\documentclass[11pt]{amsart}
\usepackage[T1]{fontenc}
\usepackage[margin=1.5in]{geometry}
\usepackage{amssymb, amsmath, amsthm}
\usepackage{graphicx, tikz}
\usepackage{txfonts}
\usepackage{booktabs}
\usepackage{xcolor}
\usepackage{enumerate}
\usepackage{subcaption}
\usepackage[numbers]{natbib}
\numberwithin{equation}{section}
\usepackage[textsize=tiny]{todonotes}
\usepackage{multirow}

\usepackage{mathtools}
\newcommand{\defeq}{\vcentcolon=}

\let\oldtocsection=\tocsection
\let\oldtocsubsection=\tocsubsection
\let\oldtocsubsubsection=\tocsubsubsection
\renewcommand{\tocsection}[2]{\hspace{0em}\oldtocsection{#1}{#2}}
\renewcommand{\tocsubsection}[2]{\hspace{1em}\oldtocsubsection{#1}{#2}}
\renewcommand{\tocsubsubsection}[2]{\hspace{2em}\oldtocsubsubsection{#1}{#2}}

\definecolor{ocean}{rgb}{0,0.1,0.6}

\definecolor{imperialGreen}{RGB}{2,137,59}
\definecolor{imperialBlue}{RGB}{0, 62, 116}
\definecolor{imperialBrick}{RGB}{165,25,0}
\definecolor{imperialProcess}{RGB}{0,133,202}

\usepackage{hyperref}
\hypersetup{colorlinks=false, linkbordercolor=imperialProcess,
citebordercolor=imperialBrick}

\newcommand{\ind}{1\hspace{-2.1mm}{1}}
\newcommand{\eps}{\varepsilon}
\newcommand{\half}{\frac{1}{2}}

\newcommand{\D}{\mathrm{d}}
\newcommand{\EE}{\mathbb{E}}
\newcommand{\ev}{\mathbb{E}}

\newcommand{\R}{\mathbb{R}}

\newcommand{\x}{\boldsymbol{x}}

\newcommand{\bt}{\boldsymbol{\theta}}

\newtheorem{theorem}{Theorem}[section]
\theoremstyle{definition}

\newtheorem{remark}[theorem]{Remark}

\makeatletter
\newcommand{\subjclassjel}[2][1991]{%
  \let\@oldtitle\@title%
  \gdef\@title{\@oldtitle\footnotetext{#1 \emph{JEL subject classification.} #2}}%
}
\graphicspath{{Figures/}}

\title[Uncertainty and Explainability in Deep Rough Volatility]
{Uncertainty and Explainability in Deep Rough Volatility:\\
A Neural Information-Theoretic Posterior Approach}

\author{Damiano Brigo}
\email{damiano.brigo@imperial.ac.uk}
\address{Department of Mathematics, Imperial College, London, United Kingdom.}
\author{Rapha\"el Huser}
\email{raphael.huser@kaust.edu.sa}
\address{Computer, Electrical and Mathematical Sciences and Engineering Division (CEMSE),
  King Abdullah University of Science and Technology (KAUST),
  Thuwal, Saudi Arabia}
\author{Dan Leonte}
\email{dan.leonte@kaust.edu.sa}
\address{Computer, Electrical and Mathematical Sciences and Engineering Division (CEMSE),
  King Abdullah University of Science and Technology (KAUST),
  Thuwal, Saudi Arabia}

\date{September 24, 2026}
\thanks{\small The opinions here expressed are solely those of the authors and do not represent in any way those of their employers.
For the purpose of open access, the authors have applied a Creative Commons Attribution (CC BY) license to any Author Accepted Manuscript version arising.}

\begin{document}
\begin{abstract}
Deep learning has substantially accelerated the calibration of complex
stochastic-volatility models, but neural point calibration alone does not capture the uncertainty remaining after an implied-volatility (IV) surface has been observed. We develop a simulation-based inference framework for rough Heston (rHeston) calibration that learns the posterior distribution of the model parameters conditional on an  
IV surface. Using neural ratio estimation, we obtain calibrated posterior samples that can be propagated through heteroscedastic neural surrogate
pricers for path-dependent exotic options.  The resulting posterior-predictive
distributions combine residual parameter uncertainty with conditional
surrogate uncertainty and yield uncertainty-aware price intervals.

We further introduce Hellinger-SHAP, an information-theoretic
explainability method for posterior inference.  Rather than attributing a
single parameter point estimate, it applies local-background Kernel SHAP to
a posterior-information functional measuring contraction from the prior to
the posterior.  This identifies maturity--moneyness regions associated with
posterior information gain for individual rHeston parameters.  In a
simulation study, posterior-predictive intervals provide calibrated or
conservative coverage across forward-start, barrier, and realized-variance
claims, while point plug-in prices can be materially unreliable for selected
contract regimes.  Together, the UQ and XAI analyses provide a transparent
framework for uncertainty-aware neural calibration and downstream exotic
pricing under the specified prior-predictive model.
\end{abstract}

\maketitle

\section{Introduction}
\label{sec:introduction}

Deep learning has substantially accelerated the calibration and valuation of
complex derivative portfolios.  In risk-critical applications, however, a
neural network that maps market data to a single parameter vector does not
by itself quantify the uncertainty that remains conditional on the observed
data. This issue is especially relevant for stochastic-volatility models,
where an observed implied-volatility (IV) surface is finite (i.e., available only at finitely many maturities and strikes) and subject to numerical simulation or 
observation errors, and may leave residual uncertainty about the
parameters under the specified model and prior. A principled treatment of
this uncertainty is needed both for downstream pricing and for meaningful
explainability analyses.

We study this problem for the rough Heston (rHeston) model, a representative
rough-volatility model (RVM)~\cite{roughvol}.  We use the rHeston
formulation in~\cite{el2019roughening, el2019characteristic, euch2018perfect}.
Rough-volatility models can
capture salient short-maturity features of equity-option markets, but their
simulation, calibration, and path-dependent pricing remain numerically
demanding. The nonlinear parameter-to-price map and the numerical
construction of an IV surface make point calibration alone an incomplete
inferential output when downstream claims depend nonlinearly on the
parameters. Moreover, propagating parameter uncertainty to such claims
requires repeated evaluations of (potentially path-dependent) option prices, for which direct
Monte Carlo simulation is computationally prohibitive.

Three specific sources of numerical fragility motivate both the uncertainty-quantification framework and the surrogate-based exotic pricing approach developed in this paper; Section~\ref{sec:roughheston} discusses
these challenges in detail, together with the adopted simulation scheme. 
First, even the semi-closed-form route to European option prices relies on a numerical
solution of a singular fractional Riccati equation, whose associated
Fourier-inversion step is notoriously unstable precisely in the
short-maturity, deep out-of-the-money regime that matters most for skew
calibration~\cite{el2019characteristic, lewis2001option}.  Second, the
volatility process is non-Markovian, so path-dependent pricing and the
repeated simulation required throughout this paper must either carry the
full process history or replace it with a finite-dimensional Markovian
lift; the lift restores tractable, affine dynamics only by introducing a
stiff, multi-scale factor system that can be expensive to
resolve~\cite{abi2019multifactor}. Third, as for classical square-root diffusions, the discretized variance process must be kept nonnegative, which requires dedicated truncation or correction schemes and their
attendant bias-variance trade-offs. Each difficulty can be controlled for
a one-off evaluation at a fixed parameter value, but together they become
prohibitive when the model must be evaluated repeatedly across
many parameter configurations. This fragility is the primary motivation
for the neural surrogate and uncertainty-quantification framework
developed in this paper.

A complementary objective is to understand how different parts of the IV surface inform parameter inference. Previous work has applied post-hoc XAI methods to neural point estimators for
the Heston and rHeston models; see~\cite{brigo2021interpretability}
and~\cite{brigo2024interpretability}. Such analyses can identify regions of
the IV surface associated with changes in a fitted point prediction, using
methods such as LIME~\cite{ribeiro2016should},
DeepLIFT~\cite{shrikumar2017learning}, LRP~\cite{bach2015pixel}, and
SHAP~\cite{lundberg2017unified}. In the present setting, however, the
inferential object of interest is a posterior distribution rather than only a
single parameter summary. We therefore explain posterior information
gain---the contraction from a chosen prior to the posterior induced by an IV
surface---rather than a point estimate alone.

Our approach combines simulation-based inference (SBI), uncertainty-aware
neural exotic-option surrogate pricers, and information-theoretic explainability.
Telescoping neural ratio estimation (TRE) learns an approximation to the
conditional posterior distribution of the rHeston parameters given an IV
surface. Posterior samples are propagated through conditional surrogate
predictive distributions to obtain posterior-predictive laws for
path-dependent exotic prices. This construction retains both residual
parameter uncertainty and the conditional uncertainty represented by the
surrogate, rather than collapsing the calibration to a single parameter
vector. Separately trained MSE- and MAE-based inverse networks are used as
point-estimation benchmarks.

Specifically, the main contributions of this work are as follows:
\begin{itemize}
    \item \textbf{Posterior inference for neural rough-volatility calibration:}
    we use SBI to learn 
    the full conditional distribution of rHeston parameters given a synthetic IV surface, and
    assess the quality of the learnt approximate distribution through simulation-based calibration checks on held-out datasets.

     \item \textbf{Posterior-predictive uncertainty quantification for exotic
    prices:} we propagate posterior draws through heteroscedastic neural
    surrogate pricers for forward-start, barrier, and realized-variance exotic
    contracts.  This yields predictive intervals that combine parameter and
    conditional surrogate uncertainty, together with a law-of-total-variance
    attribution of their relative contributions. We therefore bridge the gap between deep uncertainty quantification and practical quantitative trading.

    \item \textbf{Information-theoretic XAI:} we introduce
    Hellinger-SHAP, which uses local-background Kernel SHAP to attribute
    posterior information gain for individual parameters across the IV
    surface.  Squared Hellinger distance provides a bounded, robust complement to the
    Kullback--Leibler information-gain analysis.
\end{itemize}

We refer to the introduction of \cite{brigo2024interpretability} for a more
comprehensive review of machine-learning work on rHeston calibration and
interpretability.

The remainder of this paper is organized as follows. Section~\ref{sec:roughheston} details the rough Heston model and the generation of our
implied-volatility dataset. Section~\ref{sec:uq_methodology} introduces the
simulation-based inference framework, point-estimation benchmarks, and exotic
surrogate pricing methodology. Section~\ref{sec:xai_hellinger} develops the
Hellinger-SHAP methodology. Section~\ref{sec:simstudy} presents the numerical
study, including posterior recovery and posterior-predictive coverage. Section~\ref{sec:simstudy-xai} presents the XAI
results. Finally, Section~\ref{sec:conclusion} concludes the paper and
discusses directions for future research. Further details and simulation results can be found in the appendix. 

\section{Rough Heston model and synthetic data generation}
\label{sec:roughheston}

\subsection{Rough volatility (and rough Heston)}

We have abundant evidence that traditional SVMs cannot reproduce the volatility surfaces observed in real markets, see for example~\cite{bayer2024roughbook, bolko2023gmm}. 
As a response, SVMs driven by a fractional Brownian motion with Hurst exponent in $(0,\half)$ have been proposed.
In Gatheral, Jaisson and Rosenbaum~\cite{roughvol}, such a model was introduced based on statistical observations under the historical measure, and triggered a large research response that has extended the boundaries of such models and their 
applications~\cite{abi2019multifactor, abi2019affine, keller2018affine, el2019characteristic, euch2018perfect, guennoun2018asymptotic, jacquier2018vix}.
The overall consensus coming from these and related papers is that, while rough volatility behaviour can be mimicked by standard Markovian models, \emph{`rough volatility as the null statistical hypothesis is very hard to beat'} (quoting the original paper of Gatheral and Rosenbaum).
We now briefly introduce the rHeston model, whose 
implied volatility surface is studied in this paper.  
A fractional Brownian motion~$W^H$ is a continuous-time centered Gaussian process starting from zero, with covariance function
$$
\EE\left[W_{t}^{H} W_{s}^{H}\right]=\frac{1}{2}\left(|t|^{2 H}+|s|^{2 H}-|t-s|^{2 H}\right),
\qquad\text{for all } s,t\geq 0.
$$
Here, $H\in(0,\frac{1}{2})$ is called the Hurst parameter and describes the roughness of the paths of the process, in the sense of H\"older $(H-\eps)$-regularity for any $\eps \in (0,H)$, 
and $H=\half$ corresponds to the standard Brownian motion.
Such a process can be built on top of a two-sided standard Brownian motion~$W$ via the Mandelbrot-Van Ness representation~\cite{mandelbrot1968fractional}
\begin{equation}\label{eq:MandelbrotfBm}
W_{t}^{H} =\frac{1}{\Gamma(H+\half)}
\left(
\int_{-\infty}^{0}
\left[(t-s)^{H-\half}-(-s)^{H-\half}\right] \D W_{s}+
 \int_{0}^{t}(t-s)^{H-\half} \D W_{s}\right),
\end{equation}
almost surely for all $t\geq 0$, where~$\Gamma$ denotes the Gamma function.
Other types of fractional Brownian motions exist, and in fact the general class of interest for applications corresponds to continuous Gaussian Volterra processes of the form
\begin{equation}\label{eq:fBmKernel}
W^{H}_{t} = \int_{0}^{t}K(t-s)\D W_s,
\end{equation}
almost surely for all $t\geq 0$, for some square integrable kernel function~$K(\cdot)$ on~$[0,\infty)$ and some standard Brownian motion~$W$.
In this formulation, both~$W$ and~$W^H$ generate the same filtration~\cite{decreusefond1999stochastic}
and the H\"older regularity of~$W^H$ is generated by the local singularity behaviour of~$K(\cdot)$ around the origin.
A common example uses $K(u) = u^{H-\half}\ind_{\{u>0\}}$,
giving rise to the Riemann-Liouville (or type~II) fractional Brownian motion, 
while the representation~\eqref{eq:MandelbrotfBm} can also be written as~\eqref{eq:fBmKernel}, 
albeit with a more complicated kernel function (but also with H\"older regularity $H-\eps$ for any $\eps \in (0,H)$).

Inspired by this convolutional representation, the rough Heston model \citep{el2019characteristic} is defined as \begin{align}
\frac{\D S_{t}}{S_{t}} &= \sqrt{V_{t}} \D B_{t}, \qquad S_0 >0, \label{eq:S_t}\\
V_{t} &= V_{0} + \frac{1}{\Gamma\left(H+\frac{1}{2}\right)}
\int_{0}^{t}(t-s)^{H-\frac{1}{2}} \left\{\kappa \left(\theta-V_{s}\right) \D s
+ \nu\sqrt{V_{s}} \D W_{s}\right\},\label{eq:V_t}
\end{align}
under the risk-neutral measure (where we ignore interest rates and dividends for simplicity),
where~$S$ denotes the stock price process, $V$ the instantaneous variance process, 
$B$ and $W$ are standard Brownian motions with instantaneous covariation $\rho$ as in
\[
\mathrm{d}\langle B,W\rangle_t = \rho\,\mathrm{d}t,
\]
the ``leverage parameter'', and the parameters satisfy $\kappa,\theta,\nu,V_0>0$, $\rho\in[-1,1]$, $H\in(0,\half)$. 

A different formulation~\cite{guennoun2018asymptotic} is also possible, 
but the one from~\cite{el2019characteristic} is more amenable to computations and most popular in practice,
and the reader is referred to~\cite[Chapter~4]{bayer2024roughbook} for a full comparison. While simulating these rough paths was exceptionally computationally intensive a few years ago, a wealth of sophisticated numerical techniques have recently been devised to accelerate this process~\cite{bennedsen2017hybrid, abi2019multifactor, fukasawa2021refinement, bayerFuk2022weak, bayer2022weak, BayerBreneis_2024_weak}.
One important benefit of adopting this rHeston model is that the affine property of the standard Heston model is preserved (in an infinite-dimensional setting), so that pricing European options can be achieved in principle via Fourier transform methods, see for example 
Lewis~\cite{lewis2001option}.
Even in this comparatively favourable case of European payoffs, the rough Heston characteristic function has no closed form and is itself \emph{approximated} by a numerical solution to the fractional Riccati ODE with a singular kernel at $t=0$~\cite{el2019characteristic}, typically through an Adams predictor--corrector scheme~\cite{diethelm2002predictor} whose number of time-steps has to be chosen by the user. This approximation is then used within a Fourier-inversion pricing routine, e.g., the Lewis formula~\cite{lewis2001option} or the COS expansion~\cite{fang2009COS}, which introduces a second layer of approximation through the truncation of the semi-infinite inversion integral and through the number of quadrature or cosine nodes used to discretize it, together with the choice of a damping parameter needed to ensure integrability. Adequate values for these hyper-parameters are not trivial to determine and vary across the options surface: short maturities, where the characteristic function has fatter tails, require substantially finer settings than long ones, so in practice the parameter ranges are tuned per maturity band rather than chosen once and fixed. As a consequence, the evaluation of the characteristic function via the fractional Riccati equation is notoriously unstable for specific market regimes: for short maturities and deep out-of-the-money strikes, the Adams predictor-corrector scheme suffers from numerical instability, and the highly oscillatory integrals required by Fourier inversion fail to converge. In these extreme wing regions, one must often abandon the semi-closed-form approach entirely and resort to heavy Monte Carlo simulations.

Beyond European options, difficulties compound. For path-dependent options, Fourier approaches are fundamentally not amenable, and one is strictly forced to rely on Monte Carlo methods regardless of the strike or maturity. Moreover, the volatility process is non-Markovian and its kernel is singular at the origin, which makes the dynamics inherently difficult to simulate as $H \to 0$~\cite{abi2019multifactor}. Consequently, any task that requires repeated path evaluations, e.g., Monte Carlo pricing of path-dependent payoffs or simulation-based inference, must either carry the full history of the process or replace it by a finite-dimensional Markovian lift~\cite{abi2019multifactor, BayerBreneis_2023_MarkovianApprox}. This lift approximates the singular kernel by a weighted sum of exponentials, representing the volatility process as a superposition of mean-reverting factors, and in doing so restores non-fractional, affine, Fourier-amenable dynamics. The added dimensions, however, are not benign and constitute a heterogeneous, multi-scale system: resolving the singularity forces a subset of factors to mean-revert on extremely fast timescales, rendering the system stiff and imposing small discretization steps~\cite{BayerBreneis_2024_weak}, while the factors' speeds and weights must themselves be determined numerically, through quadrature or optimization routines. A further difficulty is shared with the classical square-root diffusions considered in Section~\ref{subsec:hqe}: the discretized variance must be kept nonnegative, which calls for dedicated truncation or correction schemes~\cite{lord2010comparison, alfonsi2025nonnegativity}. Although the discretization and stiffness costs are well controlled for one-off European pricing at a fixed parameter value, they become prohibitive in the large-scale simulation regimes required for exotic pricing and simulation-based inference, where the model must be evaluated repeatedly across many rough Heston parameter configurations.

The fact that even state-of-the-art numerical pricing on the rough manifold is fragile motivates shifting the computational burden offline: high-quality simulations are generated in advance and used to train neural surrogates that subsequently allow fast evaluations. At the same time, replacing direct numerical evaluation by learned approximations makes rigorous uncertainty quantification critical, and the framework we propose accounts for both residual uncertainty in the calibrated parameters and predictive uncertainty from the surrogate pricers.

\subsection{Simulation algorithm}\label{subsec:hqe}

We simulate the rough Heston variance process $(V_t)_{t \ge 0}$ using the hybrid quadratic-exponential (HQE) discretisation scheme \citep{HQE_Gatheral_SSRN}. HQE combines two ideas that address the main difficulties induced by roughness: i) the \emph{hybrid} scheme of \cite{bennedsen2017hybrid} from Brownian semi-stationary processes tames the singularity while staying computationally feasible, and ii) the quadratic--exponential (QE) approximate sampler~\cite{andersen2008QE}, 
adapted by~\cite{HQE_Gatheral_SSRN} to rough Heston, replaces the potentially negative Gaussian update of the volatility process by a non-negative, approximate distribution constructed through moment-matching and drastically reduces truncation bias. 

First, the standard left-point, Riemann discretisation fails to capture the steepness of the fractional kernel $(t-s)^{H-1/2}$ inside the 
intervals of the time partition immediately preceding $t$, and its error deteriorates substantially as $H \downarrow 0$
. The hybrid scheme of~\citep{bennedsen2017hybrid} resolves this by splitting the partition intervals into two groups. The intervals closest to the singularity $s=t$, referred to as the \emph{head} cells, are handled by a specialised scheme that eliminates kernel discretisation error. In the remaining intervals, the kernel varies smoothly without a singularity, and its contribution can be approximated well by a single evaluation of the kernel, or alternatively by a cell average, multiplied by a Brownian increment. This construction removes the singularity while staying computationally feasible, and the remaining approximation error arises only from freezing the variance process at the left endpoint of each interval. 
 
Second, although the paths $V_\cdot$ are non-negative, the Gaussian discretisation does not preserve this property. Standard remedies such as the Lord--Koekkoek--van Dijk (LKvD) full truncation \citep{lord2010comparison} keep the square-root term well-defined, but in our experiments, this positivity correction introduces non-negligible bias which gets worse in the relevant parameter regimes, e.g., low $H$ and $V_0$. The QE step addresses this specific issue.

We contribute to the HQE scheme on three fronts
. First, since the QE approximation only preserves the first two moments of the aforementioned \emph{head} increment, the resulting discrete asset price is no longer a martingale, and we propose a martingale correction which is available analytically and easy to implement. Second, we show that a version of the conditional representation due to Romano and Touzi~\citep{RomanoTouzi1997} survives the rough kernel despite $V_\cdot$ no longer being a semimartingale, which allows us to analytically integrate the Brownian motion orthogonal to the one driving $V_\cdot$ for a number of contracts, e.g., European and forward start options. By significantly reducing MC variance, this guarantees that our finite sample estimators satisfy some highly desirable properties, e.g., convexity of the price in the strike and hence the absence of butterfly arbitrage. Third, we consider graded, non-uniform grids and several variance reduction techniques, which further reduces the required number of discretisation steps and MC paths. 

\subsubsection{The hybrid construction and head cell decomposition}
\label{sec:hybrid_construction_and_cell_decomp}
For ease of notation, we present the uniform-grid case,  
with all changes necessary for the non-uniform case detailed in Appendix~\ref{app:hqe_non_uniform_grids}. Let $\alpha = H+1/2$ and fix a time grid $0 = t_0 < t_1 < \ldots < t_N = T$ with equidistant step-size $\Delta  = T/N$. Throughout, we write $V_{t_n}$ for the process~\eqref{eq:V_t} at a mesh point, and use a hat for the corresponding quantity produced by discretisation: $\widehat V_n\approx V_{t_n}$, and so on for all other processes.  For $t_{n+1}\le T$, splitting the integral from~\eqref{eq:V_t} cell by cell gives
\begin{equation}
V_{t_{n+1}} = V_0 + \underbrace{\frac{\kappa}{\Gamma(\alpha)}   \sum_{m=0}^{n}  \int_{t_m}^{t_{m+1}} (t_{n+1}-s)^{\alpha-1}(\theta - V_s)\,\D s}_{\text{drift}} + \underbrace{ \frac{\nu}{\Gamma(\alpha)} \sum_{m=0}^{n}  \int_{t_m}^{t_{m+1}}   (t_{n+1}-s)^{\alpha-1}\sqrt{V_s}\,\D W_s}_{\text{diffusion}}.
\label{eq:V-cell-decomposition}
\end{equation}

For the drift term, we freeze $V_s \approx V_{t_m}$ at the left endpoint of each cell $[t_m,t_{m+1}]$. The same approximation is also used for the diffusion on the history cells. However, by It\^{o}'s isometry, the mean-square error of the diffusion approximation involves the squared kernel $(t_{n+1}-s)^{2H-1}$, which is more singular near $s=t_{n+1}$ than the kernel $(t_{n+1}-s)^{H-1/2}$ appearing in the drift term. We therefore use left-point freezing in the drift and in the diffusion for all $m<n$, while treating the current \emph{head} cell $[t_n,t_{n+1}]$ separately. To this end, for a lag $j\defeq n-m \in \{0,\ldots,n\}$, denote the (Toeplitz) drift weights $a_{n,m} \defeq a_j$ by

\begin{equation}
a_j \defeq \frac{1}{\Gamma(\alpha)}\int_{t_m}^{t_{m+1}} (t_{n+1}-s)^{\alpha-1}\,\D s = \frac{\Delta^{\alpha}}{\Gamma(\alpha+1)}\bigl[(j+1)^{\alpha}-j^{\alpha}\bigr].
\label{eq:aj}
\end{equation}
The drift term is then approximated by $\kappa \sum_{m=0}^n a_{n-m} (\theta - \widehat{V}_m^{+}),$ where $\widehat V_n^{+} = \max\{\widehat V_n,0\}$ 
implements LKvD full truncation; this amounts to retaining potentially negative values, and only clipping when consuming the volatility.  For the diffusion term, we split the \emph{history} cells $m<n$, where the kernel is not singular and can be replaced by the constant cell average $\frac{a_{n-m}}{\Delta}$, from the \emph{head} cell $m=n$, where the kernel is singular and requires special treatment
. 
Define the exact head increment of the variance process and the exact stochastic increment 
\begin{align*}
u_n &\defeq \frac{\nu}{\Gamma(\alpha)}\int_{t_n}^{t_{n+1}} (t_{n+1}-s)^{\alpha-1}\sqrt{V_s}\,\D W_s,
\\
\chi_n &\defeq \int_{t_n}^{t_{n+1}} \sqrt{V_s}\,\D W_s,
\end{align*}
Let the simulated counterparts be $\widehat u_n$ and $\widehat \chi_n$. Then the diffusion term is approximated by $\frac{\nu}{\Delta} \sum_{m=0}^{n-1} a_{n-m} \widehat \chi_m + \widehat u_n$, where the index stops at $m=n-1$ to not double-count the head cell. Grouping the drift with the diffusion $m<n$ \emph{past} cells, define 
\begin{align}
\widehat\Xi_n \defeq & V_0 + \kappa\sum_{m=0}^{n} a_{n-m}\left(\theta-\widehat V_{m}^{+}\right) + \frac{\nu}{\Delta} \sum_{m=0}^{n-1}a_{n-m}\widehat\chi_m, \label{eq:xi_n_hat}\\
\widehat V_{n+1} =& \widehat\Xi_n + \widehat u_n,  \label{eq:v_n_1_u_hat_increment}
\end{align}
Note that for $m < n$, $\widehat \chi_m$ is sampled at the $m^{\text{th}}$ step and reused (rather than sampled again) 
at step $n$. It is also worth emphasizing that $u_n$ and $\chi_n$ are driven by the same infinitesimal Brownian motion $\mathrm{d}W_s$, therefore dependent, and the crux of the discretization scheme is to retain the joint law of $(u_n,\chi_n)$, or at least the first two moments. Formally, let $\mathcal{G}_n =\{\widehat{V}_0,\ldots,\widehat{V}_n, \widehat{\chi}_0, \ldots, \widehat{\chi}_{n-1}\}$ be the information available from the discretized scheme after simulating $\widehat{V}_n$ and before simulating $\widehat{V}_{n+1}$. The hybrid and HQE schemes differ precisely in the choice of $(\widehat u_n, \widehat\chi_n)$.  
 In the hybrid scheme, $u_n$ and $\chi_n$ are approximated through $V_s\approx \widehat V_n$ for $s \in [t_n,t_{n+1}]$, which yields
\begin{align*}
    \widehat u_n = & \frac{\nu}{\Gamma(\alpha)} \sqrt{\widehat V_n^{+}} \int_{t_n}^{t_{n+1}} (t_{n+1}-s)^{\alpha-1}\,\D W_s, \\
    \widehat \chi_n =  & \sqrt{\widehat V_n^{+}} (W_{t_{n+1}} - W_{t_n}),
\end{align*}

Then the pair $(\widehat u_n, \widehat \chi_n) | \mathcal{G}_n$ is a bivariate Gaussian with conditional mean $\bf{0}$ and variance-covariance matrix given by
\begin{equation*} 
\widehat{V}_n^{+}
\begin{pmatrix} 
\frac{\nu^2}{\Gamma(\alpha)^2} \int_{t_n}^{t_{n+1}} (t_{n+1}-s)^{2(\alpha-1)}  \mathrm{d}s & \frac{\nu}{\Gamma(\alpha)}  \int_{t_n}^{t_{n+1}} (t_{n+1}-s)^{\alpha-1} \mathrm{d} s
\\ \frac{\nu}{\Gamma(\alpha)}  \int_{t_n}^{t_{n+1}} (t_{n+1}-s)^{\alpha-1} \mathrm{d} s & \int_{t_n}^{t_{n+1}} \mathrm{d}s
\end{pmatrix}
= \widehat{V}_n^{+}
\begin{pmatrix} 
\frac{\Delta^{2H} \, \nu^2}{2H \, \Gamma(\alpha)^2}  & \frac{\nu \Delta^\alpha}{\Gamma(\alpha+1)}  
\\ \frac{\nu \Delta^\alpha}{\Gamma(\alpha+1)}    & \Delta
\end{pmatrix},
\end{equation*}
and drawing samples is trivial. Also, since $\ev [\widehat u_n \mid \mathcal{G}_{t_n}] = 0 $, we have that $\widehat \Xi_n = \ev[\widehat V_{n+1} \mid \mathcal{G}_{n}]$ can be interpreted as the one-step-ahead conditional mean predictor of $\widehat V_{n+1}$.

Nevertheless, this approach does not enforce non-negativity by default. The clipped path $\widehat{V}_\cdot^{+}$ systematically overstates the volatility, biasing the implied volatility upwards. Further, when $\widehat V_n^{+} =0$, the diffusion term from~\eqref{eq:V-cell-decomposition} is identically zero, thus maintaining $V_\cdot \equiv 0$ until the drift returns above $0$, which means $V_\cdot$ is truncated for non-negligible periods of time. Importantly, the truncation issue aggravates as either $H$ or $V_0 \to 0$.

\subsubsection{The HQE scheme}
We now discuss the HQE scheme of \cite{HQE_Gatheral_SSRN}, which improves the approximation of $V_s$ on the head-cell $[t_n,t_{n+1}]$ and also enforces non-negativity by design. 

To begin with, note that as $H \downarrow 0$ in $u_n  = \nu\int_{t_n}^{t_{n+1}} (t_{n+1}-s)^{\alpha-1}\sqrt{V_s}\,\D W_s$, the singular kernel ${(t_{n+1}-s)}^{\alpha-1} = {(t_{n+1}-s)}^{H-1/2}$ assigns most of the mass towards the right endpoint $V_{t_{n+1}}$ and negligible weight on the left endpoint $V_{t_n}$. Thus, although the singularity itself is treated well by the hybrid scheme, approximating the variance on the entire cell by left endpoint $\widehat{V}_n$ loses accuracy as $H \downarrow 0$.  
HQE proposes the $1^{\text{st}}$ order approximation $$\ev[V_s \mid \mathcal{G}_{n}] \approx \frac{(s-t_n)}{\Delta} \widehat {\Xi}_n + \frac{(t_{n+1}-s)}{\Delta} \widehat V_n $$ given by the linear interpolation between $\widehat V_n$ and $\widehat \Xi_n = \ev[\widehat V_{n+1} \mid \mathcal{G}_{n}]$. By direct calculation and It\^{o}'s Lemma, we obtain
$\mathrm{Var}(u_n \mid \mathcal{F}_{t_n}) \approx 
\frac{\nu^2 \, \Delta^{2H}}{\Gamma(\alpha)^{2} \, 2H} \,\bar v_n,$ 
where 
\begin{equation}
    \bar v_n = \frac{2H}{2H+1} \widehat V_n+ \frac{1}{2H+1}\widehat\Xi_n. 
    \label{eq:v_bar_n}
\end{equation}
The weighted average \eqref{eq:v_bar_n} improves the local approximation of the first two conditional moments, but it does not by itself produce a non-negative update for the variance process. If one were to retain a Gaussian head-cell update at this stage, the resulting one-step approximation still becomes negative with non-negligible probability, especially in the rough regimes of small $H$ or small $V_0$, and truncation-based fixes reintroduce undesirable bias. HQE therefore combines the hybrid discretisation of the singular Volterra kernel with Andersen's quadratic--exponential (QE) moment-matching construction, which enforces positivity at the one-step level while preserving the local mean--variance information encoded in \eqref{eq:v_bar_n}. Accordingly, freeze $V_s$ to the weighted average $\bar v_n$ on $[t_n, t_{n+1}]$ (see \cite{HQE_Gatheral_SSRN} for the formal treatment), which then gives the approximate conditional variance-covariance matrix 
\begin{align*}
\widehat \Sigma_n = 
\mathrm{Var}\left({\begin{pmatrix}
\widehat{u}_n  \\\widehat{\chi}_n
\end{pmatrix} \Bigg| \mathcal{G}_n
}
\right)
=  
\bar{v}_n \begin{pmatrix} 
    \frac{\nu^2 \Delta^{2H}}{2H\,\Gamma(\alpha)^{2}}  &  \frac{\nu \Delta^{\alpha}}{\Gamma(\alpha+1)}\\  \frac{\nu \Delta^{\alpha}}{\Gamma(\alpha+1)} & \Delta
    \end{pmatrix},
    \end{align*}
It is worth noting that as $H \to 1/2$, $\bar v_n \to \frac{1}{2}(\widehat V_n + \widehat \Xi_n)$ recovers the quadrature rule from the standard, non-rough Heston model. In the rough limit  $H \downarrow 0$, the singularity dominates and the weighted average collapses to the right endpoint prediction $\bar v_n \to \widehat \Xi_n$. Having determined more accurate approximations for the first two conditional moments by replacing $\widehat{V}_n$ with $\bar{v}_n$, we next enforce positivity via Andersen's Quadratic-Exponential (QE) sampler.

We write $X\sim\mathrm{QE}(m,s^2)$ for a non-negative QE random variable with mean $m$ and variance $s^2$. The sampler switches between a quadratic-Gaussian and a zero-inflated exponential representation depending on the relative variance $s^2/m^2$; the explicit sampling formulas are given in Appendix~\ref{app:HQE_app}. Conditionally on $\mathcal G_n$, draw two independent QE variables
\begin{align}
Y_n
&\sim
\mathrm{QE}\left(
\frac{\widehat\Xi_n}{2},
\bar v_n\frac{\nu^2\Delta^{2H}}{\Gamma(\alpha+1)^2}
\right), \label{eq:Y_n}
\\
E_n
&\sim
\operatorname{QE}\left(
\frac{\widehat\Xi_n}{2},
\bar v_n\frac{\nu^2\Delta^{2H}}{\Gamma(\alpha)^2}
\frac{(H-\frac12)^2}{2H\alpha^2}
\right),\label{eq:E_n}
\end{align}
and reconstruct
\begin{equation}
\widehat V_{n+1}=Y_n+E_n,
\qquad
\widehat\chi_n
=
\frac{\Gamma(\alpha+1)\Delta^{1-\alpha}}{\nu}
\left(Y_n-\frac{\widehat\Xi_n}{2}\right),
\qquad
\widehat u_n=\widehat V_{n+1}-\widehat\Xi_n.
\label{eq:hqe-qe-reconstruction}
\end{equation}
Since $Y_n,E_n\geq0$, this guarantees $\widehat V_{n+1}\geq0$. Since the two QE samples $Y_n$ and $E_n$ are conditionally independent and because of their prescribed means and variances, then
\[
\ev[\widehat u_n\mid\mathcal G_n]=\ev[\widehat\chi_n\mid\mathcal G_n]=0,
\qquad
\operatorname{Var}\!\left(
\begin{pmatrix}\widehat u_n\\ \widehat\chi_n\end{pmatrix}
\Bigm|\mathcal G_n
\right)=\widehat\Sigma_n.
\]
Thus HQE preserves the first two conditional moments of the head-cell construction while enforcing non-negativity. It is worth keeping in mind though that the resulting $\widehat\chi_n$ is no longer conditionally Gaussian.

\subsubsection{Martingale correction, Romano-Touzi and other improvements} 
Recall that the Brownian motion driving the stock price admits the orthogonal decomposition
\begin{equation}
B_t = \rho W_t + \sqrt{1-\rho^2}\,W_t^\perp,
\label{eq:orthogonal-bm}
\end{equation}
where $W$ and $W^\perp$ are independent Brownian motions. 
With
$X_t\defeq\log S_t$, It\^{o}'s formula gives 
\begin{equation*}
X_{t_{n+1}}-X_{t_n}
=
-\frac12\int_{t_n}^{t_{n+1}}V_s\,\D s
+ \rho \int_{t_n}^{t_{n+1}} \sqrt{V_s}\,\D W_s 
+\sqrt{1-\rho^2}
\int_{t_n}^{t_{n+1}}\sqrt{V_s}\,\D W_s^\perp.
\end{equation*}
over $[t_n,t_{n+1}]$. In the following, let $Z_n^\perp$ denote the standard Gaussian increment associated with $W^\perp$ over the interval $[t_n,t_{n+1}]$ and let $\widetilde v_n \defeq \half(\widehat V_n+\widehat V_{n+1}) \approx \frac{1}{\Delta} \int_{t_n}^{t_{n+1}}V_s\,\D s$. Then the (uncorrected) log-price increment is given by
\begin{equation}
\Delta\widehat {X}_n^u = -\half\widetilde v_n\Delta + \rho\widehat\chi_n + \sqrt{1-\rho^2}\sqrt{\widetilde v_n\Delta}\,Z_n^{\perp},
\label{eq:price-uncorrected}
\end{equation}
where $Z_n^\perp \sim \mathcal{N}(0,1)$. While the continuous-time stock price is a martingale, the HQE discretisation does not satisfy this property in general, because i) the QE approximations changes the conditional exponential moments of $\widehat\chi_n$, and ii) the trapezoidal approximation $\widetilde v_n=\half(\widehat V_n+\widehat V_{n+1})$ is itself random at time $t_n$ and correlated with $\widehat\chi_n$ through $V_{n+1}$, hence with $\hat{u}_n$. Note that preserving the martingale property under the HQE discretisation is equivalent to 
\begin{equation*}
\ev\left[e^{\Delta\widehat X_n}\mid\mathcal G_n\right]=1.
\end{equation*} 
We therefore propose the (corrected) update
\begin{equation}
\Delta\widehat X_n = -\half \widetilde v_n\Delta  + \rho\widehat\chi_n - C_n + \sqrt{1-\rho^2}\sqrt{\widetilde v_n\Delta}\,Z_n^{\perp},
\label{eq:price-corrected}
\end{equation}
where 
\begin{equation*}
C_n \defeq \log \ev \left[e^{\Delta\widehat X_n^{(0)}}\mid\mathcal G_n\right]
\end{equation*}
is available analytically; see Appendix~\ref{app:QE_and_MGF_for_QE} for the derivation and closed-form expression. This yields $\ev[e^{\Delta\widehat X_n}\mid\mathcal F_{t_n}]=1$ at negligible computational cost and enforces the martingale property. This correction is also practically important beyond the martingale property itself. Since the conditional Monte Carlo representation used to generate the
European-option prices underlying the IV dataset is built on the same
corrected log-price dynamics, it yields a forward convention that is
numerically consistent with the simulated pricing engine. In particular,
the implied volatilities obtained by put--call parity and forward-based
inversion, and hence the IV surfaces used as inputs to the inverse
calibration and posterior-inference procedures, are computed under this
same convention. The martingale correction therefore removes a small but
systematic inconsistency that the QE replacement would otherwise introduce.
\begin{remark}
    Note that the trapezoidal approximation
$\widetilde v_n=\half(\widehat V_n+\widehat V_{n+1})$
provides a more accurate approximation of the integrated variance over the cell than the left-point approximation $\widetilde v_n=\widehat V_n$. Even if the latter was used instead,
the martingale correction would still be needed, since the QE scheme is such that the uncorrected
asset-price update does not satisfy
$\ev[e^{\Delta\widehat X_n}\mid\mathcal G_n]=1$ in general. 
\end{remark}

We now turn our attention to variance reduction techniques.  
The decomposition \eqref{eq:orthogonal-bm} immediately suggests a
conditional Monte Carlo estimator. Since the HQE construction
only modifies the variance-driven increment $\widehat\chi_n$, the orthogonal
Gaussian innovation $Z_n^\perp$ remains independent of the simulated
variance path. Conditioning on the latter therefore yields a Gaussian
mixing representation similar to the Romano--Touzi trick \citep{RomanoTouzi1997}.
Summing the corrected log-price increments
\eqref{eq:price-corrected} over the time grid, define
\begin{equation}
\widehat M \defeq \sum_{n=0}^{N-1}\widehat\chi_n,
\qquad
\widehat Q \defeq \sum_{n=0}^{N-1}\widetilde v_n \Delta,
\qquad
\widehat C \defeq \sum_{n=0}^{N-1} C_n.
\end{equation}
The log-price at time $T$ can then be expressed as
\begin{equation}
\log \widehat S_T =
 -\frac{1}{2}\widehat Q + \rho \widehat M
-\widehat C + \sqrt{1-\rho^2}
\sum_{n=0}^{N-1} \sqrt{\widetilde v_n\Delta}\, Z_n^\perp.
\label{eq:logST}
\end{equation}
Using the fact that the variables $Z_n^\perp$ are independent standard Gaussians, it follows that
\[
\sum_{n=0}^{N-1}
\sqrt{\widetilde v_n\Delta}\,
Z_n^\perp
\;\Big|\;
\mathcal{G}_n 
\sim
\mathcal N(0,\widehat Q).
\]
Therefore
\begin{equation}
\log \widehat S_T
\mid \mathcal{G}_n
\sim
\mathcal N
\left(
\rho \widehat M
-\half \widehat Q
-\widehat C,
\,
(1-\rho^2)\widehat Q
\right),
\label{eq:mixing-law}
\end{equation}
There are two main types of options we must consider. For contracts such as European and forward-start options, the
orthogonal Brownian motion can be integrated out analytically instead of simulated. Besides substantially reducing the Monte Carlo variance, for European calls, each conditional price is decreasing and convex in the strike; the finite-sample Monte Carlo average therefore inherits these properties and, in particular, is free of butterfly arbitrage. For path-dependent contracts such as barriers, although a closed-form integration over $W^\perp$ as in~\eqref{eq:mixing-law} is generally not available, we can exploit a similar structure: for each simulated variance path, we draw multiple independent realizations of $W^\perp$, each paired with its antithetic counterpart, and construct the corresponding asset-price paths without repeating the computationally expensive rough variance discretisation. Averaging these inner replicates almost eliminates the MC variability arising from $W^\perp$.

Neither the martingale correction nor the other techniques presented in this section have been employed in the existing works \citep{HQE_Gatheral_SSRN,italian_thesis} or computer implementations \citep{gatheral_HQE_github,breneis_bayer_github_stoch_vol_models} for the HQE scheme. Finally, we also employ graded time grids and antithetic sampling adapted to the QE construction. On a graded grid, the same construction applies broadly, at the cost of losing the Toeplitz structure available on a uniform grid, see Appendix~\ref{app:HQE_app} for details.

\section{Simulation-Based Inference methodology}\label{sec:uq_methodology}

\label{sec:TRE_and_NBE}
Throughout the paper, $\bt = (\kappa, \theta, \nu, \rho, V_0, H) \in \Theta \subset \R^m$ with $m=6$ is parameterizing the rough Heston model, and $\x$ denotes market data; in our case, $\x$ is an implied-volatility (IV) surface, whose detailed construction is deferred to Section~\ref{sec:simstudy}.
This section is organized around two inferential tasks. Subsection~\ref{sec:sbi-landscape} addresses the inverse problem of recovering model parameters from market data. Given an observed IV surface $\x_o$, we train neural networks (NN) that return either a point estimator $\hat{\bt} \defeq \hat{\bt}(\x_o)$ or a full posterior distribution $p(\bt \mid \x_o)$. Posterior uncertainty quantification is essential because, under the
specified prior-predictive simulator, an observed finite and numerically
generated IV surface can leave residual uncertainty about $\bt$. A point
estimate alone therefore does not reflect the residual uncertainty that can propagate downstream into
pricing uncertainty.
Subsection~\ref{sec:sbi-surrogate} studies the forward map, from parameters to prices. A second NN, which we refer to as the surrogate pricer, inputs $\bt$ and returns the price of an exotic option together with an error scale quantifying the discrepancy between the surrogate output and the actual price, derived by Monte Carlo (MC). Composing the two maps amounts to propagating posterior draws of $\bt$ through the surrogate, while also accounting for the surrogate's estimated error distribution. 
 This yields a predictive distribution for each exotic price, whose quantiles may be interpreted as model-implied bid and ask quotes. The distribution reflects both i) parameter uncertainty after observing market data, represented by $p(\bt \mid \x_o)$, and approximation uncertainty arising from replacing repeated MC 
 evaluations with the surrogate model.

The methodology has two stages. Offline, we simulate a dataset of parameter--data pairs $(\bt,\x)$ once and use it to train point and posterior estimators for the inverse map, as well as surrogate pricers for the forward map. 
Online, a newly
observed IV surface is passed through the trained models to deliver either a
point calibration together with the corresponding surrogate exotic prices, or a
posterior distribution over parameters together with the induced predictive
distribution over exotic prices. The training cost is therefore incurred once
and amortized across subsequent calibrations.
\subsection{The neural point and density estimation landscape}
\label{sec:sbi-landscape}

Parameter calibration is often formulated as a point-estimation problem \citep{hernandez2017,horvath2019deep,romer2022empirical}: given an observed IV surface $\x \in \mathcal X$, one selects the parameter value $\hat\bt$ whose model-implied surface best matches it and then treats that value as correct. We instead adopt a Bayesian formulation, assigning $\bt$ a prior density $p(\bt)$ on $\Theta$ and inferring the posterior $p(\bt \mid \x)$. Point summaries can be recovered from the posterior, but not conversely.

Evaluating the posterior requires access to the likelihood $\bt, \x \mapsto p(\x \mid \bt)$, which is intractable for the rough Heston model, whereas simulating from the model, and subsequently the IV surface $\x$ for any $\bt$ is fast through the HQE scheme from Section~\ref{subsec:hqe}; this is the regime SBI is designed for. Modern SBI is dominated by amortized, neural methods see \citep{zammit_mangion_annual_review_2025} for a review. Among the available methods, we employ neural ratio estimation (NRE) \citep{Cranmer2015ApproximatingLR,amortized_hermans_2020} for three reasons. First, its classification formulation places no parametric or architectural restrictions on the posterior (as in normalizing flows). Second, it enables fast posterior inference: the posterior density is available in a single NN forward pass, and, through the telescoping extension from \citep{rhodes_tre}, posterior sampling is MCMC-free, vectorizable, and GPU-friendly. By contrast, in flow-matching and diffusion models \citep{lipman2024flowmatchingguidecode,songscore}, each density evaluation requires integrating the divergence of the learned vector field along an ODE trajectory, which is not fully and efficiently vectorizable for GPU use. Third, this formulation inherits the calibration toolbox of binary classification, which supplies additional simulation-based checks on, and post-training corrections to, the quality of the learnt approximation.

Instead of estimating the likelihood or the posterior directly, NRE replaces the intractable density-estimation problem with a binary classification one. The key insight is that a classifier trained to distinguish samples from two distributions implicitly learns their density ratio; NRE applies this to the joint $p(\x, \bt)$ and the product of marginals $p(\x) p(\bt)$ to learn the likelihood-to-evidence ratio:
\begin{equation}
\label{eq:ratio}
r(\x, \bt) :=
\frac{p(\x \mid \bt)}{p(\x)}
=
\frac{p(\x, \bt)}{p(\x) p(\bt)}
=
\frac{p(\bt \mid \x)}{p(\bt)},
\end{equation}
where $p(\x) = \int p(\x \mid \bt)p(\bt)\mathrm{d}\bt$ is the evidence. Formally, a classifier $c(\cdot, \cdot): \mathcal X \times \Theta \to [0, 1]$ trained to minimize the binary cross-entropy (BCE) loss
\begin{equation}
\label{eq:nre-bce}
\mathcal L(c) = -\mathbb E_{p(\x,\bt)}[\log c(\x, \bt)] - \mathbb E_{p(\x)p(\bt)}[\log\{1-c(\x, \bt)\}],
\end{equation}
to distinguish between samples from $p(\x, \bt)$ and $p(\x)p(\bt)$, has the Bayes-optimal solution \citep{amortized_hermans_2020}
\begin{equation}
\label{eq:nre-optimal}
c^\star(\x, \bt) =
\frac{p(\x, \bt)}{p(\x, \bt) + p(\x) p(\bt)},
\qquad\text{so that}\qquad
r(\x, \bt) =
\frac{c^\star(\x, \bt)}{1-c^\star(\x, \bt)}.
\end{equation}
In practice, the loss from \eqref{eq:nre-bce} is unavailable, and we replace it by its Monte Carlo (MC) approximation over simulated pairs and minimize it over a flexible class of functionals that satisfy a universal approximation theorem, e.g., the neural-network class $\{c_{\boldsymbol\psi}\}$, parameterized by weights and biases $\boldsymbol\psi$. The fitted weights $\hat{\boldsymbol\psi}$ yield the approximation
\begin{equation}
\hat r(\x, \bt) :=
\frac{c_{\hat{\boldsymbol\psi}}(\x, \bt)}{1-c_{\hat{\boldsymbol\psi}}(\x, \bt)}
\approx r(\x, \bt).
\end{equation}
The posterior can be recovered up to classifier approximation error as $p(\bt \mid \x) \approx p(\bt) \hat r(\x, \bt)$. Finally, note that both classes are cheap to simulate: drawing $\bt \sim p(\bt)$ and then $\x \sim p(\cdot \mid \bt)$ produces pairs from $p(\x, \bt)$; replacing $\bt$ in each pair with a fresh draw from the prior breaks the dependence between $\x$ and $\bt$ while preserving the marginals, hence producing pairs from $p(\x)p(\bt)$.

Applying \eqref{eq:ratio} directly becomes numerically unreliable as the parameter dimension $m$ grows \citep{rhodes_tre}. Indeed, the joint $q$ concentrates like a point-mass relative to $\tilde q$, and the classifier increasingly separates the two classes without accurately learning the level curves of $c^\star$, and consequently, of the posterior. In machine learning terms, the loss function saturates and cannot tell a good classifier from a poor one \citep{rhodes_tre,songunderstanding}. To bypass this issue, we follow the telescoping ratio estimation (TRE) development of \citep{rhodes_tre} and learn the ratio sequentially in the coordinates of $\bt$. Writing $\bt^{1:i} = (\theta^1, \ldots, \theta^i)$, this sequential decomposition gives
\begin{equation}
\label{eq:tre}
r(\x, \bt) =
\prod_{i=1}^m r_i(\x, \bt^{1:i}),
\qquad\text{where}\qquad
r_i(\x, \bt^{1:i}) =
\frac{p(\theta^i \mid \x, \bt^{1:i-1})}
{p(\theta^i \mid \bt^{1:i-1})}
=
\frac{p(\theta^i \mid \x, \theta^1, \ldots, \theta^{i-1})}
{p(\theta^i \mid \theta^1, \ldots, \theta^{i-1})}.
\end{equation}
For each $i$, the factor $r_i$ is estimated by training a binary classifier to distinguish between samples $\theta^i$ drawn from the densities $\theta^i \mapsto p(\theta^i \mid \x, \bt^{1:i-1})$ and $\theta^i \mapsto p(\theta^i \mid \bt^{1:i-1})$, conditionally on $\x, \bt^{1:i-1}$. Because each classification subproblem introduces only one coordinate of $\bt$ at a time, the two classes overlap substantially and the loss saturation that plagues NRE is alleviated \citep{rhodes_tre}. The factorized structure also yields a fast, MCMC-free posterior sampler, which is essential in financial applications. For example, propagating parameter uncertainty into exotic prices (Subsection~\ref{sec:sbi-surrogate}) requires drawing thousands of posterior samples within the strict latency limits of a trading desk, a regime where MCMC algorithms are simply too slow.

Crucially, each learned factor $r_i$ gives access to a one-dimensional conditional density
\begin{equation}
p(\theta^i \mid \x, \bt^{1:i-1}) = r_i(\x, \bt^{1:i}) p(\theta^i \mid \bt^{1:i-1}).
\end{equation}
With only a small number of NN evaluations, we can construct a polynomial approximation of this one-dimensional density that is accurate to machine precision \citep{olver2013fast}. Because the cumulative distribution function (CDF) is simply the antiderivative of this density, it can be computed through direct algebraic manipulation of the approximate polynomial coefficients, completely bypassing the need for further NN evaluations. This permits rapid inverse transform sampling, proceeding sequentially coordinate by coordinate: we first draw $\theta^1 \sim p(\theta^1 \mid \x)$, then $\theta^2 \sim p(\theta^2 \mid \x, \theta^1)$, and so on, up to $\theta^m \sim p(\theta^m \mid \x, \theta^1, \ldots, \theta^{m-1})$. Consequently, generating a large number of independent posterior samples is almost instantaneous, fully vectorizable, GPU-friendly; see \citep{rhodes_tre} for details.

\textbf{Changing the prior at inference time.}
A useful consequence of \eqref{eq:ratio} is that the learned ratio does not depend on the distribution used to generate the training parameters, beyond a normalising constant. Let $\pi$ be any prior with $\operatorname{supp}\pi(\bt) \subseteq \operatorname{supp}p(\bt) \subset \Theta$, which ensures the trained NN is not extrapolating out of the training distribution. Since $\pi(\bt \mid \x) \propto \pi(\bt)r(\x, \bt)$ while $p(\bt \mid \x) = p(\bt)r(\x, \bt)$, the two posteriors differ only through the ratio of priors,
\begin{equation}
\label{eq:priorchange}
\frac{\pi(\bt \mid \x)}{p(\bt \mid \x)}
\propto
\frac{\pi(\bt)}{p(\bt)}
=: w(\bt),
\end{equation}
with $w$ available in closed form, as we choose both densities. Thus, approximate posterior samples under $\pi$ can be obtained by reweighting posterior samples under $p$ according to $w(\bt)$. This is particularly important in finance, where priors encode market regimes.

\begin{remark}[Simulation designs]
\label{rem:grid}
Deep-calibration studies use a range of simulation designs, including
space-filling or randomized designs and, in some numerical experiments,
fixed grids; see, for example,
\citep{liu2019neural,baschetti2024deep}.  Here the random design
$\bt\sim p$ is not merely a numerical choice: it induces a joint law for
simulated pairs $(\bt,\x)$ under which posterior estimation, posterior
summaries, and direct point estimation can all be viewed as learning
different functionals of the same probabilistic object.  The ratio $r$
identifies the full conditional law, while squared-error and
coordinatewise absolute-error losses target the conditional mean and
conditional median, respectively
\citep{zammit_mangion_annual_review_2025, matt_american_statistician}. 
A deterministic grid may still be paired with explicit quadrature weights
or regarded as an empirical design distribution.  Without such a choice,
however, it does not directly induce the prior-predictive joint law that
underlies the posterior and risk interpretations used here.
\end{remark}

For comparison with point calibration, we also train neural inverse maps
$f_{\boldsymbol{\eta}}:\x\mapsto\bt$.  A network trained with mean
squared-error (MSE) loss targets the conditional mean under the simulation design,
whereas 
mean absolute-error (MAE) training targets conditional medians;
see Remark~\ref{rem:grid}.  These directly trained inverse maps are
implemented separately from the TRE posterior estimator.  Posterior means and
medians computed from $\hat p(\bt\mid\x)$ have related population targets, but
need not coincide numerically with the outputs of separately trained MSE- and
MAE-based networks because of finite-sample, architectural, optimization, and
posterior-approximation errors.  Moreover, a direct inverse point estimator is
tied to its training design, whereas the ratio-based posterior can be
reweighted under an alternative prior with support contained in the training
support.
It is worth mentioning the relation to  forward-surrogate calibration. Indeed, an alternative deep-calibration strategy first learns the forward map from
model parameters to vanilla-option prices or implied volatilities and then
obtains a calibration by minimizing a weighted discrepancy between the learned
surface and the observed market surface; see, for example,
\citep{horvath2019deep}.  This approach is highly useful when a practitioner
requires a point calibration under a specified fitting objective, possibly
incorporating quote weights, bid--ask spreads, missing observations, or
parameter constraints.  It reintroduces an optimization-based procedure at
inference time, however, and its output is a single selected parameter vector.

Our aim is complementary.  Under the prior-predictive data-generating law, an
IV surface can leave residual uncertainty about $\bt$.  The TRE therefore
targets the full conditional law $p(\bt\mid\x)$ rather than a particular
optimizer-dependent point solution.  This distinction is consequential for
downstream nonlinear payoffs: propagating only one calibrated vector discards
the residual parameter uncertainty represented by $p(\bt\mid\x)$, whereas the
posterior-predictive construction in \eqref{eq:predictive} retains it.  The
MSE- and MAE-trained inverse networks are included as point-estimation
benchmarks, not as substitutes for a conventional weighted forward-surface
calibration.

\subsection{Propagating posterior uncertainty through exotic surrogates}
\label{sec:sbi-surrogate}
Let $\Pi(\cdot \mid \bt)$ denote the distribution of the surrogate prediction conditional on parameters $\bt$, which may be degenerate in the deterministic-surrogate case or non-degenerate if the surrogate returns both a predictive mean and a predictive uncertainty. The downstream predictive law for an exotic price $P^\textrm{exo}$ given observed market data $\x_o$ is then
\begin{equation}
\label{eq:predictive}    
p(P^\textrm{exo} \mid \x_o)
=
\int \Pi(P^\textrm{exo} \mid \bt)\, p(\bt \mid \x_o)\, \mathrm{d}\bt.
\end{equation}
This construction combines two distinct sources of uncertainty represented
by the fitted pipeline.  First, $p(\bt \mid\x_o)$ describes residual
parameter uncertainty conditional on the observed IV surface under the
specified prior-predictive simulator.  Second, $\Pi(\cdot\mid\bt)$
represents the conditional uncertainty of the exotic surrogate, including its
approximation discrepancy and the Monte Carlo variability absorbed by its
training targets.  The posterior is therefore not an uncertainty quantity
computed after a separate optimization-based calibration; it is an
inferential object conditional on $\x_o$, which can be summarized by a point
decision or propagated directly to exotic prices. 
We approximate $\Pi$ by a neural network $\Pi_{\boldsymbol{\xi}}$ with weights and biases $\boldsymbol{\xi}$; the remainder of this subsection details the methodology and training details.

Fix a single exotic contract, i.e., its payoff, strike and maturity. For each $\bt \in \Theta,$ the corresponding price $P^\textrm{exo}(\bt)$ under the rough Heston model can be estimated by Monte Carlo (MC) with the HQE simulator from Section~\ref{subsec:hqe}. The surrogate pricer replaces these repeated 
MC evaluations at inference time with a single forward pass, with price--parameter pairs generated under the same random design $\bt \sim p(\bt)$ as in Section~\ref{sec:sbi-landscape}. Two aspects are in order. First, even for a fixed contract, the price can vary by orders of magnitude as $\bt$ ranges over $\Theta$, and a loss in absolute terms would concentrate training effort only on regions where prices are large and let the relative error deteriorate where these are small. We therefore train with prices in log-space, regressing
\begin{equation}
\label{eq:logprice}
y(\bt) := \log\left(P^\textrm{exo}(\bt) + \epsilon\right),
\end{equation}
where the jitter $\epsilon \geq 0$ prevents numerical instability coming from settings where none of the MC paths give a positive price, e.g., deep out-of-the-money options. Second, we design the surrogate NN to also quantify its discrepancies from the MC prices. To this end, let it output both a location $\mu_{\boldsymbol{\xi}}(\bt)$ and a scale $\sigma_{\boldsymbol{\xi}}(\bt) > 0$ in log space, trained by minimising the Gaussian negative log-likelihood
\begin{equation}
\label{eq:surrogate-nll}
\mathcal L(\boldsymbol{\xi}) = \mathbb E_{\bt\sim p}
\left[
\log\sigma_{\boldsymbol{\xi}}(\bt) +
\frac{\left\{y(\bt) - \mu_{\boldsymbol{\xi}}(\bt)\right\}^2}
{2\sigma_{\boldsymbol{\xi}}(\bt)^2}
\right].
\end{equation}
Up to the jitter, this assumes the log-normal family for the law of the surrogate prediction
\begin{equation}
\label{eq:surrogate-law}
P^\textrm{exo} + \epsilon \mid \bt \sim \operatorname{LogNormal}\left(\mu_{\boldsymbol{\xi}}(\bt), \sigma_{\boldsymbol{\xi}}(\bt)^2\right).
\end{equation}
A deterministic surrogate corresponds to the degenerate limit $\sigma_{\boldsymbol{\xi}} \to 0$; richer families are possible, for instance heavier-tailed location--scale distributions in log space, but \eqref{eq:surrogate-law} suffices for our purposes. Because the training targets are themselves MC estimates, the learnt scale absorbs both the network's approximation error and the MC noise in the targets. With $\Pi_{\boldsymbol{\xi}}$ specified, the downstream predictive law from \eqref{eq:predictive} is a posterior-weighted mixture of (potentially shifted) log-normals,
\begin{equation}
\label{eq:posterior-predictive}
\hat p(P^\textrm{exo} + \epsilon \mid \x_o) =
\frac{1}{k}\sum_{j=1}^k \Pi_{\boldsymbol{\xi}}(P^\textrm{exo}\mid \bt_j),
\end{equation}
where $\bt_1, \ldots, \bt_k \sim p(\cdot \mid \x)$ and whose quantiles define uncertainty-aware predictive intervals for the exotic price; in particular, upper and lower quantiles may be interpreted as model-implied ask and bid levels. Finally, note that if $\eps >0$, the shifted log-normal from~\eqref{eq:posterior-predictive} allows for negative prices, specifically prices in $[-\eps,0]$. More flexible conditional predictive families can be employed in future work to remove this numerical artefact. 

\section{A Novel XAI approach: Hellinger-SHAP}\label{sec:xai_hellinger}

Classical post-hoc explainability methods such as SHAP 
are usually applied to a point prediction. In the present setting, however, the
inferential object of interest is the posterior distribution over model
parameters conditional on an implied-volatility (IV) surface, rather than
only a single calibrated parameter vector. Even when point recovery is
accurate on average, the inverse map from numerically generated or
Monte Carlo noisy IV surfaces to model parameters can leave residual
parameter uncertainty under the specified prior-predictive model. For
nonlinear downstream exotic prices, collapsing this conditional uncertainty
to a point calibration can be consequential. Our goal is therefore to explain
posterior information gain---the reduction in prior uncertainty induced by
observing an IV surface---rather than merely a point estimate.

To identify the regions of the observed IV surface most associated with this
information gain, we apply Shapley values from cooperative game theory to a
scalar functional measuring posterior contraction from the prior. Let
$\mathcal S:=\{1,\ldots,d_x\}$ index the entries of an IV surface. For a scalar
coalition value function $v$ and feature $k\in\mathcal S$, the Shapley
attribution is
\[
\phi_k(v,\x_o)
=
\sum_{I\subseteq\mathcal S\setminus\{k\}}
\frac{|I|!(d_x-|I|-1)!}{d_x!}
\left\{
 v(I\cup\{k\};\x_o)-v(I;\x_o)
\right\}.
\]
Because exact evaluation is combinatorial in $d_x$, we approximate these
attributions using Kernel SHAP \citep{lundberg2017unified} with a local
empirical background distribution, defined next.

For each explained surface $\x_o$, let $B_{\x_o}$ denote the empirical
distribution supported on its $100$ nearest IV surfaces from the simulated
dataset, where nearness is measured with the Euclidean distance in the space of retained principal components. For a coalition $I\subseteq\mathcal S$ and a
background surface $\x_b$, write $(\x_o)_I,(\x_b)_{-I}$ for the hybrid surface
that retains the entries in $I$ from $\x_o$ and replaces the complementary
entries by those of $\x_b$. If $g$ is a scalar model output or
posterior-information functional, the local-background game is
\[
v_g(I;\x_o)
:=
\mathbb E_{\x_b\sim B_{\x_o}}
\left[
 g\!\left((\x_o)_I,(\x_b)_{-I}\right)
\right].
\]
Thus, Kernel SHAP attributes marginal changes in $g$ relative to nearby,
empirically generated reference surfaces. The use of the full simulated
dataset as the candidate pool provides a substantially denser set of local
backgrounds than selection from the held-out test set alone.

This construction reduces, but does not eliminate, the off-manifold issue
inherent in coalition-based explanations of structured surfaces. Although
both $\x_o$ and $\x_b$ are simulated IV surfaces, their feature-wise hybrid
need not itself lie exactly on the rough-Heston IV-surface manifold. The
resulting scores should consequently be interpreted as local relative
attributions under the stated background-completion distribution, rather than
as strictly manifold-preserving counterfactual effects.

The key modelling choice is the scalar functional $g$. For a parameter
coordinate $\theta^i$, let
\[
\hat r_i(\x,\theta^i)
:=
\frac{\hat p(\theta^i\mid\x)}{p(\theta^i)}
\]
denote the estimated marginal posterior-to-prior density ratio. We consider
two information-theoretic measures of posterior contraction,
\[
g_i^{\mathrm{KL}}(\x)
:=
D_{\mathrm{KL}}\!\left(
\hat p(\theta^i\mid\x)\,\middle\|\,p(\theta^i)
\right)
=
\mathbb E_{\theta^i\sim\hat p(\cdot\mid\x)}
\left[\log \hat r_i(\x,\theta^i)\right],
\]
and
\[
g_i^{H^2}(\x)
:=
H^2\!\left(
\hat p(\theta^i\mid\x),p(\theta^i)
\right)
=
1-
\mathbb E_{\theta^i\sim p}
\left[\sqrt{\hat r_i(\x,\theta^i)}\right],
\]
where we use the convention
\begin{equation*}
    H^2(p,q) = \frac{1}{2}\int \left(\sqrt{p(y)} - \sqrt{q(y)}\right)^2 \mathrm{d}y = 1 - \int \sqrt{p(y) q(y)} \mathrm{d}y
\end{equation*}
for the squared Hellinger distance. The corresponding Kernel-SHAP value
function is $v_{g_i^D}(I;\x_o)$, with $D$ equal to $\mathrm{KL}$ or $H^2$. Its
attributions identify the surface regions whose inclusion, relative to the
local background-completion distribution, most increases posterior
information about $\theta^i$.

The Kullback--Leibler and squared-Hellinger constructions are complementary
summaries of posterior contraction. In particular, the squared Hellinger
distance is bounded and therefore provides a useful sensitivity analysis to
the Kullback--Leibler version. Agreement between their attribution maps is an
empirical robustness finding, rather than a consequence of correct model
specification.

Two implementation points are worth noting. First, the full-posterior
estimator $\hat{p}(\bt \mid \x)$ described in Section~\ref{sec:sbi-landscape}
is not convenient for this analysis, as evaluating its marginals would require
numerically integrating over the remaining $m-1$ coordinates of $\bt$. We
therefore train $m$ neural ratio estimators, one for each marginal
posterior-to-prior ratio $\hat r_i$. Second, once the one-dimensional
$\hat r_i$'s are available, both $g_i^{\mathrm{KL}}$ and $g_i^{H^2}$ are
evaluated by Clenshaw--Curtis quadrature.

The same construction extends beyond one-dimensional marginals. For
example, when posterior dependence between two components is of interest,
one may define an information functional using
\[
D\!\left(
\hat p(\theta^i,\theta^j\mid\x)
\,\middle\|\,
p(\theta^i,\theta^j)
\right)
\]
and apply the same local-background Shapley decomposition. Thus, the
methodology can be used both for single-parameter recovery and for joint
posterior-information analysis.

Finally, the posterior-information functional, and hence its Shapley
decomposition, is prior- or regime-specific. A different prior changes both
the posterior contraction being measured and the appropriate local reference
distribution. The reported attributions should therefore be understood as
properties of the specified prior-predictive rough-Heston experiment, rather
than as prior-free statements about IV-surface informativeness.

\section{Uncertainty quantification results}\label{sec:simstudy} 
We next validate our uncertainty-quantification (UQ) 
methodology in a simulation study. To this end, for each parameter vector $\bt$, we apply the HQE scheme described in Section~\ref{subsec:hqe} to produce a set of discretized asset-volatility paths $\left(S_t^{(m)}, V_t^{(m)}\right)$, for $m=1,\ldots,M$. These paths are used to generate datasets for the two SBI tasks, specifically generate implied-volatility (IV) surfaces derived from European option prices for the inverse problem considered in Section~\ref{sec:simstudy_European_and_IV}, and prices for $J=3$ exotics contract classes: forward-start call, down-and-out put barrier and realized-variance (RV) call options, considered in Section~\ref{sec:simstudy-exotics}. Each exotic family is evaluated over several maturities and other contract specific terms, which are omitted for brevity until Section~\ref{sec:simstudy-exotics}. Formally, let
\begin{align*}
    C(T,K;\bt)
    =& \quad \ev[(S_T-K)^{+}],\\
    P_j^{\textrm{exo}}(\bt)
    =& \quad \ev[\Psi_j(S_\cdot,V_\cdot)],
\end{align*}
where $(T,K)$ are the maturity and strike of the European call, and $\Psi_j$ denote the potentially path-dependent exotic payoff functionals for $j=1,2,3$. 
We write $\widehat{C}_M(T,K;\bt)$ and $\widehat{P}_{j,M}^{\textrm{exo}}(\bt)$ for the corresponding Monte Carlo (MC) approximations obtained from the HQE scheme.
Schematically, for each value of $\bt$
\begin{equation*}
    \bt
    \;\longrightarrow\;
    \left\{\big(\widehat S_\cdot^{(m)},\widehat V_\cdot^{(m)}\big)\right\}_{m=1}^{M}
    \;\longrightarrow\;
    \begin{cases}
        \widehat C_M(T,K;\bt)
        \;\longrightarrow\;
        \widehat\sigma_{\textrm{IV}}(T,K;\bt)
        \;\longrightarrow\;
        \x(\bt),\\[0.6em]
        \widehat P_{j,M}^{\textrm{exo}}(\bt),
        \qquad j=1,2,3.
    \end{cases}
\end{equation*}
Here $\widehat\sigma_{\mathrm{IV}}(T,K;\bt)$ is the implied volatility obtained by inverting the corresponding MC European price, and $\x(\bt)$ collects these IVs over an observation grid of pairs $(T,K)$. Thus, the upper branch produces the observation used for the inverse SBI problem, in which we approximate $p(\bt \mid \x)$, whereas the lower branch produces quantities that are held out for downstream pricing applications. After an IV surface $\x_o$ has been observed and a posterior for $\bt$ obtained, these exotic pricing maps are used to propagate parameter uncertainty into posterior distributions for financially relevant, non-European claims. Given an observed IV surface $\x_o$, the posterior for $\bt$ can subsequently be propagated through the exotic pricing maps to quantify uncertainty in non-European claims. To simplify notation, we write the Monte Carlo approximation of an exotic price as $\widehat{P}^{\textrm{MC}}$, with the contract understood from context. In the following, Section~\ref{sec:simstudy_European_and_IV} describes the parameter design, construction and numerical validation of the IV dataset. Section~\ref{sec:sim_study_posterior_recovery} evaluates posterior recovery, while Section~\ref{sec:simstudy-exotics} studies posterior uncertainty propagation to exotic prices.

\subsection{Synthetic dataset generation for the inverse 
problem}\label{sec:simstudy_European_and_IV}

\subsubsection{Parameter and IV surface design}\label{sec:gridconst}

Recall the rough Heston parameterization $\bt = (\kappa, \theta, \nu, \rho, V_0, H) \in \Theta$, where we take $\Theta$ to be the Cartesian product of the support ranges listed in Table~\ref{tab:rheston_supports}, and let the simulation prior $p(\bt)$ be uniform on $\Theta$. As discussed in Section~\ref{sec:TRE_and_NBE}, this choice ensures that our neural posterior estimator $\bt,\x \mapsto \hat p(\bt\mid \x)$ is trained over a broad parameter region, while allowing alternative priors to be imposed at inference time.

\begin{table}[ht]
\centering
\caption{Parameter supports for the rough Heston model synthetic dataset.}
\label{tab:rheston_supports}
\begin{tabular}{lcc}
\toprule
\textbf{Parameter} & \textbf{Symbol} & \textbf{Support Range} \\ \midrule
Mean-reversion Speed & $\kappa$ & $(0.10, 2.00)$ \\
Long-run Variance & $\theta$ & $(0.01, 0.25)$ \\
Vol-of-Vol & $\nu$ & $(0.05, 0.45)$ \\
Leverage correlation & $\rho$ & $(-0.95, -0.1)$ \\
Initial Variance & $V_0$ & $(0.01, 0.25)$ \\
Hurst Exponent & $H$ & $(0.05, 0.45)$ \\ \bottomrule
\end{tabular}
\end{table}

Previous deep-calibration studies for stochastic-volatility models often work with a cone-shaped subset of the implied-volatility surface, excluding parts of the $(T,k)$ grid that are either too illiquid in practice or numerically unstable. Here we instead work with a standardized surface indexed by maturities $T$ and normalized moneyness coordinates. For each maturity $T$, we define the log-forward moneyness
\[
k=\log\!\left(\frac{K}{F_0(T)}\right),
\]
where $F_0(T)$ denotes the forward price for maturity $T$. Under our zero-rate specification, $F_0(T)=S_0$, but the forward notation is convenient and keeps the construction consistent with standard implied-volatility parametrizations. We then standardize moneyness by the at-the-money volatility scale,
\[
z=\frac{k}{\bar{\sigma}_{\mathrm{ATM}}(T)\sqrt{T}},
\]
where $\bar{\sigma}_{\mathrm{ATM}}(T)$ denotes the at-the-money implied volatility estimated from the same Monte Carlo sample paths used to compute all strikes at maturity $T$. Equivalently, if one fixes a standardized coordinate $z$, the corresponding log-moneyness is
\[
k(T,z)=z\,\bar{\sigma}_{\mathrm{ATM}}(T)\sqrt{T}.
\]

This normalization reduces scale heterogeneity across maturities and improves the conditioning of the inverse map $\bt \mapsto x(\bt)$ without increasing computational cost, since the ATM volatility is obtained from the same simulated option prices. We take
\[
z\in\{-2.5,-2.0,-1.5,-1.0,-0.5,0,0.5,1.0,1.5\}
\]
at maturities
\[
T\in\left\{\frac{5}{252},\frac{10}{252},\frac{1}{12},\frac{1}{4},\frac{1}{2},1,2\right\},
\]
which yields a $63$-dimensional implied-volatility surface.

Because a common set of simulated paths is used to construct entries of a
given surface, the Monte Carlo and inversion errors across strikes and
maturities are generally dependent.  This dependence is retained in the
simulated joint distribution of $\x$ on which the amortized estimators are
trained; we do not impose an independent-noise model on individual IV
entries.

The key advantage of this standardized representation is that each feature corresponds to a comparable location on the smile across parameter draws: near-the-money and wing observations remain informative even when the overall volatility level changes. In particular, this avoids allocating too many features to contracts that become numerically negligible in low-volatility regimes while preserving a fixed and interpretable observation layout for the neural estimators.

\subsubsection{IV inversion}
The standardized observation grid from Section~\ref{sec:simstudy_European_and_IV} is indexed by the nominal log-forward moneyness
$k=\log\!\left(\frac{K}{F_0(T)}\right),$
which reduces to $k=\log(K/S_0)$ under our zero-rate specification. However, when inverting Monte Carlo European prices into implied volatilities, we do not use a naive Monte Carlo average of terminal asset values as the forward input. Instead, the inversion is performed relative to the forward level induced by the same martingale-corrected conditional Black--Scholes representation used in the HQE pricing step.

 In particular, whenever one side of the smile is more stable in put rather than call representation, we recover the corresponding European price through the standard put--call parity relation using that same forward convention, and then invert implied volatility consistently throughout. This avoids a small inconsistency that would arise from mixing parity relations under the simulated forward with inversion formulas centered directly at $S_0$, and in practice reduces spurious inversion noise, especially for deep in- or out-of-the-money contracts.

The same inversion convention is used to obtain the at-the-money implied volatilities entering the standardized $z$-grid construction from Section~\ref{sec:gridconst}. In other words, the grid itself is indexed by the nominal coordinates $(T,z)$, but the numerical inversion producing the associated implied volatilities is carried out under the forward convention consistent with the martingale-corrected conditional pricing scheme.

\subsubsection{Numerical validation}
\label{sec:simstudy-IV_benchmark}

We validate the HQE generator by benchmarking the resulting European implied volatilities against a Lewis inversion based on the rough Heston characteristic function. For each parameter vector $\bt$ and grid point $(T,z)$, define
\begin{equation}
    \Delta_{\mathrm{IV}}^{\mathrm{HQE}}(T,z;\bt)
    =
    \left|
        \widehat{\sigma}^{\mathrm{HQE}}(T,z;\bt)
        -
        \sigma^{\mathrm{Lewis}}(T,z;\bt)
    \right|.
    \label{eq:HQE-Lewis-error}
\end{equation}
Table~\ref{tab:hqe_lewis_quantiles}
 reports quantiles of $\Delta_{\mathrm{IV}}$ across the simulation prior for different time-grid resolutions and grading rates.

\begin{table}[t]
    \centering
    \caption{Quantiles of the absolute difference between the HQE and Lewis implied-volatilities $\Delta_{\mathrm{IV}}^{\mathrm{HQE}}(T,z;\bt)$ defined in~\eqref{eq:HQE-Lewis-error} over all pairs $(T,z)$, for different time-grid resolutions $N \in \{504,1008,2016,4032\}$ and grading rates $r \in \{1,2,3\}$. We use MC estimators with $M= 5 \cdot 10^5$ samples; quantiles are estimated based on $500$ pairs $(\bt,\x)$. Non-uniform grids perform better, and we settle on $N=2016$ and $r=2$ for the dataset generation.}
    \label{tab:hqe_lewis_quantiles}
    \small
    \setlength{\tabcolsep}{3pt}
\begin{tabular}{c | @{\hspace{5pt}}
                cccc @{\hspace{8pt}}
                cccc @{\hspace{8pt}}
                cccc @{\hspace{8pt}}
                cccc}
        \toprule
        & \multicolumn{4}{c}{$N=504$}
        & \multicolumn{4}{c}{$N=1008$}
        & \multicolumn{4}{c}{$N=2016$}
        & \multicolumn{4}{c}{$N=4032$} \\
        \cmidrule(lr){2-5}
        \cmidrule(lr){6-9}
        \cmidrule(lr){10-13}
        \cmidrule(lr){14-17}
        $r$
        & $q_{50}$ & $q_{90}$ & $q_{95}$ & $q_{99}$
        & $q_{50}$ & $q_{90}$ & $q_{95}$ & $q_{99}$
        & $q_{50}$ & $q_{90}$ & $q_{95}$ & $q_{99}$
        & $q_{50}$ & $q_{90}$ & $q_{95}$ & $q_{99}$ \\
        \midrule
$1$ & 0.03 & 0.23 & 0.38 & 0.80 & 0.02 & 0.15 & 0.23 & 0.48 & 0.02 & 0.10 & 0.15 & 0.29 & 0.02 & 0.08 & 0.12 & 0.21 \\
$2$ & 0.02 & 0.10 & 0.14 & 0.24 & 0.02 & 0.08 & 0.11 & 0.19 & 0.02 & 0.07 & 0.10 & 0.15 & 0.01 & 0.07 & 0.09 & 0.15 \\
$3$ & 0.02 & 0.10 & 0.13 & 0.22 & 0.02 & 0.08 & 0.11 & 0.17 & 0.02 & 0.07 & 0.09 & 0.15 & 0.01 & 0.07 & 0.10 & 0.17 \\
        \bottomrule
    \end{tabular}
\end{table}

Two conclusions emerge clearly from these experiments. First, graded time grids substantially improve the approximation relative to the uniform case, and we therefore adopt grading rate $r=2$ in the final dataset generation. Second, once the QE correction is included, the remaining discrepancy with the Lewis benchmark is already of the same order as, or smaller than, the intrinsic Monte Carlo noise over much of the simulation prior. In particular, additional diagnostics reported in Appendix~\ref{app:sec_extended_validation_of_the_dataset} show that the main improvement comes from controlling truncation error rather than from refining the purely hybrid part of the scheme.

The Lewis benchmark itself is numerically delicate, especially in the short-maturity and wing regimes. In our implementation, benchmark values were stabilized using pair-specific Fourier-inversion settings together with Richardson extrapolation. The reported discrepancies should therefore be interpreted as total numerical disagreement, which includes Lewis pricing error, rather than as a pure HQE bias estimate. 

Nevertheless, the refinement diagnostics indicate that the HQE scheme achieves sub-$10$bp accuracy up to the $95$th percentile, while the remaining upper-tail discrepancy is small and appears to be dominated by benchmark instability and Monte Carlo noise rather than by the HQE discretization itself. This is more than sufficient for the synthetic-data generation task considered in this paper.

\subsubsection{Surface-quality checks}
\label{sec:simstudy-IV_no_arbitrage_etc}

As a further sanity check on the synthetic dataset, we verify that the generated implied-volatility surfaces satisfy the expected qualitative shape constraints, including convexity- and wing-behavior diagnostics in the spirit of static-arbitrage and Roger-Lee-type checks \citep{roger_lee_2004}. No generated surface violates any of these diagnostics. 

\subsection{Parameter calibration and posterior recovery}\label{sec:sim_study_posterior_recovery}

\begin{figure}[!htbp]
    \centering
    \includegraphics[width=\linewidth]{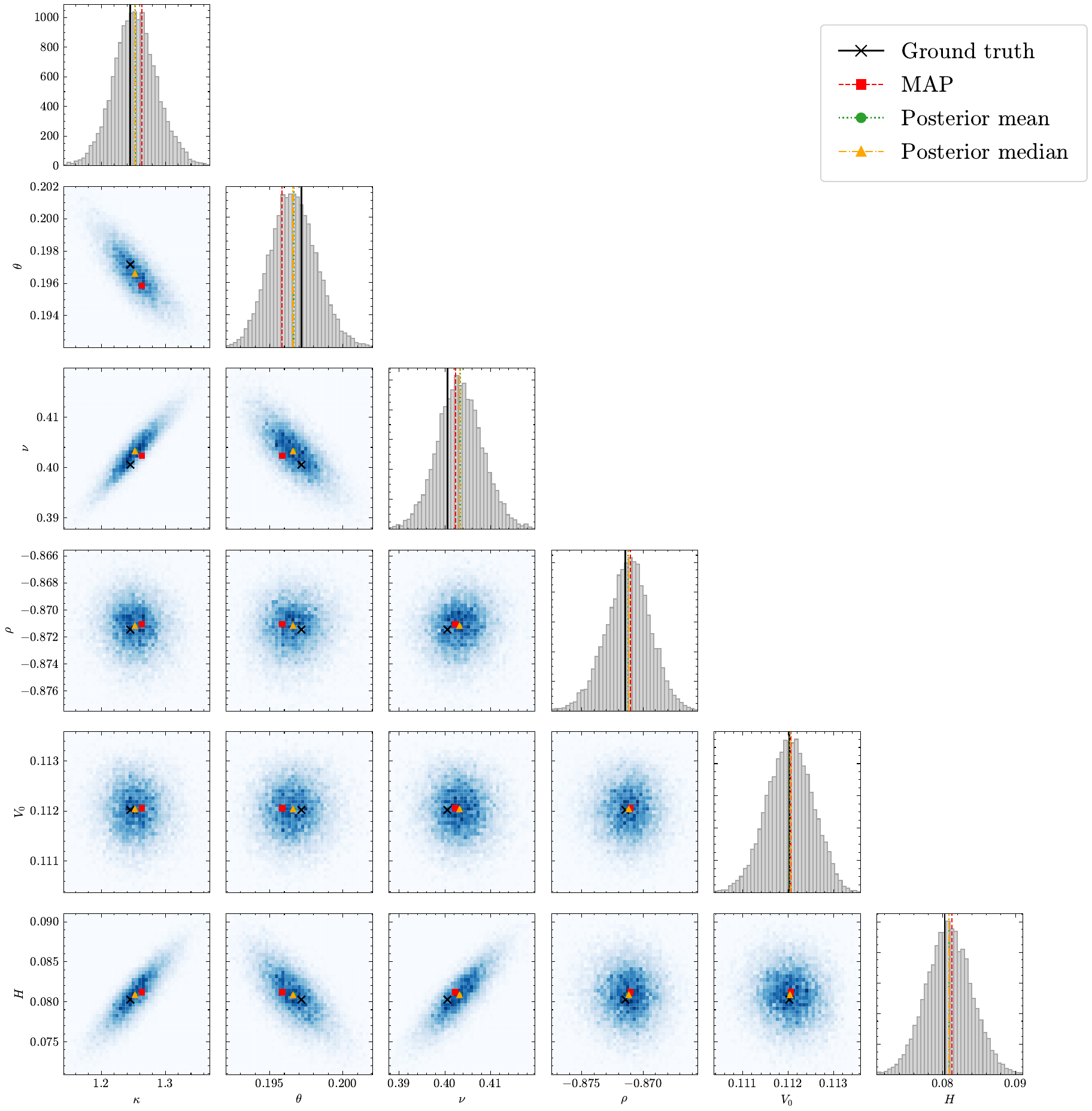}
    \caption{Posterior distribution of the parameter vector $\bt$ conditional on an IV surface $\x$ from the held-out test set. The diagonal panels display the estimated marginal posterior density of each parameter, while the off-diagonal panels show the corresponding pairwise marginal posterior distributions. For each parameter, the ground truth value is shown together with three point summaries of the posterior: the maximum a posteriori (MAP) estimate, the sample posterior mean and the sample posterior median. Parameter recovery is good, while the shape of the posterior distributions reflect the remaining parameter uncertainty and correlation. These plots are generated from $15000$ posterior draws.}
    \label{fig:posterior_recovery}
\end{figure}
\begin{figure}[!htbp]
    \centering
    \includegraphics[width=0.6\linewidth]{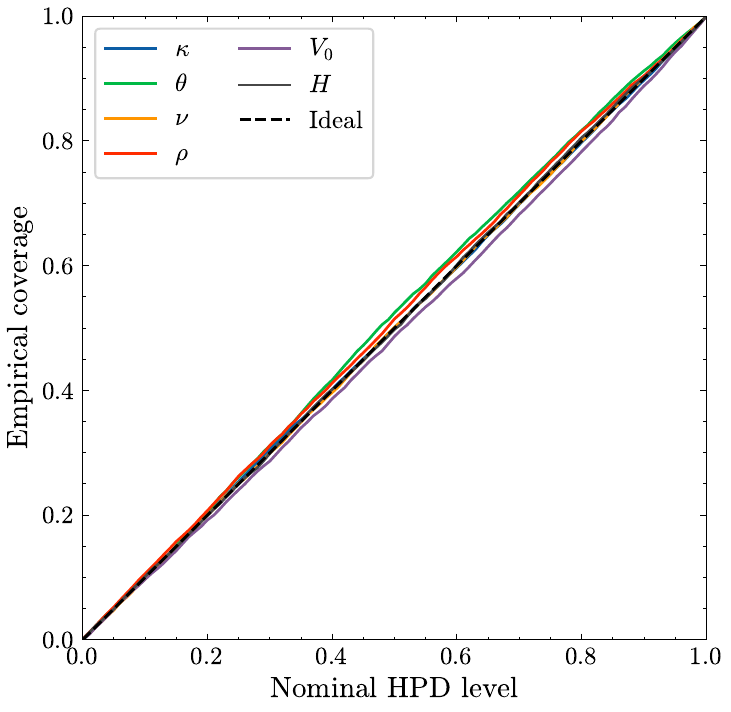}
    \caption{Empirical highest posterior density (HPD) regions coverage of the 
    learnt TRE conditional posteriors over the held-out test set. We note good coverage results, i.e., the empirical coverage of the HPD regions matches the nominal values almost exactly. This confirms the learnt approximations $\hat{p}(\kappa \mid \x),\ldots, \hat{p}(H \mid \x, \kappa, \theta,\nu,\rho, V_0)$ are neither over- nor under-confident. See Appendix~\ref{app:TRE_and_marginal_NRE}, in particular the second half of this section, for further details.}
    \label{fig:TRE_seq_coverage_beta}
\end{figure}
We generate $500000$ parameter-surface pairs $(\bt,\x)$ using the HQE scheme with $N=2016$ and grading parameter $r=2$. We randomly assign them to training, validation and test sets containing $90\%$, $5\%$ and $5\%$ of the pairs, respectively. Each IV surface $\x$ is $63$-dimensional, as explained in Section~\ref{sec:gridconst}. We train a neural posterior estimator, specifically a Telescoping ratio estimator (TRE) on these pairs to recover the parameters underlying an observed IV surface and quantify the associated uncertainty through posterior (credible) regions. 
To facilitate neural-network training, we centre the surfaces and project them onto their first $20$ principal components (PCs), with the PCA fitted on the training set only; these PCs explain more than $99.9999\%$ of the variability. The retained PC scores are then standardized to unit variance, again using training-set statistics. Higher-order PCs explain negligible variation and correspond to eigenvalues that are numerically close to zero, which makes the subsequent standardization of PC scores numerically unstable. For TRE, the parameter components of $\bt$ are also standardized to the common range $[-1,1]$ before being passed as input to the neural network. Further details on the preprocessing steps, post-training steps, hyperparameter search neural-network architectures are provided in the Appendix~\ref{app:further_training_details}. Figure~\ref{fig:posterior_recovery} illustrates posterior recovery for an arbitrary IV surface $\x$ from the test set. Beyond recovering the ground truth $\bt$, we also require the posterior uncertainty reported by TRE to be reliable, because it is precisely this uncertainty that is subsequently propagated to the prices of exotic derivatives in Section~\ref{sec:simstudy-exotics}. We therefore assess the \emph{calibration} of the approximate posterior $\bt, \x \mapsto \hat{p}(\bt \mid \x)$. Here, (posterior) calibration means that Bayesian credible regions have the expected frequentist coverage under repeated simulation from the prior. For example, a $95\%$ posterior credible region should contain the corresponding ground truth $\bt$ parameter in approximately $95\%$ of pairs $(\bt,\x)$ from the test set. Recall from Section~\ref{sec:sbi-landscape} that TRE constructs the posterior sequentially, and here, we approximate
$p(\kappa\mid\x)$, followed by $p(\theta\mid\x,\kappa)$ and up to $p(H\mid\x,\kappa,\theta,\nu,\rho,V_0)$.
We can therefore assess calibration separately for each of the above conditional posterior densities. Specifically, for each approximate conditional density, we compute the corresponding approximate highest posterior density (HPD) regions at a range of nominal levels 
and compare the empirical coverage of these HPD regions over the test set with the theoretical value
. Figure~\ref{fig:TRE_seq_coverage_beta} shows near-perfect calibration for each of the approximate conditional densities, as the displayed curves barely deviate from the line $y=x$; see Appendix~\ref{app:TRE_and_marginal_NRE} for further details.

\subsubsection{Point estimators} 
\begin{figure}[!htbp]
    \centering
    \includegraphics[width=\linewidth]{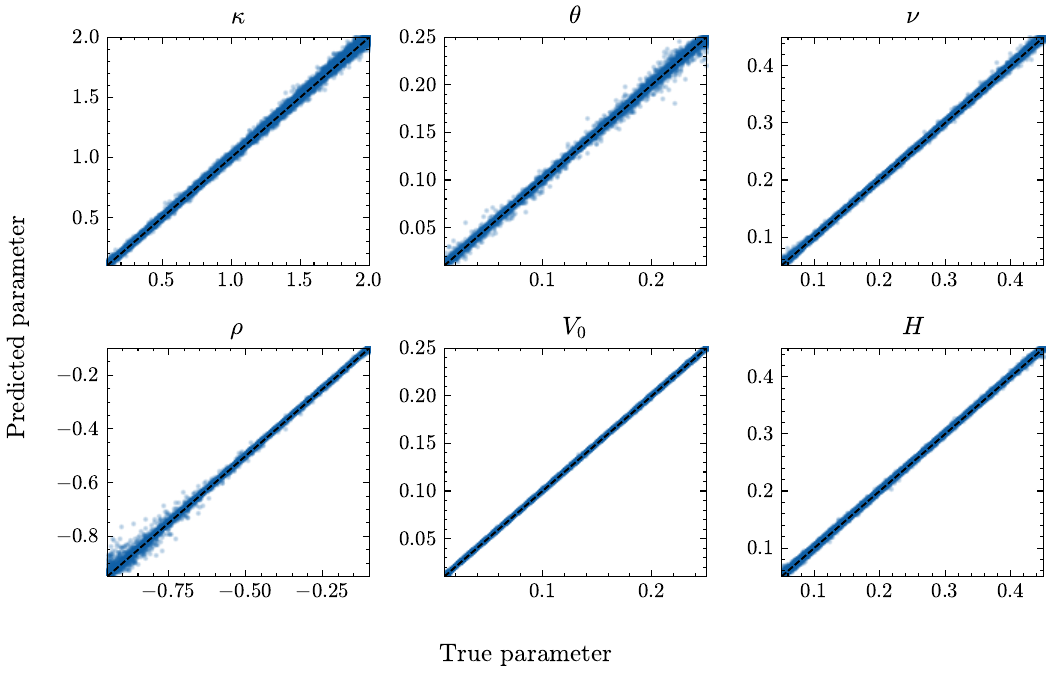}
    \caption{Neural point estimator trained with the MSE loss evaluated over the test set. Each subfigure illustrates the true versus predicted parameter values, with the $y=x$ line representing perfect parameter recovery.}
    \label{fig:NBE_MSE}

    \centering
    \includegraphics[width=\linewidth]{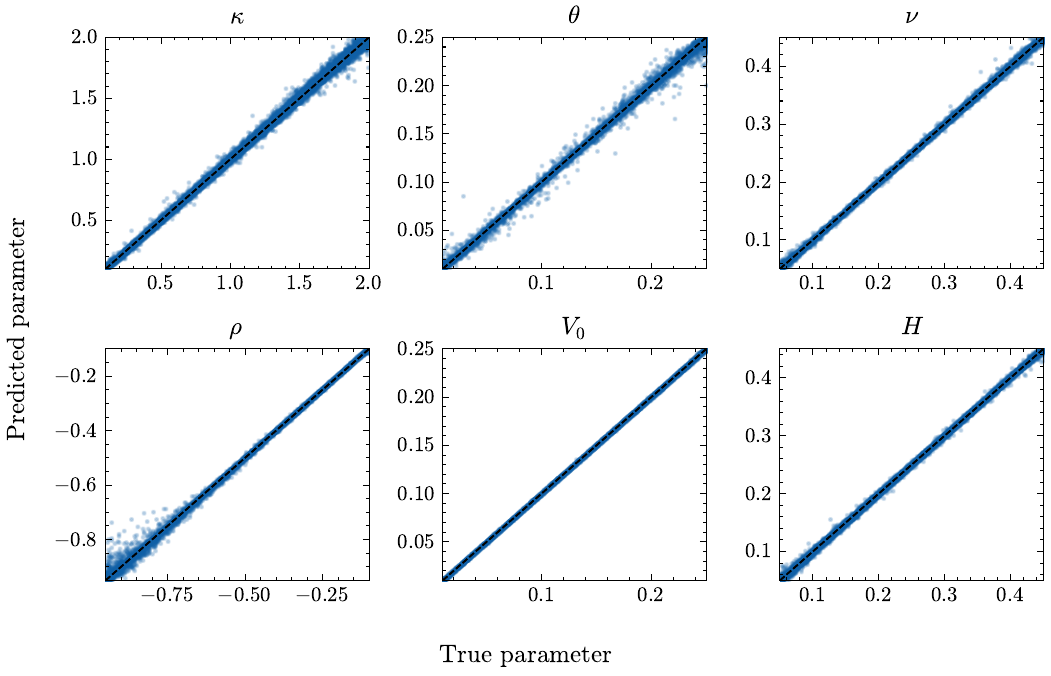}
    \caption{Neural point estimator trained with the MAE loss evaluated over the test set. Each subfigure illustrates the true versus predicted parameter values, with the $y=x$ line representing perfect parameter recovery.}
    \label{fig:NBE_MAE}
\end{figure}

For comparison with the TRE, using the same dataset and preprocessing steps, we also train neural point estimators with the MSE and MAE loss functions. As discussed in Remark~\ref{rem:grid}, these target the posterior mean and median, respectively. Figures~\ref{fig:NBE_MSE} and~\ref{fig:NBE_MAE} illustrate excellent performance of the trained neural estimators; training details are provided in Appendix~\ref{app:further_training_details}. 

\subsection{Pricing and uncertainty quantification for the forward exotic problem}
\label{sec:simstudy-exotics}

\subsubsection{Exotic contract description}\label{sec:exotic_contract_description}
As mentioned at the beginning of Section~\ref{sec:simstudy}, we consider three families of exotic contracts. First, the forward-start call option has payoff
\begin{equation*}
    \Psi^{\textrm{fwd}}(S_\cdot)
    =
    \left(S_{T_2}-e^k S_{T_1}\right)^{+},
\end{equation*}
with $\tau = T_2 - T_1$ and where $k\in\{-0.10,-0.05,0,0.05,0.10\}$, $T_1\in\left\{\frac{1}{12},\frac14,\frac12,1\right\}$ and $\tau\in\left\{\frac{1}{12},\frac14\right\}$. Second, consider the down-and-out put barrier option with payoff
\begin{equation*}
    \Psi^{\text{b}}(S_\cdot)
    =
    (K-S_T)^{+} \cdot
    \mathbf{1}\left(\inf_{0\leq t\leq T}S_t>B\right),
\end{equation*}
with $K=S_0e^{k_K}$ and $B=S_0e^{k_B}$, and where
$k_K\in\{0,-0.05\}$, $k_B\in\{-0.10,-0.15,-0.20\}$, and $T\in\left\{\frac{1}{4},\frac{1}{2},1\right\}$. The third and final contract is the realized-variance (RV) call option, with payoff given by
\begin{equation*}
    \Psi^{\text{RV}}(V_\cdot)
    =
    \left(
        \frac{1}{T}\int_0^T V_t\,\mathrm{d}t-K_{\mathrm{RV}}
    \right)^{+},
\end{equation*}
with $K_{\mathrm{RV}}=c\cdot v_{\mathrm{ref}}$ and where $v_{\mathrm{ref}}=0.13$, $c\in\{0.25,0.45,0.65\}$
, and $T\in\left\{\frac{1}{12},\frac14,\frac12,1\right\}$. In all three cases, we set $S_0 = 100$.

\subsubsection{Variance analysis of Monte Carlo exotic-price targets}
\label{sec:exo_mc_variance}

Before studying posterior uncertainty propagation through the surrogate pricers, we first assess the intrinsic Monte Carlo (MC) variability of the exotic-price targets themselves. This step is important because the surrogate models of Section~\ref{sec:sbi-surrogate} are trained on MC prices rather than exact prices, so excessive target noise would mechanically inflate the predictive uncertainty learned by the network and blur the distinction between parameter uncertainty and surrogate approximation error.

\begin{figure}[!htbp]
   \centering

    \begin{subfigure}[t]{0.32\linewidth}
        \centering
        \includegraphics[width=\linewidth]{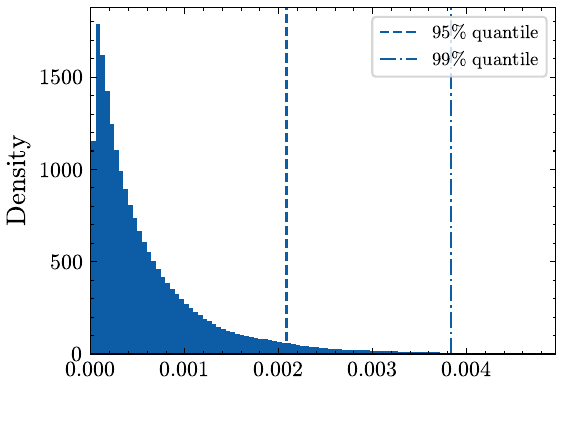}
        \caption{Forward-start}
    \end{subfigure}
    \hfill
    \begin{subfigure}[t]{0.32\linewidth}
        \centering
        \includegraphics[width=\linewidth]{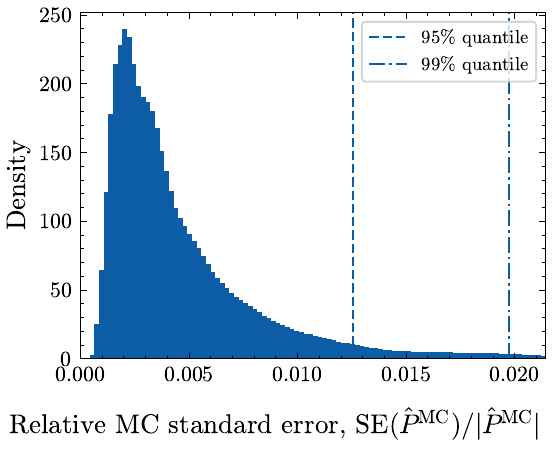}
        \caption{Barrier}
    \end{subfigure}
    \hfill
    \begin{subfigure}[t]{0.32\linewidth}
        \centering
        \includegraphics[width=\linewidth]{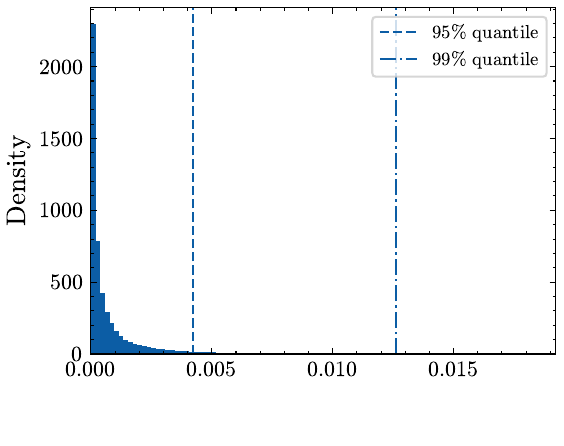}
        \caption{Realized variance}
    \end{subfigure}
    \caption{Empirical distributions of relative Monte Carlo standard errors for the three exotic contract families. The relative standard error is defined as the estimated Monte Carlo standard error divided by the corresponding estimated option price. For realized-variance options, contracts with estimated prices below $10^{-5}$ are excluded from this relative diagnostic to avoid instability from near-degenerate prices that are not informative. The results are pooled over all retained contracts within each family.}    \label{fig:mc_relative_se}
\end{figure}

For each exotic contract family, we report the quantiles of the relative MC standard error (also known as coefficient of variation). 
Relative diagnostics are informative for forward-start and barrier contracts, whose prices remain comfortably away from $0$ over the prior parameter region from Table~\ref{tab:rheston_supports}. For realized-variance (RV) calls, however, some of the most extreme upper-strike contracts become degenerate, i.e., none of the $500000$ MC estimator's paths yield non-zero contributions. Thus, for the RV family, we exclude from any subsequent analysis the contracts $(c,T) \in \{(0.65,1/12),(0.65,1/4)\}.$ Further, to keep relative summaries meaningful and focus on economically-relevant prices, when reporting results evaluated on the test-set, we exclude observations with estimated MC prices below $10^{-5}$. 
Detailed contract-by-contract diagnostics, including lower quantiles of prices, fractions of zero or near-zero realizations, and coefficient-of-variation summaries, are reported in Appendix~\ref{app:extended_validation_of_the_exotic_dataset}.

The main conclusion is that the forward-start and barrier training targets are sufficiently well resolved by MC for the purposes of surrogate learning. Forward-start options exhibit the smallest relative MC noise overall, while barriers are noisier, as expected from their path dependence and discontinuous payoff structure, but still remain within an acceptable range. The RV family is more delicate: lower and intermediate strike multipliers display stable coefficients of variation across the parameter space, whereas the two excluded contracts frequently concentrate MC prices at or near the numerical floor and do not provide useful regression targets over the broad prior considered here. For the retained contracts, the observed MC variability is small enough that the uncertainty decompositions reported in the next subsection may be interpreted as reflecting mainly posterior and surrogate effects rather than raw simulation noise.

\subsubsection{Predictive intervals and decomposition of uncertainty}
\label{sec:exo_predictive_intervals}

We now assess whether the posterior uncertainty recovered from the IV surface is economically material after propagation to exotic prices, and whether the 
resulting predictive distribution is calibrated. Here, predictive calibration refers to whether predictive intervals with a given nominal coverage level contain the corresponding held-out estimated price $P^{\mathrm{MC}}$ with approximately that frequency across the test set. 
Importantly, we distinguish between posterior-only predictive intervals, which propagate posterior uncertainty through the surrogate pricer, and full predictive intervals, which also account for the surrogate's conditional predictive uncertainty. 

Our analysis has four complementary parts. First, we assess the
downstream error incurred by 
using a neural point
estimator $\hat{\bt}$. Second, we quantify the economic size of the resulting full
posterior-predictive intervals relative to the exotic price. Third, we
compare the empirical coverage of posterior-only and full predictive
intervals. Finally, we use the law of total variance to determine how the
predictive uncertainty is divided between residual parameter uncertainty
and conditional surrogate-pricing uncertainty.

\begin{figure}[!htbp]
    \centering

    \begin{subfigure}[t]{0.92\linewidth}
        \centering
        \includegraphics[width=\linewidth]{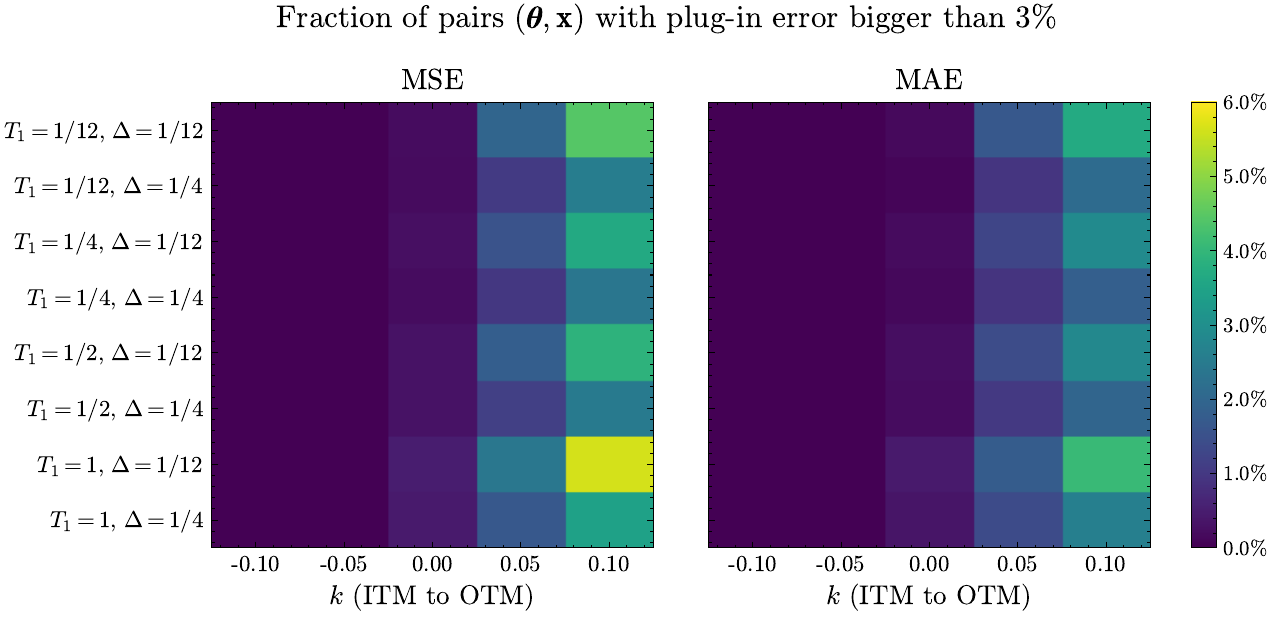}
        \caption{Forward-start}
    \end{subfigure}

    \vspace{0.4em}

    \begin{subfigure}[t]{0.92\linewidth}
        \centering
        \includegraphics[width=\linewidth]{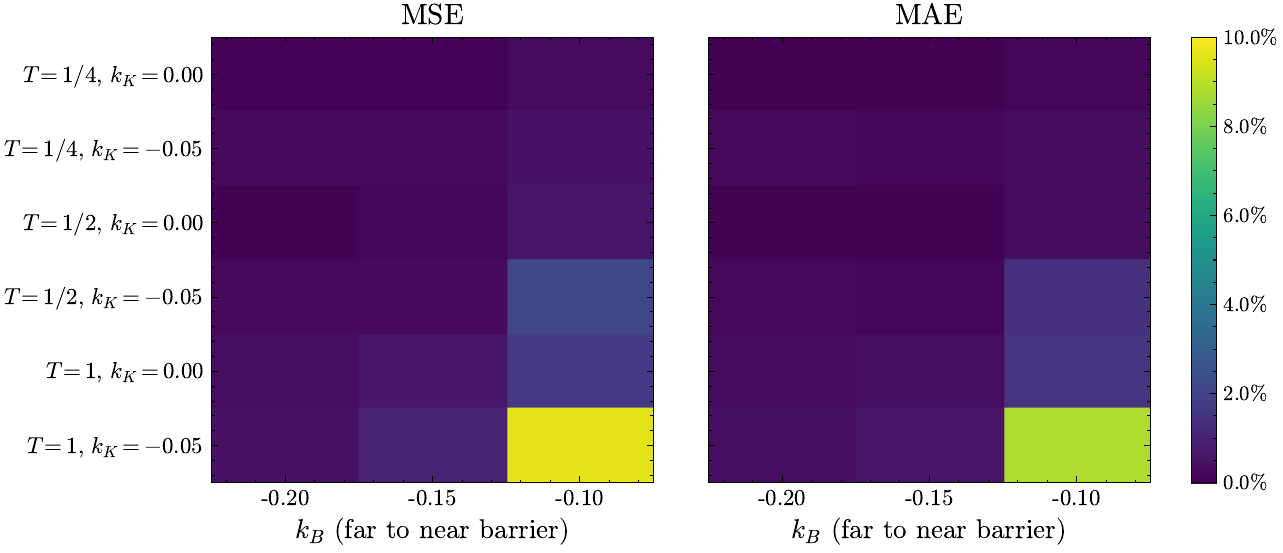}
        \caption{Barrier}
    \end{subfigure}

    \vspace{0.4em}

    \begin{subfigure}[t]{0.92\linewidth}
        \centering
        \includegraphics[width=\linewidth]{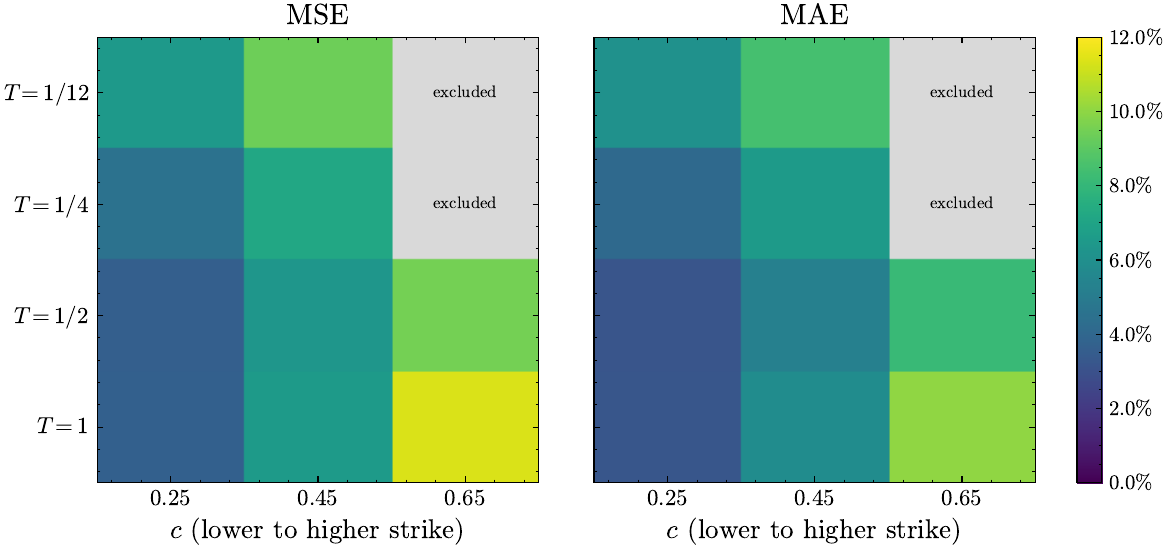}
        \caption{Realized variance}
    \end{subfigure}

\caption{
{\small{Frequency ($\%$) of test samples for which the relative plug-in
pricing error exceeds $3\%$, by exotic-contract specification. The plug-in
price evaluates the exotic surrogate at the MSE- or MAE-trained neural point
estimator, and the error is measured relative to the held-out MC reference price. Within each family, the left and right panels
show rates for the MSE- and MAE-trained estimators, respectively. Grey cells denote
excluded near-degenerate realized-variance contracts. Results for the $1\%$ and $5\%$ thresholds are reported in
Appendix~\ref{app_Extended_UQ}.}}}\label{fig:plugin_error_3pct}
\end{figure}
\textbf{Plug-in pricing error.} Recall the MC standard-error diagnostics from Figure~\ref{fig:mc_relative_se}, which
show that the sampling noise of these reference prices is small relative to the downstream effects reported below. The $99\%$ quantile of the relative MC standard error remains below $0.5\%$ 
for forward starts, is approximately $1.3\%$ for realized-variance claims, and is approximately $2\%$ for barriers. These values are smaller than, or of a lower order than, the contract-level plug-in discrepancies documented
below. Thus, although MC noise is necessarily present in the surrogate training targets and reference prices, it cannot by itself account for the observed downstream discrepancies.

For a test pair $(\bt,\x)$, let $\widehat\theta_{\mathrm{MSE}}(\x)$ and $\widehat\theta_{\mathrm{MAE}}(\x)$ denote the point calibrations produced by the networks trained under the MSE and MAE losses, respectively. For each contract specification, we compute the relative plug-in error
\begin{equation}
  E_{\mathrm{plug}}(\widehat\bt)
  :=
  \frac{\left|\widehat P^{\mathrm{MC}}-
  \widehat P^{\mathrm{exo}}(\widehat\bt)\right|}
  {\widehat P^{\mathrm{MC}}},
\end{equation}
and report the proportions for which $E_{\mathrm{plug}}(\widehat\bt)>q$ at $q\in\{0.01,0.03,0.05\}$.  This comparison isolates the practical limitation of replacing a posterior by a single parameter vector.  It should not be interpreted as a pure measure of inverse-network error: it also reflects the sensitivity of the exotic-pricing map and any residual surrogate approximation error.

\begin{table}[t]
\caption{Percentage of held-out test pairs for which the relative plug-in pricing error exceeds the stated threshold, pooled over the retained contract within each exotic family. The plug-in price is obtained by evaluating the exotic surrogate at the MSE- or MAE-trained neural point calibration, and the error is measured relative to the held-out Monte Carlo reference price. Contract-level patterns are reported in
Appendix~\ref{app_Extended_UQ}.
}\label{tab:plugin_summary}
\begin{tabular}{lrrrrrr}
\toprule
 & \multicolumn{2}{r}{Error > 1\%} & \multicolumn{2}{r}{Error > 3\%} & \multicolumn{2}{r}{Error > 5\%} \\
Family & MSE & MAE & MSE & MAE & MSE & MAE \\
\midrule
Forward-start & 6.46 & 4.86 & 1.08 & 0.85 & 0.46 & 0.40 \\
Barrier & 17.25 & 14.87 & 1.03 & 0.85 & 0.15 & 0.11 \\
RV & 22.05 & 18.38 & 6.83 & 6.05 & 4.21 & 3.81 \\
\bottomrule
\end{tabular}
\end{table}

Table~\ref{tab:plugin_summary} shows that visually accurate parameter recovery does not guarantee reliable plug-in exotic prices. For example, at the $3\%$ threshold, the fraction of failures appears moderate for forward starts and barriers, however Figure~\ref{fig:plugin_error_3pct} shows substantial heterogeneity within the same family. In particular, the high error rates are clustered among out-of-the-money contracts for forward-starts, e.g. $k=0.10$, and similarly for higher $k_B =- 0.10$ at longer maturities for barriers. The corresponding error rates are much larger for realized-variance claims for all contract specifications.
After pooling over contract specifications, the MSE and MAE point estimators exceed the $1\%$ error threshold in $6.46\%$ and $4.86\%$ of forward-start cases, $17.25\%$ and $14.87\%$ of barrier cases, and $22.05\%$ and $18.38\%$ of realized-variance cases, respectively. The corresponding rates remain non-zero at the $5\%$ threshold, especially for realized-variance contracts. Figures~\ref{fig:plugin_error_1pct} and~\ref{fig:plugin_error_5pct} in Appendix~\ref{app_Extended_UQ} give the contract-level patterns at the two additional $1\%$ and $5\%$ thresholds.

As discussed above, the errors are strongly contract-dependent. They are most prevalent for out-of-the-money forward starts, for barriers close to the prevailing level, and for the higher-strike realized-variance claims retained in the experiment.  The MAE estimator is slightly better in these aggregate summaries, but the difference between the two point estimators is secondary to the central finding: neither point summary provides an interval or a diagnostic of when its associated plug-in price is unreliable. In particular, the larger plug-in discrepancies relative to the MC standard errors establish that the effects documented here are not merely the direct transmission of sampling noise in the training prices.  They arise from collapsing a non-degenerate inverse distribution to one parameter vector, together with the sensitivity of the downstream pricing map.

\textbf{Relative widths of full predictive intervals.} To assess the economic size of the uncertainty-aware quotes, let $L_{ij}^{(\alpha)}$ and $U_{ij}^{(\alpha)}$ be the lower and upper endpoints of the nominal $\alpha$ predictive interval for test case $i$ and contract specification $j$.  We measure its scale relative to the held-out MC price by
\begin{equation}
  W_{ij}^{(\alpha)}
  :=
  100\,
  \frac{U_{ij}^{(\alpha)}-L_{ij}^{(\alpha)}}
  {\widehat P_{ij}^{\mathrm{MC}}},
  \qquad \alpha \in (0,1).
  \label{eq:relative-pi-width}
\end{equation}
The full predictive interval is used here, so that $W_{ij}^{(\alpha)}$ incorporates both posterior uncertainty and conditional surrogate uncertainty.  We summarize the empirical distribution of $W_{ij}^{(0.95)}$ by its median and upper quantiles.

Table~\ref{tab:pi-width-summary} shows that the $95\%$ predictive
intervals are economically non-negligible even after pooling across
heterogeneous contract specifications.  Their median width is $0.65\%$
of the reference price for forward starts, $1.66\%$ for realized-variance
claims, and $2.20\%$ for barriers.  The upper tail is substantially wider:
the $95\%$ quantile reaches $3.84\%$, $18.18\%$, and $5.82\%$ of the
respective reference price.  These pooled summaries conceal meaningful
heterogeneity.  The contract-level results in
Tables~\ref{tab:forward-width}, \ref{tab:barrier-width}, and~\ref{tab:rv-width}
show that some comparatively insensitive claims have narrow relative
intervals, whereas out-of-the-money forward starts, near-barrier contracts,
and several higher-strike realized-variance claims have particularly wide
bands. Hence uncertainty quantification is not 
equally valuable across contracts, but
it identifies precisely those regimes in which a point quote is least reliable.
\begin{table}[t]
\caption{Distribution of the relative width $W_{ij}^{(0.95)}$ of the full
nominal-$95\%$ predictive interval, pooled over all retained contract--test
pairs within each exotic family. For each pair, $W_{ij}^{(0.95)}$ is defined
in \eqref{eq:relative-pi-width} as the 
relative interval length expressed as a
percentage of the held-out Monte Carlo reference price. Entries are the
median, $90\%$ quantile, and $95\%$ quantile of the resulting distribution of
relative widths.  Contract-level results are reported in
Tables~\ref{tab:forward-width}--\ref{tab:rv-width}.}\label{tab:pi-width-summary}
    \begin{tabular}{lrrrrrr}
\toprule
Family & Median (\%)
& $90\%$ quantile (\%)
& $95\%$ quantile (\%)
\\
\midrule
Forward-start & 0.65 & 2.46 & 3.84 \\
Barrier & 2.20 & 4.56 & 5.82 \\
Realized variance & 1.66 & 8.17 & 18.18 \\
\bottomrule
\end{tabular}
\end{table}

\begin{table}[t]
\caption{Median and interquartile range (IQR) of empirical coverage rates for
nominal-$95\%$ predictive intervals across retained contract specifications
within each exotic family.  ``Posterior only'' (PO) propagates draws from the
post-training-calibrated TRE posterior through the conditional surrogate mean.
``Full UQ'' additionally samples from the conditional surrogate-predictive
law and therefore incorporates both posterior and conditional surrogate
uncertainty.  All entries are percentages; IQRs are reported in percentage
points.  Contract-level results are reported in
Tables~\ref{tab:forward-coverage}--\ref{tab:rv-coverage}.
}\label{tab:coverage_summary}
    \begin{tabular}{lrrrr}
\toprule
Family & PO $95\%$ median & PO $95\%$ IQR & Full UQ $95\%$ median & Full UQ $95\%$ IQR \\
\midrule
Forward-start & 94.60 & 1.58 & 97.19 & 0.33 \\
Barrier & 78.83 & 15.59 & 95.75 & 0.90 \\
Realized variance & 94.24 & 1.66 & 96.89 & 0.43 \\
\bottomrule
\end{tabular}
\end{table}
\textbf{Predictive coverage.} We next empirically evaluate predictive coverage. We remind the reader that we distinguish the posterior-only interval, obtained by propagating posterior samples through the conditional surrogate mean, from the full interval, obtained from the posterior mixture of conditional log-normal surrogate distributions in~\eqref{eq:posterior-predictive}.

The distinction is consequential.  As summarized in
Table~\ref{tab:coverage_summary}, posterior-only intervals materially
under-cover barrier claims: their median contract-level $95\%$ coverage is
$78.83\%$, with an interquartile range of $15.59$ percentage points.
Posterior-only coverage is slightly closer to nominal for forward starts and
realized-variance claims, but remains contract-dependent; see
Tables~\ref{tab:forward-coverage}, \ref{tab:barrier-coverage}, and~
\ref{tab:rv-coverage}.  Once conditional surrogate uncertainty is included,
the median coverage is restored to $97.19\%$ for forward starts, $95.75\%$
for barriers, and $96.89\%$ for realized-variance claims.  The corresponding
interquartile ranges, reported in Table~\ref{tab:coverage_summary}, are
small, indicating that full predictive calibration is substantially more
stable across retained contract specifications.

Accordingly, the evidence supports the calibration of the full predictive
construction, not a claim that posterior uncertainty alone is always
sufficient.  The modest over-coverage observed for some specifications is
consistent with conservative predictive intervals and finite-sample
estimation.  Residual departures from nominal coverage may arise because the
conditional surrogate law is approximated by a log-normal distribution whose
shape need not exactly match the parameter-dependent distribution of
surrogate and MC errors.  A richer conditional family or post-training
calibration of the exotic predictive distribution could be used if sharper
nominal calibration were required; see~\cite{salnikov2026} for such an application to neural estimators.

\textbf{Predictive-variance attribution.}
\begin{table}[t]
\caption{Median and interquartile range (IQR), across retained contract
specifications, of the case-wise shares of full predictive variance attributed
to TRE posterior uncertainty and conditional surrogate uncertainty.  For each
held-out test pair, the shares are computed from the law of total variance in
\eqref{eq:total-variance-decomposition} and then averaged within its contract
specification.  All entries are percentages; IQRs are reported in percentage
points.  Contract-level results are reported in Appendix~\ref{app_Extended_UQ}.}
\begin{tabular}{lrrrr}
\toprule
Family & TRE median & TRE IQR & Surrogate median & Surrogate IQR \\
\midrule
Forward-start & 81.67 & 8.01 & 18.33 & 8.01 \\
Barrier & 43.07 & 18.40 & 56.93 & 18.40 \\
Realized variance & 89.69 & 3.66 & 10.31 & 3.66 \\
\bottomrule
\end{tabular}
\label{tab:var_decomposition_summary}
\end{table}
Finally, we quantify the relative contributions of inverse and surrogate uncertainty.  Let $P$ denote the full predictive exotic price and write
\begin{equation}
  m_\xi(\bt):=\mathbb E[P\mid\bt],
  \qquad
  s_\xi^2(\bt):=\textrm{Var}(P\mid\bt),
\end{equation}
where, under the shifted log-normal surrogate 
in~\eqref{eq:surrogate-law}, both quantities are available in closed form.  Conditional on the observed surface $x$, the law of total variance gives
\begin{equation}
 \operatorname{Var}(P\mid \x)
 =
 \underbrace{\textrm{Var}_{\bt\mid \x}\left[m_\xi(\bt)\right]}
 _{\text{posterior (TRE) component}}
 +
 \underbrace{\mathbb E_{\bt\mid \x}\left[s_\xi^2(\bt)\right]}
 _{\text{conditional surrogate component}}.
 \label{eq:total-variance-decomposition}
\end{equation}
For each retained test pair and contract specification, we express the two terms in~\eqref{eq:total-variance-decomposition} as shares of the total predictive variance.  The reported contract-level values from Table~\ref{tab:var_decomposition_summary} average these case-wise shares over the held-out test set, thus avoiding a raw-variance aggregation across contracts with different price scales.

The decomposition is clearly contract-specific.  For realized-variance claims, the posterior component accounts for approximately $86\%$--$92\%$ of total predictive variance across the displayed specifications (Table~\ref{fig:rv_variance_decomposition}); residual parameter uncertainty is therefore the leading source of pricing uncertainty.  For barriers, by contrast, the conditional surrogate component is often comparable to or larger than the posterior component, and reaches more than $80\%$ for some near-barrier, longer-maturity claims (Table~\ref{fig:barrier_variance_decomposition}). This is consistent with the greater difficulty of approximating a discontinuous path-dependent payoff.  Forward starts predominantly exhibit posterior-driven uncertainty away from the most in-the-money specifications, but their surrogate share can still be material; see Table~\ref{fig:forward_start_variance_decomposition}.  These results reinforce the need to propagate both layers: omitting surrogate uncertainty is especially costly for barriers, whereas treating the surrogate layer as the only uncertainty source would miss the dominant contribution for most realized-variance claims.

Taken together, the plug-in, width, coverage, and variance-decomposition diagnostics show that UQ is not merely a cosmetic enlargement of a neural point estimate.  A point calibration can be highly accurate on average while still generate economically consequential errors for selected downstream claims. The full posterior-predictive construction flags these regimes through wider intervals and 
attains well-calibrated or conservative coverage by retaining uncertainty from the surrogate-pricing stage.

\section{XAI results}
\label{sec:simstudy-xai}

\begin{figure}[!htbp]
    \centering

    \begin{subfigure}[t]{0.48\linewidth}
        \centering
        \includegraphics[width=\linewidth]{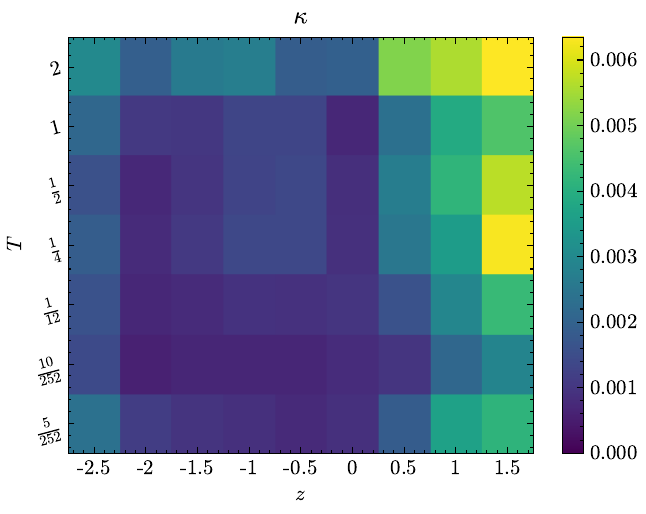}
    \end{subfigure}
    \hfill
    \begin{subfigure}[t]{0.48\linewidth}
        \centering
        \includegraphics[width=\linewidth]{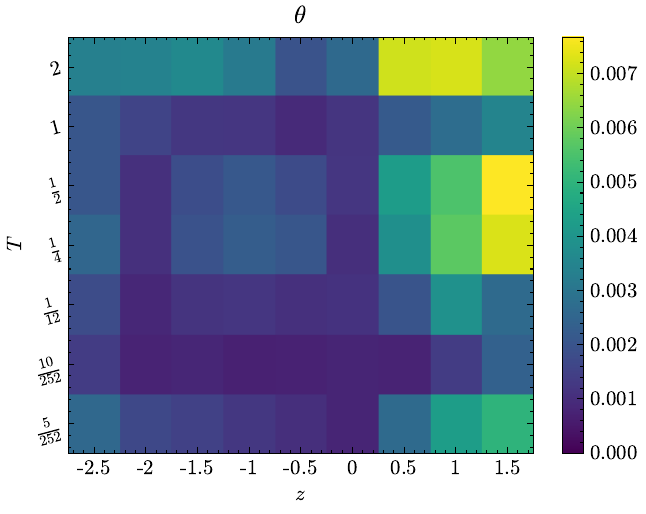}
    \end{subfigure}

    \vspace{0.5em}

    \begin{subfigure}[t]{0.48\linewidth}
        \centering
        \includegraphics[width=\linewidth]{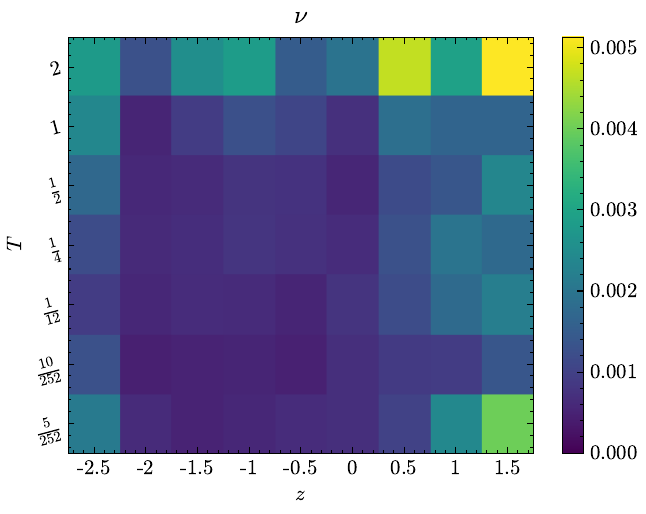}
    \end{subfigure}
    \hfill
    \begin{subfigure}[t]{0.48\linewidth}
        \centering
        \includegraphics[width=\linewidth]{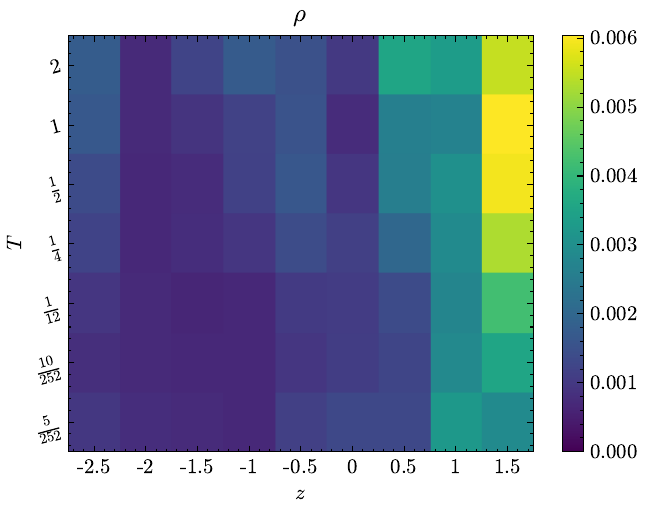}
    \end{subfigure}

    \vspace{0.5em}

    \begin{subfigure}[t]{0.48\linewidth}
        \centering
        \includegraphics[width=\linewidth]{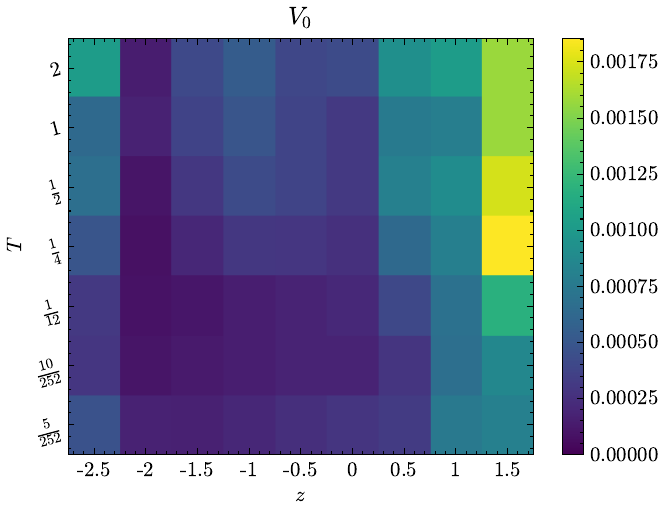}
    \end{subfigure}
    \hfill
    \begin{subfigure}[t]{0.48\linewidth}
        \centering
        \includegraphics[width=\linewidth]{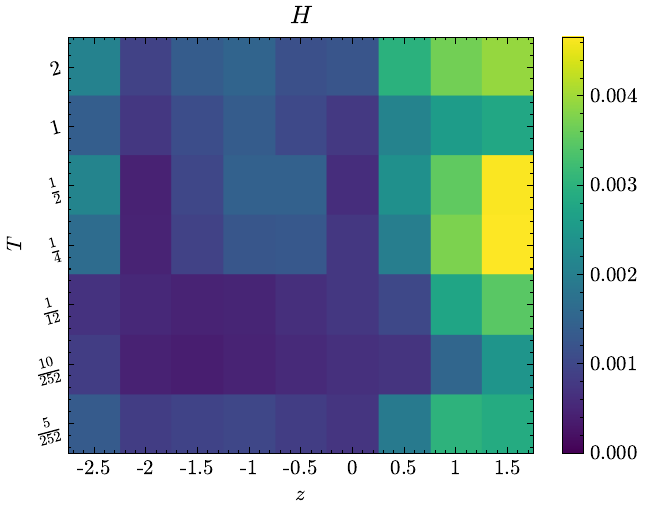}
    \end{subfigure}

    \caption{Mean absolute squared-Hellinger Kernel-SHAP attributions for the six rough-Heston parameters, averaged over $500$ held-out IV surfaces from the test set. Each panel displays the average magnitude of the local attribution at a maturity--standardized-moneyness location; it does not display a signed effect.}
    \label{fig:Hellinger_XAI_heatmap}
\end{figure}

This section applies the local-background Hellinger--SHAP construction from
Section~\ref{sec:xai_hellinger} to identify the regions of the IV surface
most strongly associated with marginal posterior contraction under the
specified prior-predictive rough-Heston experiment. Figure~\ref{fig:Hellinger_XAI_heatmap}
reports mean absolute squared-Hellinger Kernel-SHAP attributions, averaged
over $500$ held-out IV surfaces. Thus, each panel summarizes the average
magnitude of a feature's local contribution to posterior information gain about
the corresponding parameter. Since the attributions are taken in absolute
value before averaging, the maps describe strength rather than direction:
they do not show whether changing a particular IV entry increases or
decreases posterior contraction for an individual surface. We also remind the reader that the results
correspond to a surface constructed in the standardized coordinate $z$ 
from Section~\ref{sec:gridconst}, and not to a raw-strike wing construction.

A common pattern across the panels of Figure~\ref{fig:Hellinger_XAI_heatmap} is the prominence of the positive
standardized-moneyness wing, especially at medium and long maturities. This
pattern is particularly visible for $\kappa$, $\theta$, $\rho$, $V_0$, and
$H$, and is also present, though more localized, for $\nu$. Further, we stress that these results do not imply that one
right-wing quote identifies all parameters on its own. Rather, the finding is that these entries
recurrently make large contributions in the local Shapley coalitions, and
their common importance is consistent with the residual posterior dependence
between parameters visible in Figure~\ref{fig:posterior_recovery}.

The maturity profiles nevertheless differ across parameters. For $\kappa$
and $\theta$, the largest attributions are concentrated in the positive wing
at the longest maturities, indicating that long-horizon smile information is
particularly associated with contraction of the mean-reversion-speed and
long-run-variance marginals. The attribution map for $\nu$ is more localized,
with pronounced contributions in the positive wing at long maturity and at
selected intermediate maturities. This is consistent with the role of
curvature and wing behaviour in distinguishing the volatility-of-volatility
component. For $\rho$, the positive wing is important over a broader range of
maturities, in line with its association with smile asymmetry in this
simulation design.

The profiles for $V_0$ and $H$ further show that the information is not
concentrated exclusively at the shortest maturity. The $V_0$ map 
retains an important positive-wing contribution at intermediate and longer maturities. The
map for $H$ is more distributed across the maturity dimension, with
substantial positive-wing contributions at both shorter and longer horizons.
Thus, within the present standardized $63$-point surface design, roughness is
associated with cross-maturity smile information rather than with a single
isolated short-dated quote. More generally, the absence of a uniformly
dominant shortest-maturity at-the-money region emphasizes that posterior
contraction is driven by the joint term-structure and smile geometry.

Appendix~\ref{app:sec_XAI} compares these maps with the Kullback--Leibler version of the
posterior-information functional and with Kernel-SHAP maps for the
MSE- and MAE-trained inverse networks. The Kullback--Leibler maps reproduce
the principal maturity--moneyness patterns in Figure~\ref{fig:Hellinger_XAI_heatmap},
including the repeated positive-wing concentration and the long-maturity
importance for $\kappa$ and $\theta$. This agreement is an empirical
sensitivity result: it indicates that the qualitative localization observed
here is not driven solely by the use of the bounded squared-Hellinger
functional. The point-estimator maps also display broadly similar prominent
regions. This provides useful qualitative triangulation, but not equality of
estimands: point-estimator SHAP explains variation in a conditional point
summary, whereas Hellinger--SHAP explains contraction of a marginal posterior
relative to its prior.

As specified in Section~\ref{sec:xai_hellinger}, these results remain local,
background-completion, and prior-specific attributions. They should therefore
be interpreted as properties of the stated prior-predictive rough-Heston
experiment, rather than as prior-free or strictly manifold-preserving
counterfactual statements about IV-surface informativeness.

\section{Conclusions}\label{sec:conclusion}

We have developed an uncertainty-aware neural calibration framework for the
rough Heston model.  Rather than reducing an observed implied-volatility
surface to a single calibrated parameter vector, the TRE learns an
approximation to the posterior distribution of the parameters conditional on
that surface under the specified prior-predictive simulator. The held-out
coverage diagnostics show that the 
conditional
posterior approximations are well calibrated in the simulation setting.

Propagating the learnt posterior distribution through heteroscedastic neural surrogate pricers
yields predictive distributions for potentially path-dependent exotic prices.  The
numerical study shows that visually accurate point calibration does not
guarantee reliable plug-in exotic prices for every contract specification.
The full posterior-predictive construction instead retains both residual
parameter uncertainty and conditional surrogate uncertainty.  Its
contract-specific coverage and variance decompositions show why both layers
matter: posterior uncertainty is the leading contribution for most
realized-variance claims, whereas conditional surrogate uncertainty is often
material, and sometimes dominant, for barrier claims.

We have also introduced Hellinger-SHAP to explain posterior contraction rather
than a single parameter point prediction.  The resulting attributions are
local relative contributions to posterior information gain under the stated
background-completion distribution.  They are therefore prior-specific and
should not be interpreted as strictly manifold-preserving counterfactuals;
the local-background construction makes
this limitation explicit.  Comparison with the Kullback--Leibler version
provides a useful sensitivity analysis of the information-gain target.

Several extensions merit investigation.  First, observed market surfaces
require an observation model that accounts for features such as bid--ask
spreads, missing quotes, liquidity, and potential model misspecification.
Second, a useful benchmark would compare the directly trained inverse point
estimators with calibration obtained by optimizing a learned forward
IV-surface surrogate under a specified weighted fitting criterion \citep{horvath2019deep}.  Such a
comparison would clarify when distinct point 
estimators agree and how a
forward-optimizer calibration affects the downstream plug-in error. Nevertheless, this would
still be complementary to the posterior propagation methodology developed in this paper, and not a substitute for it. Finally, the framework can be applied to alternative volatility models,
including these based on Trawl processes \citep{ole_brazilian,leonte2024simulation}
, to assess how model structure affects posterior uncertainty, posterior-predictive exotic pricing, and
information-theoretic attribution patterns.

\section*{Acknowledgements}

Large language models were used in this work to assist with text and code generation, text rewriting and code refactoring, as well as grammatical checks and improvements to phrasing and structure. All authors have reviewed and take full responsibility for any AI assisted content.

\bibliographystyle{siam}
\bibliography{bibliography_paper1}
\newpage
\appendix
\section{Synthetic dataset generation: methodological aspects}
\label{app:HQE_app}
This appendix is organised as follows:
\begin{itemize}
    \item Section~\ref{app:QE_and_MGF_for_QE} gives parameterization of the quadratic exponential (QE) distribution, together with the martingale correction used for the HQE scheme on the uniform grid.
    \item Section~\ref{app:hqe_non_uniform_grids} extends the HQE scheme to non-equidistant grids.
    \item Section~\ref{app:further_variance_reduction} details further variance reduction techniques used with the HQE scheme.
\end{itemize}

\subsection{The QE step and martingale property}
\label{app:QE_and_MGF_for_QE}
Write $X\sim \textrm{QE}(m,s^2)$ for the non-negative moment-matching variable of
\cite{andersen2008QE}, with mean $m>0$, variance $s^2\ge0$, and relative variance
$\psi\defeq s^2/m^2$. Setting
$q\defeq\half\min(\psi,3/2)$ and
$\omega \defeq 2/(\max(\psi,3/2)+1)$, and drawing
$Z \sim \mathcal N(0,1)$ and $U\sim\mathcal U[0,1)$, let
\begin{equation}
X=
\begin{dcases}
\big(\sqrt B+\sqrt A\,Z\big)^2,
\quad A=\frac{mq}{1+\sqrt{1-q}},
\quad B=\frac{m(1-q+\sqrt{1-q})}{1+\sqrt{1-q}},
& \psi\le 3/2,\\[6pt]
\frac{m}{\omega}\log\frac{\omega}{1-U} \mathbf 1_{\{1-U<\omega\}},
& \psi> 3/2.
\end{dcases}
\label{eq:qe-draw}
\end{equation}
The above construction stays non-negative and satisfies $\ev[X]=m$ and
$\textrm{Var}(X)=s^2$; we refer the reader to ~\cite{andersen2008QE} and~\cite{HQE_Gatheral_SSRN} for the
derivation and accuracy of the approximation, and why it is suitable in the Heston and rough Heston models. We use the above parameterization in $(A,B)$ because it remains well defined as $\psi\downarrow 0$. The branchwise moment-generating is
\begin{equation}
\log M(\gamma)=
\begin{dcases}
\dfrac{\gamma B}{1-2\gamma A}
-\half\log(1-2\gamma A),
& \psi\le 3/2,\quad \gamma<\tfrac{1}{2A},\\[7pt]
\log \left(1-\omega+\dfrac{\omega}{1-\gamma m/\omega}\right),
& \psi>3/2,\quad \gamma<\tfrac{\omega}{m}.
\end{dcases}
\label{eq:qe-logmgf}
\end{equation}
We next adapt the martingale-correction from \cite{andersen2008QE} to the HQE scheme. Recall $\alpha=H+\frac{1}{2}$, $\mathcal G_n=\sigma(\widehat V_0,\ldots,\widehat V_n,
\widehat\chi_0,\ldots,\widehat\chi_{n-1})$,   
$\widetilde v_n= \frac{1}{2}(\widehat V_n+\widehat V_{n+1})$. Let $\beta \defeq \frac{\nu\Delta^{\alpha-1}}{\Gamma(\alpha+1)}$,
so that 
\begin{align}
    \widehat\chi_n= & (Y_n-\half\widehat\Xi_n) / \beta, \label{eq:chi_n_revision}\\
    \widehat{V}_{n+1} = &  Y_n + E_n. \label{eq:_wide_hat_v_n_1_revision}
\end{align}

\textbf{Martingale correction} Recall the uncorrected log-asset price increment from~\eqref{eq:price-uncorrected}
\begin{equation*}
\Delta\widehat {X}_n^u = -\half\widetilde v_n\Delta + \rho\widehat\chi_n + \sqrt{1-\rho^2}\sqrt{\widetilde v_n\Delta}\,Z_n^{\perp}.
\end{equation*}
Since the Gaussian innovation $Z_n^\perp \sim \mathcal{N}(0,1)$ is independent of $Y_n$ and $E_n$ conditionally independent on $\mathcal{G}_n$, it follows by the tower property that 
\begin{align}
C_n
\defeq \log \ev\left[e^{\Delta \widehat X_n^u}\mid \mathcal G_n\right]
&= \log
\ev\left[
e^{-\half\widetilde v_n\Delta+\rho\widehat\chi_n}
\ev\left[
e^{\sqrt{1-\rho^2}\sqrt{\widetilde v_n\Delta}\,Z_n^\perp}
\Bigm|\mathcal G_n,Y_n,E_n
\right]
\Bigm|\mathcal G_n
\right] \notag\\
&= \log
\ev\left[
\exp\left\{
\rho\widehat\chi_n
-\frac{\rho^2}{2}\widetilde v_n\Delta
\right\}
\Bigm|\mathcal G_n
\right],
\label{eq:integrate-perp}
\end{align}
where we used the closed form expression for the moment generating function of a standard Gaussian. We therefore have
\begin{equation}
C_n =
\log\ev\left[
\exp\left\{
\rho\widehat\chi_n-\frac{\rho^2}{2}\widetilde v_n\Delta
\right\}
\Bigm|\mathcal G_n
\right].
\label{eq:Cn-app-def}
\end{equation}
Plugging in~\eqref{eq:chi_n_revision} and~\eqref{eq:_wide_hat_v_n_1_revision}, note that 
\begin{equation}
\rho\widehat\chi_n-\frac{\rho^2}{2}\widetilde v_n\Delta
=
 \left(\frac{\rho}{\beta}-\frac{\rho^2\Delta}{4}\right) Y_n+ -\frac{\rho^2\Delta}{4} E_n
-\frac{\rho\widehat\Xi_n}{2\beta}
-\frac{\rho^2\Delta}{4}\widehat V_n.
\label{eq:affine-exponent}
\end{equation}
 Hence
\begin{equation*}
C_n
= -\frac{\rho\widehat\Xi_n}{2\beta}
  -\frac{\rho^2\Delta}{4}\widehat V_n
  +\log M_{Y_n}\left(\frac{\rho}{\beta}-\frac{\rho^2\Delta}{4} \right)+\log M_{E_n}\left(-\frac{\rho^2\Delta}{4}\right).
\label{eq:Cn-closed}
\end{equation*}
Thus $C_n$ is $\mathcal G_n$-measurable and the corrected increment given by $\Delta\widehat X_n=\Delta\widehat X_n^u-C_n$ preserves the martingale property.
\subsection{The HQE scheme with non-uniform grids}
\label{app:hqe_non_uniform_grids}
We extend the HQE scheme to non-equidistant partitions $0=t_0<t_1<\cdots<t_N=T$. Let $\Delta_n\defeq t_{n+1}-t_n$ be the width of the $n^\textrm{th}$ cell. In our experiments, we use the graded mesh

\begin{equation*}
t_i=T\left(\frac{i}{N}\right)^r, \quad i=0,\ldots,N,
\end{equation*}
with $r\ge 1$ and where $r=1$ recovers the uniform mesh. For graded meshes, most quantities of interest, e.g., the coefficients $a_j$ from~\eqref{eq:aj}, or the parameters of the QE draws $Y_n$ and $E_n$ from~\eqref{eq:Y_n} and~\eqref{eq:E_n} depend on the cell width $\Delta_n$. While this comes with an extra computational cost, we find empirically that it yields substantial improvements in the HQE scheme's accuracy that far exceed the additional time required (see Tables~\ref{tab:hqe_lewis_quantiles} and~\ref{tab:hybrid_hqe_lewis_quantiles}-~\ref{tab:zeta_values_by_Tz}). To begin with, define
\begin{equation*}
A_{n,m} =
\frac{(t_{n+1}-t_m)^\alpha-(t_{n+1}-t_{m+1})^\alpha}
{\Gamma(\alpha+1)}, \quad m\le n,\\
\end{equation*}
and set
\begin{align*}
    \bar v_n=&(2H\widehat V_n+\widehat\Xi_n)/(2H+1),\\
    \widetilde v_n=&\half(\widehat V_n+\widehat V_{n+1}). 
\end{align*}
where $\bar v_n$ and $\widetilde v_n$ are unchanged from the main body. Conditional on $\mathcal{G}_n = \{\widehat{V}_0,\ldots,\widehat{V}_n, \widehat{\chi}_0,\ldots,\widehat{\chi}_{n-1} \}$, the $n^\text{th}$ iteration of the HQE scheme is:

\begin{enumerate}
\item Define the one-step ahead conditional-mean predictor and corresponding increment by $\widehat{u}_n$, matching~\eqref{eq:xi_n_hat} and~\ref{eq:v_n_1_u_hat_increment}
\begin{align*}
\widehat\Xi_n = &  V_0+\kappa\sum_{m=0}^{n}A_{n,m}(\theta-\widehat V_m) 
+\sum_{m=0}^{n-1} \frac{\nu A_{n,m}}{\Delta_m} \widehat\chi_m, \\ 
\widehat{V}_{n+1} = & \widehat \Xi_n + \widehat{u}_{n}
\end{align*}
\item Draw independent samples $Y_n$ and $E_n$ using~\eqref{eq:qe-draw} from 
\begin{align*}
& Y_n\sim \textrm{QE}\left(\frac{\widehat\Xi_n}{2},\ \bar v_n  \frac{\nu^2 \Delta_n^{2 \alpha -1}}{\Gamma(\alpha+1)^2}\right) = \textrm{QE} \left(\frac{\widehat\Xi_n}{2},\ \beta_n^2\bar v_n\Delta_n\right),\\
& E_n\sim \textrm{QE}\left( \frac{\widehat\Xi_n}{2}, \bar v_n\frac{\nu^2\Delta_n^{2H}}{\Gamma(\alpha)^2} \frac{(H-\half)^2}{2H\alpha^2} \right),
\end{align*}
where $\beta_n=\nu\Delta_n^{\alpha-1}/\Gamma(\alpha+1)$. This matches~\ref{eq:Y_n} and~\ref{eq:E_n}.
\item Reconstruct 
\begin{equation*}
    \widehat V_{n+1}=Y_n+E_n, \qquad \widehat\chi_n=\beta_n^{-1}\left(Y_n-\frac{\widehat\Xi_n}{2}\right), \qquad \widehat u_n=\widehat V_{n+1}-\widehat\Xi_n.
\end{equation*}
\item For the discretised log-asset price, evaluate $C_n$ in~\eqref{eq:integrate-perp} with $\Delta_n$ and $\beta_n$ in place of $\Delta$ and $\beta$ and accumulate 
\begin{equation*}
\widehat M_n \defeq \sum_{j=0}^{n-1}\widehat\chi_j,
\qquad
\widehat Q_n \defeq \sum_{j=0}^{n-1}\widetilde v_j \Delta_j,
\qquad
\widehat C_n \defeq \sum_{j=0}^{n-1} C_j.
\end{equation*}
For contracts admitting the conditional mixing representation no $Z_n^\perp$ draw is necessary, as we discuss in Section~\ref{app:further_variance_reduction}; otherwise, we use \[ \Delta\widehat X_n =-\half\widetilde v_n\Delta_n+\rho\widehat\chi_n-C_n +\sqrt{1-\rho^2}\sqrt{\widetilde v_n\Delta_n}\,Z_n^\perp. \]
\end{enumerate} 
We next discuss the application of the HQE scheme to pricing options. Whereas a direct approach based on a discretisation of the (log-) asset price is possible, it generally yields unsatisfactory estimators without further variance reduction techniques.

\subsection{The HQE scheme applied for option pricing and further variance reduction}
\label{app:further_variance_reduction}

We consider European calls, together with the families of exotic contracts from Section~\ref{sec:exotic_contract_description}: forward-start calls, calls on realised variance, and down-and-out put options. In all cases, the HQE variance paths are generated using antithetic samples for the QE draws. Specifically, let $(Z_n^Y, U_n^Y)$ be the samples used to generate $Y_n$ and $(Z_n^E,U_n^E)$ the samples used for $E_n$ in the QE sampler~\eqref{eq:qe-draw}, and consider the variance paths using the flipped draws
\begin{equation*}
(Z_n^Y,Z_n^E,U_n^Y,U_n^E) \mapsto(-Z_n^Y,-Z_n^E,1-U_n^Y,1-U_n^E),
\end{equation*}
for all $n \in \{1,\ldots,N\}.$ Unlike a flipped Gaussian increment, the nonlinear, two-branch QE map does not in general yield as high of a variance reduction. Nevertheless, it provides variance reduction at no additional computational cost. Hence, we apply this strategy for all four contracts.

\subsubsection{European and forward-start calls}
For a European call with maturity $t_n$, the contribution of the orthogonal Brownian motion $W^\perp$ can be integrated out analytically, as discussed before. Conditional on the simulated HQE variance paths,
\begin{equation*}
\log \widehat S_{t_n}
\sim
\mathcal N\left(
\log S_0+\rho\widehat M_n-\half\widehat Q_n-\widehat C_n,
(1-\rho^2)\widehat Q_n
\right).
\end{equation*}
Thus, the conditional MC estimator is given by
\begin{equation*}
\ev\left[(\widehat S_{t_n}-K)^{+}\mid\mathcal G_n\right]
=
C_{\mathrm{BS}}\left(
\widehat F_n^{\,c},
K,
(1-\rho^2)\widehat Q_n
\right),
\end{equation*}
where
\begin{align*}
\widehat F_n^{\,c}
&=
S_0\exp\left(
\rho\widehat M_n-\half\rho^2\widehat Q_n-\widehat C_n
\right),\\
C_{\mathrm{BS}}(F,K,w)
&=F\Phi(d_1)-K\Phi(d_2),
\end{align*}
with $d_1=\frac{\log(F/K)+\half w}{\sqrt w}$, $d_2=d_1-\sqrt w$ for $w >0$ and $\Phi(\cdot)$ the cumulative distribution function of the standard normal.

The same conditional approach applies to the forward-start payoff $(S_{T_2}-e^kS_{T_1})^+$. With $T_1=t_{n_1} < T_2=t_{n_2}$, write the payoff as
\[
\left(S_{T_2} - e^k \cdot S_{T_1}\right)^{+} = S_{T_1}\left(\frac{S_{T_2}}{S_{T_1}}-e^k\right)^+.
\]
By conditional independence and subsequently by the tower property, we have that
\begin{align*}
\ev\left[
(\widehat S_{T_2}-e^k\widehat S_{T_1})^{+} \Bigm| \mathcal{G}_{n_2}
\right] =& \ev\left[S_{T_1}\Bigm| \mathcal{G}_{n_1} \right] \cdot \ev \left[
\left(\frac{S_{T_2}}{S_{T_1}}-e^k\right)^+
\Bigm| \mathcal{G}_{n_2} 
\right] \notag\\
=&
\widehat{F}_{n_1}^{\,c} \cdot
C_{\mathrm{BS}} \left(
\widehat{F}_{n_1,n_2}^c,
e^k,
(1-\rho^2)(\widehat Q_{n_2}-\widehat Q_{n_1})
\right),
\end{align*}
where 
\begin{equation*}
    \widehat{F}_{n_1,n_2}^c \defeq \exp \left\{
\rho(\widehat M_{n_2}-\widehat M_{n_1})
-\half\rho^2(\widehat Q_{n_2}-\widehat Q_{n_1})
-(\widehat C_{n_2}-\widehat C_{n_1})
\right\}
\end{equation*}
Overall, the contribution of $W^\perp$ is integrated out analytically in both cases.

\subsubsection{Realised-variance calls}
For the realised-variance payoff
\begin{equation*}
\left(
\frac{1}{T}\int_0^T V_t\,\mathrm dt-K_{\mathrm{RV}}
\right)^{+},
\end{equation*}
no additional variance reduction is applied beyond the QE antithetic construction. If $T=t_n$, the discretised payoff is directly available from the discretised $V.$ path
\begin{equation*}
\left(
\frac{\widehat Q_n}{T}-K_{\mathrm{RV}}
\right)^{+}.
\end{equation*}

\subsubsection{Barrier options}
For the path-dependent, non-smooth down-and-out put payoff
\begin{equation*}
(K-S_T)^+\mathbf 1\left(\inf_{0\le t\le T}S_t>B\right),
\end{equation*}
the conditional MC representation is not available. Instead, we reduce the variability arising from $W^\perp$ by reusing each simulated variance path across $R$ independent asset-price paths. Each of the $R$ paths is paired with its antithetic counterpart obtained by replacing $Z_j^\perp$ with $-Z_j^\perp$, and the resulting payoffs are averaged. Based on empirical validation, we use $R=3$; this recovers much of the variance reduction obtained by analytic conditioning, without repeating the comparatively expensive variance discretisation.

To account for barrier crossings between consecutive grid points, we additionally apply a Brownian-bridge continuity correction \citep{glasserman2004monte}. If both endpoints of a cell lie above the barrier, the within-cell survival probability is approximated by
\begin{equation*}
p_j^{\mathrm{surv}}
=
1-\exp\left(
-\frac{2(\log\widehat S_{j}-\log B)(\log\widehat S_{j+1}-\log B)}{\widetilde v_j\Delta_j}
\right),
\end{equation*}
The barrier payoff is weighted by the product of the survival probabilities.

\clearpage
\newpage
\clearpage
\section{Extended numerical validation of the synthetic dataset}
\label{app:sec_extended_validation_of_the_dataset}
This appendix is organised as follows:
\begin{itemize}
    \item  Section~\ref{app:extended_validation_of_the_IV_dataset} presents additional diagnostics on the quality of the approximations given by the hybrid and HQE schemes for generating implied volatility surfaces. In particular, we present a benchmark against Lewis pricing, which relies on numerically inverting the characteristic function of the asset price.
    \item  Section~\ref{app:extended_validation_of_the_exotic_dataset} summarizes the exotic contract dataset.
\end{itemize}
\subsection{European option prices and IV surfaces}
\label{app:extended_validation_of_the_IV_dataset}
Recall the notation from~\eqref{eq:HQE-Lewis-error}. For each parameter vector $\bt$ and grid point $(T,z)$, define the absolute value of the difference between the European implied volatilities against a Lewis inversion based on the rough Heston characteristic function. 
\begin{equation*}
    \Delta^\textrm{scheme}_{\mathrm{IV}}(T,z;\bt)
    =
    \left|
        \widehat{\sigma}^{\mathrm{scheme}}(T,z;\bt)
        -
        \sigma^{\mathrm{Lewis}}(T,z;\bt)
    \right|,
\end{equation*}
where $\textrm{scheme} \in \{\textrm{hybrid},\textrm{HQE}\}.$ We implement two version of the hybrid scheme \citep{bennedsen2017hybrid}, one where only the head-cell is dealt with in closed form, i.e. $K_h =1$, and one in which both the head-cell and the following cell are treated exactly, i.e., $K_h =2$. We refer the reader to ~\cite{bennedsen2017hybrid} for full details on the hybrid scheme. For Lewis inversion, we use the standard Adams predictor-corrector scheme followed by numerical inversion of the characteristic function. We stress that while extensive hyperparameter tuning was performed to obtain accurate results for Lewis inversion, it is worth keeping in mind that the values from Tables~\ref{tab:hybrid_hqe_lewis_quantiles}-~\ref{tab:zeta_values_by_Tz} reflect approximation errors for both Lewis and HQE.

We start by analyzing the results from Table~\ref{tab:hybrid_hqe_lewis_quantiles}. Note that $K_h=2$ barely improves over $K_h=1$ in the hybrid scheme, and that the HQE scheme is substantially more accurate than the hybrid scheme, regardless of the rate $r$. We attribute the inefficiency of the hybrid scheme in the rough Heston model to the bias induced by the truncation $V^{+} = \max\{V,0\}$, especially in the following regimes: rough regime (low $H$), low initial volatility $V_0$, or low long-term volatility $\theta$. Indeed, we monitor the fraction of steps where truncation is necessary and note empirically that more truncation yields larger values for $\Delta^\textrm{hybrid}_{\mathrm{IV}}(T,z;\bt)$. The QE step in the HQE scheme directly addresses this limitation and indeed yields far lower values $\Delta^\textrm{HQE}_{\mathrm{IV}}(T,z;\bt)$, especially at the $95\%$ and $99\%$ quantiles. Also note that graded meshes further and substantially improve accuracy. Indeed, for the HQE scheme, $N=1008$ with $r=2$ gives better results than $N=4032$ with $r=1$. This amounts to a $16$ fold reduction in the $O(N^2)$ complexity of the scheme, with only minor additional computation required for the non-uniform partition. To ensure good accuracy over the entire prior, we decide to use the HQE scheme with $N=2016$ discretisation steps between $0$ and $T_\text{max}=2$, rate $r=2$ on the graded mesh, and $M=5 \cdot 10^5$ paths in the remainder of the study. Having settled on these values, we provide a breakdown of the $\Delta^\textrm{HQE}_{\mathrm{IV}}(T,z;\bt)$ values over pairs $(T,z)$ in Table~\ref{tab:hqe_lewis_by_Tz}. We note good accuracy, with most $95\%$ and $99\%$ quantiles remaining below the $10$ and $15$ basis point levels, respectively. Slightly larger values are observed in the wings, at very short maturities. Finally, Tables~\ref{tab:CV_european} and~\ref{tab:zeta_values_by_Tz} provide quantiles of the coefficient of variation, i.e., the standard error of the MC estimator divided by the corresponding MC price, first pooled over all pairs $(T,z)$ and then with a breakdown for each pair. Based on the results we conclude that the potential bias incurred by the HQE scheme is on the same order, or smaller than the MC noise. We emphasize that $M = 5 \cdot 10^5$ is already a large number of paths to use for MC pricing of each option. We next turn our attention to exotic contracts.

\begin{table}[t]
    \centering
    \caption{Quantiles $\Delta^\textrm{hybrid}_{\mathrm{IV}}(T,z;\bt)$ and $\Delta^\textrm{HQE}_{\mathrm{IV}}(T,z;\bt)$ of the absolute difference between implied volatilities derived from Lewis pricing and those produced by inverting prices computed  with the hybrid and HQE schemes, pooled over all $(T,z)$. The HQE scheme gives IVs that are substantially closer to the Lewis benchmark than the hybrid scheme, even when $K_h=2$ for the hybrid scheme, in which both the head-cell and the following cell are treated exactly. We use $M= 5\cdot 10^5$ paths for the MC estimation. Quantiles are estimated based on $500$ pairs $(\bt,\x)$ drawn from the prior.}
    \label{tab:hybrid_hqe_lewis_quantiles}
 \small
\setlength{\tabcolsep}{3pt}
\begin{tabular}{c @{\hspace{8pt}} c | @{\hspace{5pt}}
                cccc @{\hspace{10pt}}
                cccc @{\hspace{10pt}}
                cccc @{\hspace{10pt}}
                cccc}
        \toprule
        & & \multicolumn{4}{c}{$N=504$}
        & \multicolumn{4}{c}{$N=1008$}
        & \multicolumn{4}{c}{$N=2016$}
        & \multicolumn{4}{c}{$N=4032$} \\
        \cmidrule(lr){3-6}
        \cmidrule(lr){7-10}
        \cmidrule(lr){11-14}
        \cmidrule(lr){15-18}
        \shortstack{Simulation\\scheme} & $r$
        & $q_{50}$ & $q_{90}$ & $q_{95}$ & $q_{99}$
        & $q_{50}$ & $q_{90}$ & $q_{95}$ & $q_{99}$
        & $q_{50}$ & $q_{90}$ & $q_{95}$ & $q_{99}$
        & $q_{50}$ & $q_{90}$ & $q_{95}$ & $q_{99}$ \\
        \midrule

        \multirow{3}{*}{Hybrid ($K_h=1$)}
        & 1 & 0.06 & 0.64 & 1.14 & 2.45 & 0.04 & 0.46 & 0.97 & 2.11 & 0.03 & 0.34 & 0.84 & 1.80 & 0.02 & 0.28 & 0.72 & 1.54 \\
        & 2 & 0.03 & 0.40 & 0.98 & 1.84 & 0.02 & 0.32 & 0.85 & 1.59 & 0.02 & 0.27 & 0.73 & 1.39 & 0.02 & 0.22 & 0.62 & 1.21 \\
        & 3 & 0.03 & 0.39 & 0.96 & 1.78 & 0.02 & 0.32 & 0.83 & 1.54 & 0.02 & 0.26 & 0.71 & 1.34 & 0.02 & 0.22 & 0.60 & 1.17 \\
        \addlinespace         \addlinespace

        \multirow{3}{*}{Hybrid ($K_h=2$)}
        & 1 & 0.06 & 0.65 & 1.13 & 2.44 & 0.04 & 0.44 & 0.95 & 2.08 & 0.03 & 0.35 & 0.83 & 1.78 & 0.02 & 0.28 & 0.71 & 1.54 \\
        & 2 & 0.03 & 0.40 & 0.96 & 1.83 & 0.02 & 0.33 & 0.84 & 1.59 & 0.02 & 0.27 & 0.72 & 1.38 & 0.02 & 0.22 & 0.61 & 1.21 \\
        & 3 & 0.03 & 0.39 & 0.95 & 1.76 & 0.02 & 0.32 & 0.83 & 1.55 & 0.02 & 0.26 & 0.71 & 1.32 & 0.02 & 0.22 & 0.60 & 1.16 \\
        \addlinespace         \addlinespace

        \multirow{3}{*}{HQE}
        & 1 & 0.03 & 0.23 & 0.38 & 0.80 & 0.02 & 0.15 & 0.23 & 0.48 & 0.02 & 0.10 & 0.15 & 0.29 & 0.02 & 0.08 & 0.12 & 0.21 \\
        & 2 & 0.02 & 0.10 & 0.14 & 0.24 & 0.02 & 0.08 & 0.11 & 0.19 & 0.02 & 0.07 & 0.10 & 0.15 & 0.01 & 0.07 & 0.09 & 0.15 \\
        & 3 & 0.02 & 0.10 & 0.13 & 0.22 & 0.02 & 0.08 & 0.11 & 0.17 & 0.02 & 0.07 & 0.09 & 0.15 & 0.01 & 0.07 & 0.10 & 0.17 \\

        \bottomrule
    \end{tabular}
\end{table}

\begin{table*}[!ht]
    \centering
    \caption{Absolute errors $\Delta_{\mathrm{IV}}^\textrm{HQE}(T,z;\bt)$ between HQE and Lewis implied volatilities. This table provides a breakdown by $(T,z)$ pairs of the values reported in Table~\ref{tab:hybrid_hqe_lewis_quantiles}.
    }
\label{tab:hqe_lewis_by_Tz}
\small

\begin{minipage}[t]{0.48\textwidth}
\vspace{0pt}
\centering
\setlength{\tabcolsep}{4pt}
\begin{tabular}{cc|rrrr}
\toprule
$T$ & $z$ & $q_{50}$ & $q_{90}$ & $q_{95}$ & $q_{99}$ \\
\midrule
\multirow{9}{*}{$5/252$}
& $-2.5$ & 0.02 & 0.12 & 0.16 & 0.29 \\

& $-2$ & 0.02 & 0.10 & 0.13 & 0.21 \\

& $-1.5$ & 0.02 & 0.08 & 0.10 & 0.16 \\

& $-1$ & 0.02 & 0.07 & 0.10 & 0.16 \\

& $-0.5$ & 0.02 & 0.07 & 0.10 & 0.15 \\

& $0$ & 0.01 & 0.07 & 0.10 & 0.15 \\

& $0.5$ & 0.01 & 0.07 & 0.10 & 0.15 \\

& $1$ & 0.01 & 0.07 & 0.10 & 0.16 \\

& $1.5$ & 0.01 & 0.07 & 0.10 & 0.17 \\
\addlinespace \addlinespace

\multirow{9}{*}{$10/252$}
& $-2.5$ & 0.02 & 0.10 & 0.14 & 0.22 \\

& $-2$ & 0.02 & 0.09 & 0.11 & 0.19 \\

& $-1.5$ & 0.02 & 0.07 & 0.10 & 0.15 \\

& $-1$ & 0.02 & 0.07 & 0.09 & 0.14 \\

& $-0.5$ & 0.01 & 0.07 & 0.09 & 0.14 \\

& $0$ & 0.01 & 0.07 & 0.09 & 0.13 \\

& $0.5$ & 0.01 & 0.07 & 0.09 & 0.12 \\

& $1$ & 0.01 & 0.06 & 0.09 & 0.13 \\

& $1.5$ & 0.01 & 0.06 & 0.09 & 0.14 \\
\addlinespace \addlinespace

\multirow{9}{*}{$1/12$}
& $-2.5$ & 0.03 & 0.11 & 0.15 & 0.22 \\

& $-2$ & 0.02 & 0.09 & 0.12 & 0.18 \\

& $-1.5$ & 0.02 & 0.08 & 0.11 & 0.16 \\

& $-1$ & 0.02 & 0.08 & 0.10 & 0.15 \\

& $-0.5$ & 0.02 & 0.07 & 0.10 & 0.15 \\

& $0$ & 0.01 & 0.07 & 0.09 & 0.14 \\

& $0.5$ & 0.01 & 0.07 & 0.08 & 0.14 \\

& $1$ & 0.01 & 0.06 & 0.08 & 0.13 \\

& $1.5$ & 0.01 & 0.06 & 0.08 & 0.12 \\
\addlinespace \addlinespace

\multirow{9}{*}{$1/4$}
& $-2.5$ & 0.03 & 0.10 & 0.14 & 0.24 \\

& $-2$ & 0.02 & 0.09 & 0.12 & 0.19 \\

& $-1.5$ & 0.02 & 0.08 & 0.10 & 0.15 \\

& $-1$ & 0.02 & 0.07 & 0.09 & 0.13 \\

& $-0.5$ & 0.01 & 0.06 & 0.09 & 0.13 \\

& $0$ & 0.01 & 0.05 & 0.08 & 0.13 \\

& $0.5$ & 0.01 & 0.05 & 0.07 & 0.12 \\

& $1$ & 0.01 & 0.05 & 0.07 & 0.11 \\

& $1.5$ & 0.01 & 0.04 & 0.06 & 0.11 \\

\bottomrule
\end{tabular}
\end{minipage}
\hfill
\begin{minipage}[t]{0.48\textwidth}
\vspace{0pt}
\centering
\setlength{\tabcolsep}{4pt}
\begin{tabular}{cc|rrrr}
\toprule
$T$ & $z$ & $q_{50}$ & $q_{90}$ & $q_{95}$ & $q_{99}$ \\
\midrule

\multirow{9}{*}{$1/2$}
& $-2.5$ & 0.03 & 0.09 & 0.12 & 0.17 \\

& $-2$ & 0.02 & 0.08 & 0.10 & 0.15 \\

& $-1.5$ & 0.02 & 0.07 & 0.09 & 0.13 \\

& $-1$ & 0.02 & 0.06 & 0.08 & 0.12 \\

& $-0.5$ & 0.01 & 0.06 & 0.08 & 0.11 \\

& $0$ & 0.01 & 0.06 & 0.08 & 0.11 \\

& $0.5$ & 0.01 & 0.05 & 0.07 & 0.11 \\

& $1$ & 0.01 & 0.05 & 0.07 & 0.11 \\

& $1.5$ & 0.01 & 0.05 & 0.06 & 0.10 \\
\addlinespace \addlinespace

\multirow{9}{*}{$1$}
& $-2.5$ & 0.03 & 0.09 & 0.12 & 0.16 \\

& $-2$ & 0.02 & 0.08 & 0.10 & 0.14 \\

& $-1.5$ & 0.02 & 0.07 & 0.09 & 0.12 \\

& $-1$ & 0.02 & 0.06 & 0.08 & 0.12 \\

& $-0.5$ & 0.01 & 0.06 & 0.08 & 0.12 \\

& $0$ & 0.01 & 0.05 & 0.07 & 0.11 \\

& $0.5$ & 0.01 & 0.05 & 0.06 & 0.10 \\

& $1$ & 0.01 & 0.05 & 0.06 & 0.10 \\

& $1.5$ & 0.01 & 0.05 & 0.06 & 0.09 \\
\addlinespace \addlinespace

\multirow{9}{*}{$2$}
& $-2.5$ & 0.02 & 0.09 & 0.12 & 0.17 \\

& $-2$ & 0.02 & 0.08 & 0.11 & 0.15 \\

& $-1.5$ & 0.02 & 0.07 & 0.09 & 0.13 \\

& $-1$ & 0.02 & 0.07 & 0.08 & 0.13 \\

& $-0.5$ & 0.01 & 0.06 & 0.08 & 0.13 \\

& $0$ & 0.01 & 0.06 & 0.08 & 0.12 \\

& $0.5$ & 0.01 & 0.05 & 0.07 & 0.12 \\

& $1$ & 0.01 & 0.05 & 0.07 & 0.12 \\

& $1.5$ & 0.01 & 0.05 & 0.07 & 0.12 \\
\bottomrule
\end{tabular}
\end{minipage}
\end{table*}
\clearpage
\begin{table}[t]
    \centering
    \caption{Quantiles of the coefficient of variation for the MC estimators of European options, pooled over all pairs $(T,z)$. The coefficient of variation is defined as the standard error of the MC estimator, divided by the corresponding MC price. We conclude that for our choice of prior, $N \ge 2016$ discretisation steps with $r =2$ (or $r=3$) yields HQE discretisation bias that is roughly on the same order of magnitude, or lower than that of the randomness inherent to MC estimation with $M= 5\cdot 10^5$ samples. As before, quantiles are estimated based on $500$ pairs $(\bt,\x)$ drawn from the prior.}
    \begin{tabular}{c|cc cc cc cc}
\toprule
& \multicolumn{2}{c}{$N=504$}
& \multicolumn{2}{c}{$N=1008$}
& \multicolumn{2}{c}{$N=2016$}
& \multicolumn{2}{c}{$N=4032$} \\
\cmidrule(lr){2-3}
\cmidrule(lr){4-5}
\cmidrule(lr){6-7}
\cmidrule(lr){8-9}
$r$
& $q_{50}$ & $q_{95}$
& $q_{50}$ & $q_{95}$
& $q_{50}$ & $q_{95}$
& $q_{50}$ & $q_{95}$ \\
\midrule
$1$ & 1.57 & 12.00 & 1.16 & 7.06 & 0.92 & 4.67 & 0.82 & 3.37 \\
$2$ & 0.99 & 4.78 & 0.83 & 3.43 & 0.75 & 2.93 & 0.71 & 2.65 \\
$3$ & 0.95 & 4.53 & 0.82 & 3.35 & 0.74 & 2.91 & 0.72 & 2.60 \\
\bottomrule
\end{tabular}
\label{tab:CV_european}
\end{table}
\begin{table*}[t]
    \centering
    \caption{Quantiles of the coefficient of variation for the MC estimators of European options. This provides a breakdown of the values reported in Table~\ref{tab:CV_european} by $(T,z)$ pairs. The three largest values at the $95\%$ level are $5.12$, $4.98$, $4.90$ for maturities $T = 5/252$,$1/12$,$10/252$ respectively and $z=1.5$. These correspond to deep out-of-the-money contracts where the MC price is small, thus inflating the relative MC error.}
\label{tab:zeta_values_by_Tz}
\small

\begin{minipage}[t]{0.48\textwidth}
\vspace{0pt}
\centering
\setlength{\tabcolsep}{4pt}
\begin{tabular}{cc|rrrr}
\toprule
$T$ & $z$ & $q_{50}$ &$q_{95}$  \\
\midrule
\multirow{9}{*}{$5/252$}
& $-2.5$ & 0.68 & 2.13 \\

& $-2$ & 0.68 & 1.87 \\

& $-1.5$ & 0.66 & 1.82 \\

& $-1$ & 0.63 & 1.69 \\

& $-0.5$ & 0.59 & 1.58 \\

& $0$ & 1.11 & 3.64 \\

& $0.5$ & 1.21 & 4.02 \\

& $1$ & 1.21 & 4.60 \\

& $1.5$ & 1.25 & 5.12 \\
\addlinespace \addlinespace

\multirow{9}{*}{$10/252$}
& $-2.5$ & 0.65 & 2.00 \\

& $-2$ & 0.62 & 1.86 \\

& $-1.5$ & 0.57 & 1.79 \\

& $-1$ & 0.54 & 1.71 \\

& $-0.5$ & 0.52 & 1.59 \\

& $0$ & 1.00 & 3.50 \\

& $0.5$ & 1.13 & 4.15 \\

& $1$ & 1.24 & 4.53 \\

& $1.5$ & 1.20 & 4.90 \\
\addlinespace \addlinespace

\multirow{9}{*}{$1/12$}
& $-2.5$ & 0.67 & 1.96 \\

& $-2$ & 0.65 & 1.80 \\

& $-1.5$ & 0.63 & 1.76 \\

& $-1$ & 0.60 & 1.77 \\

& $-0.5$ & 0.53 & 1.68 \\

& $0$ & 1.04 & 3.93 \\

& $0.5$ & 1.16 & 4.27 \\

& $1$ & 1.25 & 4.80 \\

& $1.5$ & 1.17 & 4.98 \\
\addlinespace \addlinespace

\multirow{9}{*}{$1/4$}
& $-2.5$ & 0.66 & 2.10 \\

& $-2$ & 0.67 & 1.99 \\

& $-1.5$ & 0.63 & 1.82 \\

& $-1$ & 0.61 & 1.75 \\

& $-0.5$ & 0.54 & 1.63 \\

& $0$ & 0.87 & 3.41 \\

& $0.5$ & 1.05 & 3.40 \\

& $1$ & 1.10 & 3.89 \\

& $1.5$ & 1.09 & 4.30 \\

\bottomrule
\end{tabular}
\end{minipage}
\hfill
\begin{minipage}[t]{0.48\textwidth}
\vspace{0pt}
\centering
\setlength{\tabcolsep}{4pt}
\begin{tabular}{cc|rrrr}
\toprule
$T$ & $z$ & $q_{50}$ & $q_{95}$ \\
\midrule

\multirow{9}{*}{$1/2$}
& $-2.5$ & 0.63 & 1.84 \\

& $-2$ & 0.61 & 1.86 \\

& $-1.5$ & 0.59 & 1.73 \\

& $-1$ & 0.52 & 1.64 \\

& $-0.5$ & 0.51 & 1.55 \\

& $0$ & 0.78 & 3.09 \\

& $0.5$ & 0.93 & 3.43 \\

& $1$ & 1.10 & 3.75 \\

& $1.5$ & 1.18 & 4.28 \\
\addlinespace \addlinespace

\multirow{9}{*}{$1$}
& $-2.5$ & 0.68 & 1.97 \\

& $-2$ & 0.67 & 1.87 \\

& $-1.5$ & 0.64 & 1.90 \\

& $-1$ & 0.58 & 1.89 \\

& $-0.5$ & 0.53 & 1.85 \\

& $0$ & 0.69 & 2.77 \\

& $0.5$ & 0.85 & 3.35 \\

& $1$ & 0.94 & 3.59 \\

& $1.5$ & 1.04 & 4.08 \\
\addlinespace \addlinespace

\multirow{9}{*}{$2$}
& $-2.5$ & 0.63 & 1.99 \\

& $-2$ & 0.66 & 1.97 \\

& $-1.5$ & 0.64 & 1.93 \\

& $-1$ & 0.60 & 1.95 \\

& $-0.5$ & 0.60 & 1.98 \\

& $0$ & 0.59 & 2.17 \\

& $0.5$ & 0.78 & 2.58 \\

& $1$ & 0.93 & 3.13 \\

& $1.5$ & 1.11 & 3.72 \\
\bottomrule
\end{tabular}
\end{minipage}
\end{table*}
\clearpage
\subsection{Exotic contracts}
\label{app:extended_validation_of_the_exotic_dataset}
Recall the exotic contracts defined in Section~\ref{sec:exotic_contract_description}. In Figure~\ref{fig:mc_relative_se}, we displayed the coefficient of variation for MC price estimates each of the three families of exotic contracts, with results pooled over all the contracts within each family. We next provide more detailed, contract-level summary statistics
.

For the forward start and barrier contracts (Tables~\ref{tab:forward_MC_summary_stats} and~\ref{tab:barrier_MC_summary_stats}), we report the minimum and the $1\%,5\%,50\%,95\%,99\%$ quantiles of the MC prices across the parameter vectors $\bt$ in the simulated dataset; we also display the $95\%$ and $99\%$ quantiles of the coefficient of variation (i.e., ratio of the standard error of the MC estimator to the MC price) and remind the reader that we simulated $500000$ pairs $(\bt,\x)$. For both families, the minimum MC prices are bounded away from $0$, and the quantiles reported for the coefficient of variation show that our MC estimators have standard deviations that are far smaller than the corresponding estimated prices.

For the realized-variance contracts, we require a two-step filtering approach. We remind the reader that we initially considered  $(c,T) \in\{0.25,0.45,0.65\} \times \left\{\frac{1}{12},\frac14,\frac12,1\right\}$, and then discarded the two contracts corresponding to $c = 0.65$ and $T \in \{\frac{1}{12},\frac{1}{4}\}$ because they had a large fraction of MC prices equal to 0. This leaves $10$ out of $12$ economically relevant contracts. For the purposes of displaying results in Table~\ref{tab:rv_MC_diagnostics}, we apply a second, observation-level filter within each retained contract. Specifically, Table~\ref{tab:rv_MC_diagnostics} reports the fractions of MC prices that are exactly $0$ and below $10^{-5}$, respectively. We compute the reported price quantiles and coefficient-of-variation quantities only from MC prices that are above $10^{-5}$. 
This is the exact same filtering applied to RV contracts in
Sections~\ref{sec:exo_mc_variance} and~\ref{sec:exo_predictive_intervals} and
Appendix~\ref{app_Extended_UQ}. Importantly, this filtering is contract-specific:
if the MC price falls below $\epsilon$ for one RV contract at a given parameter vector $\bt$, the other retained $(c,T)$ contracts for the same $\bt$ are not discarded automatically.
\begin{remark}
We note that when training the surrogate pricers for the RV contracts, we also use $\epsilon=10^{-5}$; see Section~\ref{sec:sbi-surrogate} and Appendix~\ref{app:surrogate_pricers}. Its role there is different: $\epsilon$ is introduced as a numerical jitter in the transformed training target and is subsequently subtracted, see, for example,~\eqref{eq:posterior-predictive}. Thus, during training, $\epsilon$ serves only as a numerical stabilization mechanism and does not amount to excluding any samples, whereas here it defines the threshold used to exclude economically negligible RV prices from evaluation.
\end{remark}
\begin{table}
\caption{Summary statistics for the dataset of forward-start MC prices.}
\begin{tabular}{rrr@{\hspace{1.0em}}rrrrrr@{\hspace{1.0em}}rr}
\toprule
\multicolumn{3}{c}{} & \multicolumn{6}{c}{$\widehat{P}^{\mathrm{MC}}$} & \multicolumn{2}{c}{$\mathrm{CV}$} \\
\cmidrule(lr){4-9}\cmidrule(lr){10-11}
$T_1$ & $T_2$ & $k$ & $\textrm{min}$ & $q_{01}$ & $q_{05}$ & $q_{50}$ & $q_{95}$ & $q_{99}$ & $q_{95}$ & $q_{99}$ \\
\midrule
0.08 & 0.17 & -0.10 & 9.51711 & 9.57 & 9.68 & 10.50 & 11.32 & 11.46 & 0.000 & 0.001 \\
0.08 & 0.17 & -0.05 & 4.94331 & 5.20 & 5.50 & 6.95 & 8.06 & 8.24 & 0.001 & 0.001 \\
0.08 & 0.17 & 0.00 & 0.67721 & 1.49 & 2.06 & 4.02 & 5.31 & 5.52 & 0.001 & 0.001 \\
0.08 & 0.17 & 0.05 & 0.01395 & 0.16 & 0.41 & 1.96 & 3.20 & 3.42 & 0.002 & 0.003 \\
0.08 & 0.17 & 0.10 & 0.00001 & 0.01 & 0.05 & 0.79 & 1.75 & 1.95 & 0.005 & 0.010 \\
0.08 & 0.33 & -0.10 & 9.58587 & 10.00 & 10.48 & 12.68 & 14.44 & 14.74 & 0.001 & 0.001 \\
0.08 & 0.33 & -0.05 & 5.32628 & 6.05 & 6.81 & 9.64 & 11.64 & 11.99 & 0.001 & 0.001 \\
0.08 & 0.33 & 0.00 & 1.23481 & 2.70 & 3.72 & 6.96 & 9.12 & 9.51 & 0.001 & 0.001 \\
0.08 & 0.33 & 0.05 & 0.06900 & 0.76 & 1.61 & 4.74 & 6.92 & 7.34 & 0.002 & 0.002 \\
0.08 & 0.33 & 0.10 & 0.00336 & 0.16 & 0.55 & 3.02 & 5.09 & 5.51 & 0.002 & 0.004 \\
0.25 & 0.33 & -0.10 & 9.51771 & 9.59 & 9.72 & 10.48 & 11.24 & 11.39 & 0.001 & 0.001 \\
0.25 & 0.33 & -0.05 & 4.94699 & 5.24 & 5.56 & 6.90 & 7.95 & 8.15 & 0.001 & 0.001 \\
0.25 & 0.33 & 0.00 & 0.60623 & 1.51 & 2.12 & 3.96 & 5.19 & 5.44 & 0.001 & 0.001 \\
0.25 & 0.33 & 0.05 & 0.01690 & 0.20 & 0.48 & 1.91 & 3.09 & 3.35 & 0.002 & 0.004 \\
0.25 & 0.33 & 0.10 & 0.00003 & 0.02 & 0.09 & 0.77 & 1.67 & 1.89 & 0.005 & 0.009 \\
0.25 & 0.50 & -0.10 & 9.58937 & 10.04 & 10.54 & 12.63 & 14.30 & 14.64 & 0.001 & 0.001 \\
0.25 & 0.50 & -0.05 & 5.32549 & 6.08 & 6.87 & 9.57 & 11.49 & 11.89 & 0.001 & 0.001 \\
0.25 & 0.50 & 0.00 & 1.10548 & 2.70 & 3.77 & 6.88 & 8.97 & 9.41 & 0.001 & 0.001 \\
0.25 & 0.50 & 0.05 & 0.09346 & 0.82 & 1.67 & 4.66 & 6.78 & 7.25 & 0.002 & 0.002 \\
0.25 & 0.50 & 0.10 & 0.00739 & 0.21 & 0.62 & 2.96 & 4.96 & 5.42 & 0.002 & 0.004 \\
0.50 & 0.58 & -0.10 & 9.51828 & 9.60 & 9.74 & 10.46 & 11.19 & 11.36 & 0.001 & 0.001 \\
0.50 & 0.58 & -0.05 & 4.95095 & 5.26 & 5.59 & 6.87 & 7.89 & 8.12 & 0.001 & 0.001 \\
0.50 & 0.58 & 0.00 & 0.53772 & 1.50 & 2.13 & 3.91 & 5.12 & 5.40 & 0.001 & 0.001 \\
0.50 & 0.58 & 0.05 & 0.01827 & 0.22 & 0.51 & 1.88 & 3.03 & 3.32 & 0.002 & 0.004 \\
0.50 & 0.58 & 0.10 & 0.00005 & 0.03 & 0.10 & 0.76 & 1.63 & 1.86 & 0.005 & 0.008 \\
0.50 & 0.75 & -0.10 & 9.59377 & 10.06 & 10.56 & 12.58 & 14.22 & 14.59 & 0.001 & 0.001 \\
0.50 & 0.75 & -0.05 & 5.31797 & 6.08 & 6.88 & 9.50 & 11.41 & 11.83 & 0.001 & 0.001 \\
0.50 & 0.75 & 0.00 & 0.96526 & 2.67 & 3.77 & 6.80 & 8.88 & 9.36 & 0.001 & 0.001 \\
0.50 & 0.75 & 0.05 & 0.09211 & 0.83 & 1.68 & 4.58 & 6.70 & 7.20 & 0.002 & 0.002 \\
0.50 & 0.75 & 0.10 & 0.00814 & 0.24 & 0.65 & 2.90 & 4.89 & 5.38 & 0.002 & 0.004 \\
1.00 & 1.08 & -0.10 & 9.51816 & 9.60 & 9.73 & 10.43 & 11.16 & 11.32 & 0.001 & 0.001 \\
1.00 & 1.08 & -0.05 & 4.95197 & 5.24 & 5.56 & 6.81 & 7.85 & 8.08 & 0.001 & 0.001 \\
1.00 & 1.08 & 0.00 & 0.45711 & 1.44 & 2.06 & 3.83 & 5.09 & 5.37 & 0.001 & 0.001 \\
1.00 & 1.08 & 0.05 & 0.01799 & 0.21 & 0.48 & 1.81 & 3.00 & 3.28 & 0.002 & 0.004 \\
1.00 & 1.08 & 0.10 & 0.00004 & 0.03 & 0.10 & 0.73 & 1.60 & 1.84 & 0.005 & 0.008 \\
1.00 & 1.25 & -0.10 & 9.59483 & 10.03 & 10.50 & 12.50 & 14.18 & 14.54 & 0.001 & 0.001 \\
1.00 & 1.25 & -0.05 & 5.27416 & 6.03 & 6.79 & 9.39 & 11.37 & 11.79 & 0.001 & 0.001 \\
1.00 & 1.25 & 0.00 & 0.83495 & 2.57 & 3.63 & 6.68 & 8.85 & 9.32 & 0.001 & 0.002 \\
1.00 & 1.25 & 0.05 & 0.08667 & 0.78 & 1.57 & 4.46 & 6.67 & 7.16 & 0.002 & 0.002 \\
1.00 & 1.25 & 0.10 & 0.00757 & 0.23 & 0.61 & 2.80 & 4.86 & 5.35 & 0.003 & 0.004 \\
\bottomrule
\end{tabular}
\label{tab:forward_MC_summary_stats}
\end{table}
\begin{table}
\caption{Summary statistics for the dataset of barrier MC prices.}
\begin{tabular}{rrr@{\hspace{1.0em}}rrrrrr@{\hspace{1.0em}}rr}
\toprule
\multicolumn{3}{c}{} & \multicolumn{6}{c}{$\widehat{P}^{\mathrm{MC}}$} & \multicolumn{2}{c}{$\mathrm{CV}$} \\
\cmidrule(lr){4-9}\cmidrule(lr){10-11}
$T$ & $k_K$ & $k_B$ & $\textrm{min}$ & $q_{01}$ & $q_{05}$ & $q_{50}$ & $q_{95}$ & $q_{99}$ & $q_{95}$ & $q_{99}$ \\
\midrule
0.25 & 0.00 & -0.10 & 0.0390 & 0.07 & 0.08 & 0.16 & 0.63 & 0.91 & 0.005 & 0.005 \\
0.25 & 0.00 & -0.15 & 0.1739 & 0.28 & 0.34 & 0.64 & 1.56 & 1.90 & 0.003 & 0.003 \\
0.25 & 0.00 & -0.20 & 0.4599 & 0.73 & 0.86 & 1.44 & 2.53 & 2.81 & 0.002 & 0.003 \\
0.25 & -0.05 & -0.10 & 0.0036 & 0.01 & 0.01 & 0.02 & 0.07 & 0.11 & 0.009 & 0.010 \\
0.25 & -0.05 & -0.15 & 0.0409 & 0.07 & 0.09 & 0.18 & 0.44 & 0.54 & 0.004 & 0.005 \\
0.25 & -0.05 & -0.20 & 0.1493 & 0.28 & 0.34 & 0.58 & 1.02 & 1.14 & 0.003 & 0.003 \\
0.50 & 0.00 & -0.10 & 0.0142 & 0.02 & 0.03 & 0.06 & 0.29 & 0.53 & 0.007 & 0.008 \\
0.50 & 0.00 & -0.15 & 0.0688 & 0.12 & 0.14 & 0.28 & 0.94 & 1.37 & 0.004 & 0.005 \\
0.50 & 0.00 & -0.20 & 0.2012 & 0.33 & 0.39 & 0.74 & 1.85 & 2.34 & 0.003 & 0.004 \\
0.50 & -0.05 & -0.10 & 0.0014 & 0.00 & 0.00 & 0.01 & 0.03 & 0.06 & 0.014 & 0.016 \\
0.50 & -0.05 & -0.15 & 0.0165 & 0.03 & 0.04 & 0.08 & 0.26 & 0.38 & 0.006 & 0.007 \\
0.50 & -0.05 & -0.20 & 0.0706 & 0.13 & 0.16 & 0.30 & 0.74 & 0.95 & 0.004 & 0.005 \\
1.00 & 0.00 & -0.10 & 0.0049 & 0.01 & 0.01 & 0.02 & 0.12 & 0.26 & 0.012 & 0.014 \\
1.00 & 0.00 & -0.15 & 0.0255 & 0.04 & 0.05 & 0.11 & 0.46 & 0.82 & 0.007 & 0.008 \\
1.00 & 0.00 & -0.20 & 0.0785 & 0.13 & 0.16 & 0.32 & 1.07 & 1.63 & 0.005 & 0.005 \\
1.00 & -0.05 & -0.10 & 0.0005 & 0.00 & 0.00 & 0.00 & 0.01 & 0.03 & 0.023 & 0.026 \\
1.00 & -0.05 & -0.15 & 0.0063 & 0.01 & 0.01 & 0.03 & 0.12 & 0.23 & 0.010 & 0.012 \\
1.00 & -0.05 & -0.20 & 0.0280 & 0.05 & 0.06 & 0.13 & 0.43 & 0.65 & 0.006 & 0.007 \\
\bottomrule
\end{tabular}
\label{tab:barrier_MC_summary_stats}
\end{table}
\begin{table}
    \caption{Summary statistics for the dataset of realised-variance MC prices; quantiles and the coefficient of variations (CV) are only computed for MC prices above $10^{-5}$}
\begin{tabular}{rrr@{\hspace{1.0em}}rrrrrrr@{\hspace{1.0em}}rr}
\toprule
\multicolumn{3}{c}{} & \multicolumn{7}{c}{$\widehat{P}^{\mathrm{MC}}$} & \multicolumn{2}{c}{$\mathrm{CV}$} \\
\cmidrule(lr){4-10}\cmidrule(lr){11-12}
$T$ & $c$ & $K$ & $=0\,(\%)$ & $<10^{-5}\,(\%)$ & $q_{01}$ & $q_{05}$ & $q_{50}$ & $q_{95}$ & $q_{99}$ & $q_{95}$ & $q_{99}$ \\
\midrule
0.08 & 0.25 & 0.03 & 0.10 & 0.43 & 0.0005 & 0.006 & 0.10 & 0.20 & 0.21 & 0.003 & 0.006 \\
0.08 & 0.45 & 0.06 & 1.13 & 2.88 & 0.0001 & 0.001 & 0.07 & 0.17 & 0.18 & 0.006 & 0.020 \\
0.25 & 0.25 & 0.03 & 0.01 & 0.13 & 0.0016 & 0.010 & 0.10 & 0.19 & 0.21 & 0.002 & 0.005 \\
0.25 & 0.45 & 0.06 & 0.39 & 1.27 & 0.0001 & 0.002 & 0.07 & 0.16 & 0.18 & 0.005 & 0.014 \\
0.50 & 0.25 & 0.03 & 0.00 & 0.06 & 0.0027 & 0.013 & 0.10 & 0.19 & 0.20 & 0.002 & 0.004 \\
0.50 & 0.45 & 0.06 & 0.20 & 0.78 & 0.0002 & 0.003 & 0.07 & 0.16 & 0.18 & 0.004 & 0.011 \\
0.50 & 0.65 & 0.08 & 0.92 & 2.58 & 0.0001 & 0.001 & 0.05 & 0.14 & 0.15 & 0.008 & 0.023 \\
1.00 & 0.25 & 0.03 & 0.00 & 0.04 & 0.0035 & 0.015 & 0.10 & 0.18 & 0.20 & 0.002 & 0.004 \\
1.00 & 0.45 & 0.06 & 0.14 & 0.58 & 0.0003 & 0.005 & 0.07 & 0.16 & 0.18 & 0.004 & 0.010 \\
1.00 & 0.65 & 0.08 & 0.69 & 2.03 & 0.0001 & 0.001 & 0.05 & 0.13 & 0.15 & 0.007 & 0.021 \\
\bottomrule
\end{tabular}
\label{tab:rv_MC_diagnostics}
\end{table}

\clearpage
\newpage
\section{Extended uncertainty quantification results}
\label{app_Extended_UQ}

This appendix reports the contract-level counterparts of the pooled
uncertainty-quantification summaries in
Section~\ref{sec:exo_predictive_intervals}.  The main text pools results over
contract specifications to communicate the overall economic implications of
the posterior-predictive construction.  The figures and tables below document
the resulting heterogeneity across maturities, moneyness or strike, barrier
location, and forward-start dates.  All results use the retained contract set.

\subsection{Contract-level plug-in pricing errors}

\begin{figure}[!htbp]
    \centering

    \begin{subfigure}[t]{0.92\linewidth}
        \centering
        \includegraphics[width=\linewidth]{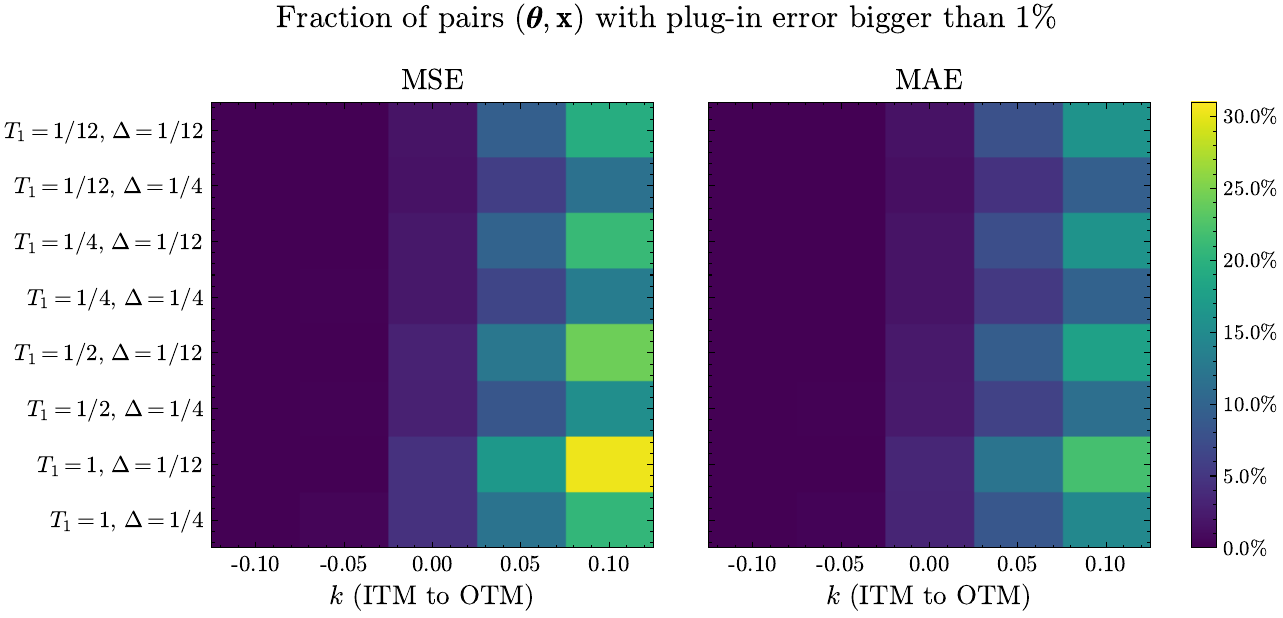}
        \caption{Forward-start}
    \end{subfigure}

    \vspace{0.4em}

    \begin{subfigure}[t]{0.92\linewidth}
        \centering
        \includegraphics[width=\linewidth]{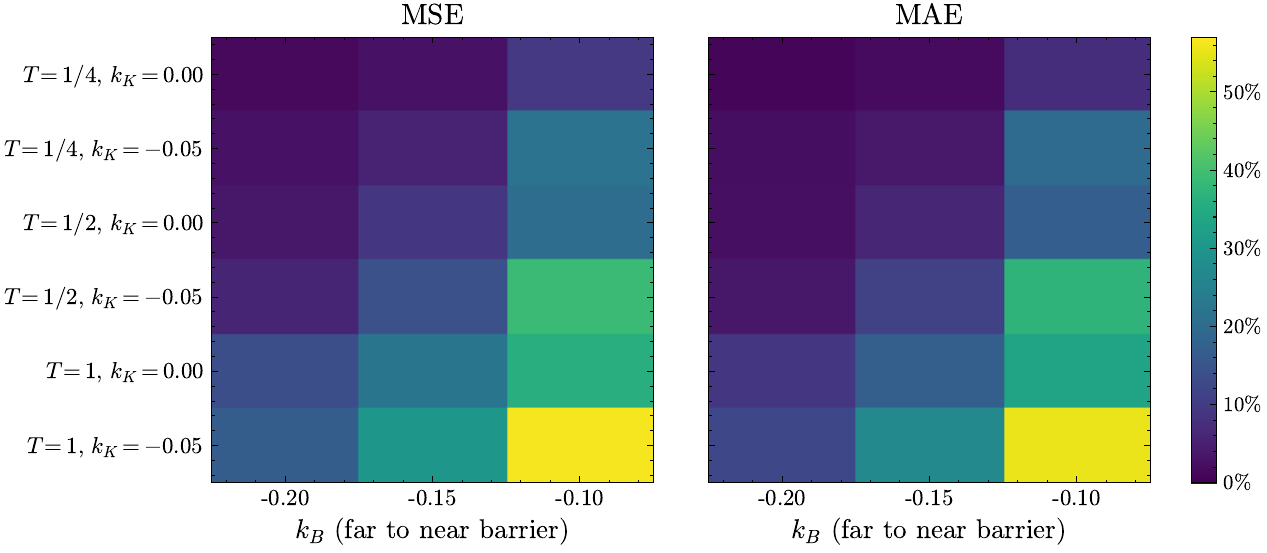}
        \caption{Barrier}
    \end{subfigure}

    \vspace{0.4em}

    \begin{subfigure}[t]{0.92\linewidth}
        \centering
        \includegraphics[width=\linewidth]{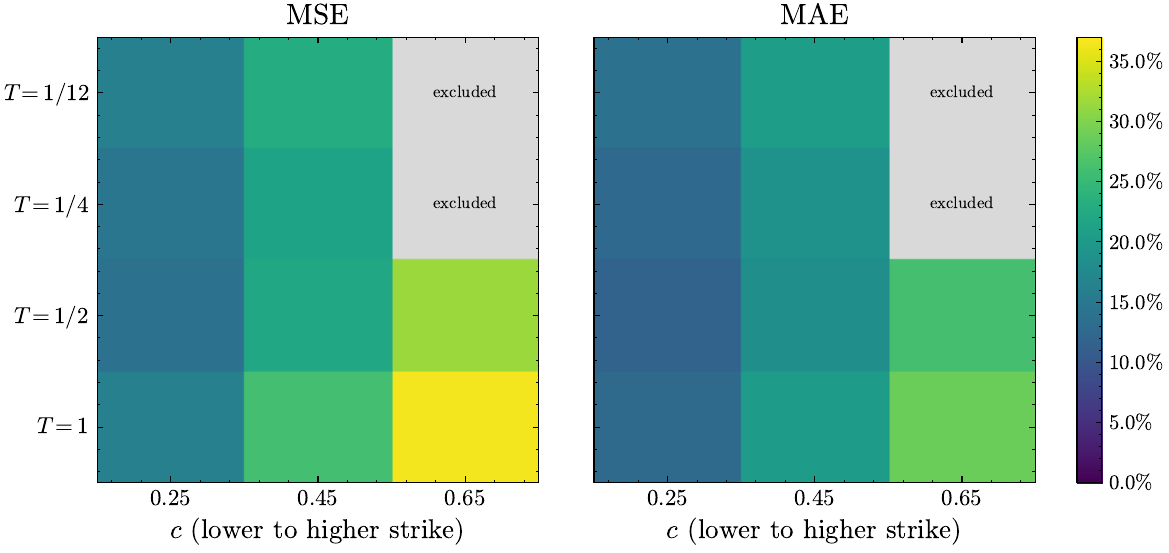}
        \caption{Realized variance}
    \end{subfigure}

\caption{As Figure~\ref{fig:plugin_error_3pct} in the main text, for the $1\%$ relative
plug-in-pricing-error threshold.
}\label{fig:plugin_error_1pct}
\end{figure}

\begin{figure}[!htbp]
    \centering

    \begin{subfigure}[t]{0.92\linewidth}
        \centering
        \includegraphics[width=\linewidth]{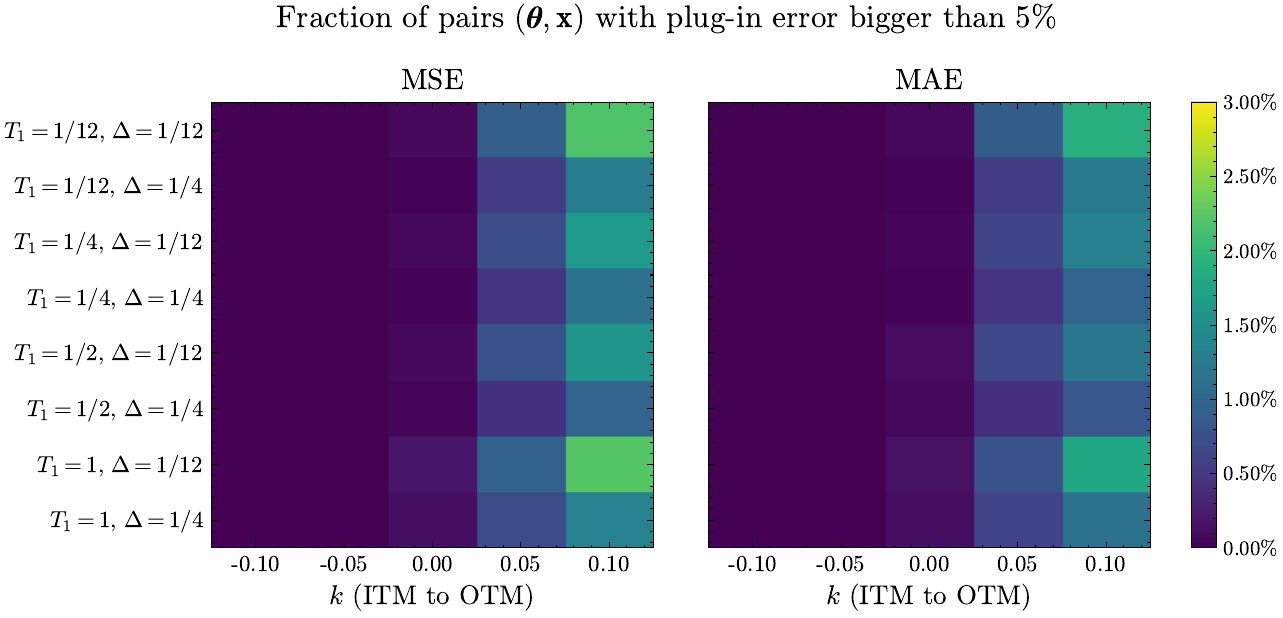}
        \caption{Forward-start}
    \end{subfigure}

    \vspace{0.4em}

    \begin{subfigure}[t]{0.92\linewidth}
        \centering
        \includegraphics[width=\linewidth]{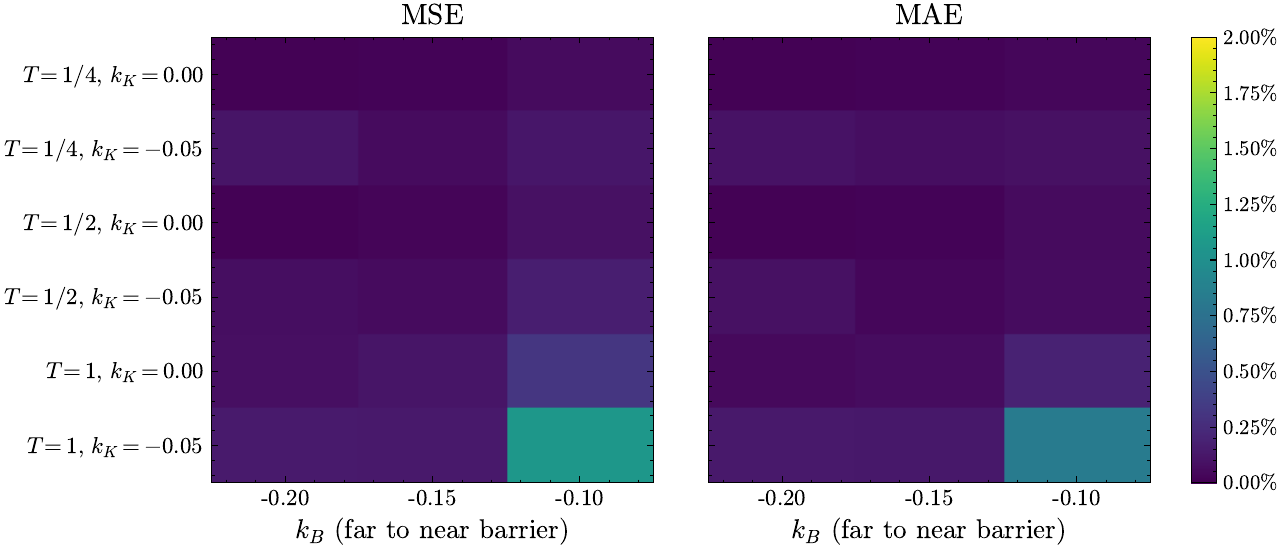}
        \caption{Barrier}
    \end{subfigure}

    \vspace{0.4em}

    \begin{subfigure}[t]{0.92\linewidth}
        \centering
        \includegraphics[width=\linewidth]{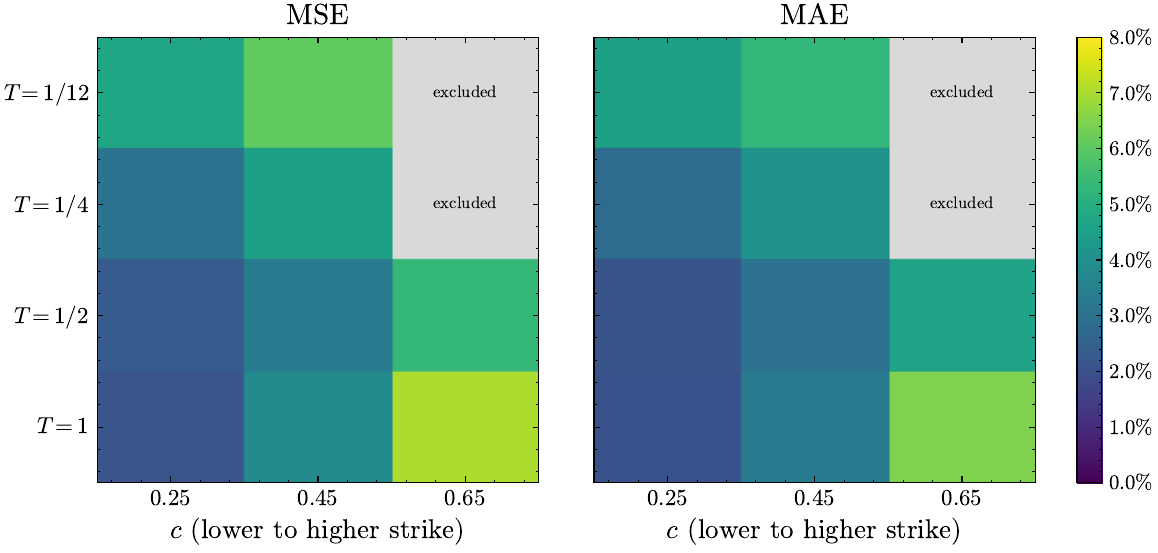}
        \caption{Realized variance}
    \end{subfigure}

\caption{As Figure~\ref{fig:plugin_error_3pct} in the main text, for the $5\%$ relative
plug-in-pricing-error threshold.
}\label{fig:plugin_error_5pct}
\end{figure}

Figures~\ref{fig:plugin_error_1pct} and~\ref{fig:plugin_error_5pct}
complement Figure~\ref{fig:plugin_error_3pct} by reporting the frequencies of
relative plug-in errors above $1\%$ and $5\%$, respectively. They make explicit that the pooled results in Table~\ref{tab:plugin_summary} do not arise uniformly across the contract grids.  The errors are most prevalent for
out-of-the-money forward starts, for barriers close to the prevailing price
region, and for the higher-strike realized-variance claims retained in the
experiment.  The tighter $1\%$ threshold is useful for identifying where a
point calibration can be economically imprecise even when average parameter
recovery appears strong. The $5\%$ results show that the issue is not
confined to negligible deviations.

\subsection{Relative widths of full predictive intervals}

Tables~\ref{tab:forward-width}--\ref{tab:rv-width} report, for each retained
contract specification, the median and upper quantiles of the relative width
$W_{ij}^{(0.95)}$ defined in \eqref{eq:relative-pi-width}.  The widths are
expressed as percentages of the held-out Monte Carlo reference price.  These
tables explain the heterogeneity behind the pooled values in
Table~\ref{tab:pi-width-summary}.  In particular, relatively wide intervals
occur for out-of-the-money forward starts, near-barrier contracts, and several
higher-strike realized-variance claims. Thus the full predictive construction
is not equally consequential for every claim: it identifies the regions in
which a point quote has the weakest reliability interpretation.

\subsection{Coverage of posterior-only and full predictive intervals}

Tables~\ref{tab:forward-coverage}--\ref{tab:rv-coverage} report empirical
coverage of nominal-$95\%$ intervals for every retained contract
specification.  ``Posterior only'' propagates draws from the calibrated TRE
posterior through the conditional surrogate mean.  ``Full UQ'' additionally
uses the conditional surrogate-predictive distribution and is therefore the
relevant interval for a downstream quote.  The comparison shows most clearly
for barriers that posterior uncertainty alone need not provide adequate
predictive coverage.  Including conditional surrogate uncertainty restores
coverage close to, and in some specifications modestly above, the nominal
level.  The detailed results also reveal the residual contract-to-contract
variation that is necessarily hidden by the family-level medians and IQRs in
Table~\ref{tab:coverage_summary}.

\subsection{Predictive-variance attribution}
Tables~\ref{fig:forward_start_variance_decomposition}-~\ref{fig:rv_variance_decomposition} report the contract-level decompositions and complement Table~\ref{tab:var_decomposition_summary}.
The detailed variance decompositions report the shares of full predictive
variance attributable to the two terms in
\eqref{eq:total-variance-decomposition}.  The decomposition is descriptive:
it attributes uncertainty under the fitted posterior-predictive model and
should not be read as a causal attribution.  Realized-variance claims are
predominantly posterior-driven, whereas for barriers the conditional
surrogate component is often comparable to, or larger than, the posterior
component.  Forward-start claims are generally posterior-driven away from the
most in-the-money specifications, although their surrogate contribution can
remain material.  Hence propagating posterior uncertainty alone is especially
inadequate for barriers, while retaining only surrogate uncertainty would
miss the principal contribution for most realized-variance claims.

\begin{table}
\caption{Forward-start options: median, $90\%$ and $95\%$ quantiles of the ratio of the length of the $95\%$ full predictive intervals to the held-out MC prices. All reported widths are percentages of the reference price.}\label{tab:forward-width}
    \begin{tabular}{rrrrrr}
\toprule
T1 & T2 & k & $95\%$ PI median & $95\%$ PI q90 & $95\%$ PI q95 \\
\midrule
0.08 & 0.17 & -0.10 & 0.14 & 0.21 & 0.24 \\
0.08 & 0.17 & -0.05 & 0.28 & 0.43 & 0.50 \\
0.08 & 0.17 & 0.00 & 0.56 & 1.34 & 1.84 \\
0.08 & 0.17 & 0.05 & 1.02 & 3.54 & 5.54 \\
0.08 & 0.17 & 0.10 & 1.70 & 6.59 & 10.50 \\
0.08 & 0.33 & -0.10 & 0.27 & 0.40 & 0.45 \\
0.08 & 0.33 & -0.05 & 0.39 & 0.69 & 0.83 \\
0.08 & 0.33 & 0.00 & 0.58 & 1.30 & 1.79 \\
0.08 & 0.33 & 0.05 & 0.84 & 2.39 & 3.63 \\
0.08 & 0.33 & 0.10 & 1.18 & 3.95 & 6.29 \\
0.25 & 0.33 & -0.10 & 0.15 & 0.23 & 0.26 \\
0.25 & 0.33 & -0.05 & 0.31 & 0.46 & 0.53 \\
0.25 & 0.33 & 0.00 & 0.62 & 1.37 & 1.90 \\
0.25 & 0.33 & 0.05 & 1.15 & 3.29 & 4.88 \\
0.25 & 0.33 & 0.10 & 1.86 & 5.66 & 8.30 \\
0.25 & 0.50 & -0.10 & 0.28 & 0.42 & 0.49 \\
0.25 & 0.50 & -0.05 & 0.43 & 0.73 & 0.87 \\
0.25 & 0.50 & 0.00 & 0.64 & 1.36 & 1.88 \\
0.25 & 0.50 & 0.05 & 0.93 & 2.46 & 3.69 \\
0.25 & 0.50 & 0.10 & 1.31 & 3.82 & 5.89 \\
0.50 & 0.58 & -0.10 & 0.18 & 0.25 & 0.28 \\
0.50 & 0.58 & -0.05 & 0.33 & 0.50 & 0.57 \\
0.50 & 0.58 & 0.00 & 0.71 & 1.51 & 2.09 \\
0.50 & 0.58 & 0.05 & 1.31 & 3.45 & 5.04 \\
0.50 & 0.58 & 0.10 & 2.07 & 5.54 & 7.98 \\
0.50 & 0.75 & -0.10 & 0.30 & 0.45 & 0.51 \\
0.50 & 0.75 & -0.05 & 0.48 & 0.79 & 0.95 \\
0.50 & 0.75 & 0.00 & 0.72 & 1.50 & 2.07 \\
0.50 & 0.75 & 0.05 & 1.05 & 2.68 & 4.03 \\
0.50 & 0.75 & 0.10 & 1.47 & 4.08 & 6.09 \\
1.00 & 1.08 & -0.10 & 0.18 & 0.28 & 0.32 \\
1.00 & 1.08 & -0.05 & 0.38 & 0.58 & 0.66 \\
1.00 & 1.08 & 0.00 & 0.82 & 1.85 & 2.62 \\
1.00 & 1.08 & 0.05 & 1.51 & 4.08 & 5.86 \\
1.00 & 1.08 & 0.10 & 2.38 & 6.40 & 9.29 \\
1.00 & 1.25 & -0.10 & 0.34 & 0.52 & 0.59 \\
1.00 & 1.25 & -0.05 & 0.54 & 0.92 & 1.10 \\
1.00 & 1.25 & 0.00 & 0.82 & 1.84 & 2.56 \\
1.00 & 1.25 & 0.05 & 1.21 & 3.26 & 4.91 \\
1.00 & 1.25 & 0.10 & 1.68 & 4.86 & 7.32 \\
\bottomrule
\end{tabular}
\end{table}
\begin{table}
\caption{Barrier options: median, $90\%$ and $95\%$ quantiles of the ratio of the length of the $95\%$ predictive intervals to the held-out MC prices. All reported widths are percentages of the reference price.}\label{tab:barrier-width}
    \begin{tabular}{rrrrrr}
\toprule
T & $k_K$ & $k_B$ & $95\%$ PI median & $95\%$ PI q90 & $95\%$ PI q95 \\
\midrule
0.25 & -0.05 & -0.20 & 1.15 & 1.75 & 2.22 \\
0.25 & -0.05 & -0.15 & 1.60 & 2.20 & 2.64 \\
0.25 & -0.05 & -0.10 & 2.86 & 3.73 & 4.20 \\
0.25 & 0.00 & -0.20 & 0.99 & 1.46 & 1.73 \\
0.25 & 0.00 & -0.15 & 1.29 & 1.82 & 2.13 \\
0.25 & 0.00 & -0.10 & 1.87 & 2.64 & 3.07 \\
0.50 & -0.05 & -0.20 & 1.60 & 2.23 & 2.63 \\
0.50 & -0.05 & -0.15 & 2.26 & 2.95 & 3.33 \\
0.50 & -0.05 & -0.10 & 4.23 & 5.51 & 5.98 \\
0.50 & 0.00 & -0.20 & 1.41 & 1.96 & 2.24 \\
0.50 & 0.00 & -0.15 & 1.83 & 2.54 & 2.92 \\
0.50 & 0.00 & -0.10 & 2.65 & 3.58 & 4.07 \\
1.00 & -0.05 & -0.20 & 2.33 & 3.13 & 3.63 \\
1.00 & -0.05 & -0.15 & 3.30 & 4.28 & 4.76 \\
1.00 & -0.05 & -0.10 & 6.49 & 8.73 & 9.40 \\
1.00 & 0.00 & -0.20 & 2.02 & 2.89 & 3.38 \\
1.00 & 0.00 & -0.15 & 2.59 & 3.56 & 4.14 \\
1.00 & 0.00 & -0.10 & 3.86 & 5.09 & 5.73 \\
\bottomrule
\end{tabular}
\end{table}
\begin{table}
\caption{Realized-variance options: median, $90\%$ and $95\%$ quantiles of the ratio of the length of the $95\%$ predictive intervals to the held-out MC prices. All reported widths are percentages of the reference price.}\label{tab:rv-width}

\begin{tabular}{rrrrrr}
\toprule
T & c & K & $95\%$ PI median & $95\%$ PI q90 & $95\%$ PI q95 \\
\midrule
0.08 & 0.25 & 0.03 & 1.19 & 7.32 & 23.04 \\
0.08 & 0.45 & 0.06 & 1.54 & 13.56 & 27.52 \\
0.25 & 0.25 & 0.03 & 1.22 & 5.40 & 13.03 \\
0.25 & 0.45 & 0.06 & 1.60 & 9.48 & 20.70 \\
0.50 & 0.25 & 0.03 & 1.31 & 4.57 & 8.08 \\
0.50 & 0.45 & 0.06 & 1.73 & 7.70 & 14.49 \\
0.50 & 0.65 & 0.08 & 2.35 & 11.65 & 22.68 \\
1.00 & 0.25 & 0.03 & 1.49 & 4.56 & 8.11 \\
1.00 & 0.45 & 0.06 & 1.97 & 7.37 & 13.60 \\
1.00 & 0.65 & 0.08 & 2.68 & 12.57 & 41.10 \\
\bottomrule
\end{tabular}
\end{table}
\clearpage
\begin{table}
\caption{Empirical coverage rates for nominal-$95\%$ predictive intervals for forward-start options. All entries as explained in percentages. These results complement Table~\ref{tab:coverage_summary}.}\label{tab:forward-coverage}
\begin{tabular}{rrrrrr}
\toprule
$T_1$ &  $T_2$ & $k$ & Posterior only $95\%$ & Full UQ $95\%$ \\
\midrule
0.08 &  0.17 & -0.10 & 87.66 & 97.39 \\
0.08 &  0.17 & -0.05 & 92.95 & 97.26 \\
0.08 &  0.17 & 0.00 & 93.84 & 96.62 \\
0.08 &  0.17 & 0.05 & 93.87 & 96.87 \\
0.08 &  0.17 & 0.10 & 93.80 & 97.01 \\
0.08 &  0.33 & -0.10 & 92.95 & 97.50 \\
0.08 &  0.33 & -0.05 & 93.99 & 97.06 \\
0.08 & 0.33 & 0.00 & 94.01 & 96.70 \\
0.08 & 0.33 & 0.05 & 94.76 & 97.18 \\
0.08 &  0.33 & 0.10 & 94.42 & 97.17 \\
0.25 &  0.33 & -0.10 & 87.80 & 97.16 \\
0.25 & 0.33 & -0.05 & 93.60 & 97.16 \\
0.25 & 0.33 & 0.00 & 95.25 & 97.27 \\
0.25 & 0.33 & 0.05 & 94.86 & 97.06 \\
0.25 &  0.33 & 0.10 & 94.30 & 97.16 \\
0.25 &  0.50 & -0.10 & 93.52 & 97.44 \\
0.25 & 0.50 & -0.05 & 95.42 & 97.52 \\
0.25 &  0.50 & 0.00 & 95.84 & 97.42 \\
0.25 &  0.50 & 0.05 & 95.59 & 97.37 \\
0.25 &  0.50 & 0.10 & 95.22 & 97.36 \\
0.50 &  0.58 & -0.10 & 89.10 & 98.28 \\
0.50 &  0.58 & -0.05 & 94.50 & 97.17 \\
0.50 &  0.58 & 0.00 & 95.62 & 97.24 \\
0.50 &  0.58 & 0.05 & 94.69 & 96.88 \\
0.50 &  0.58 & 0.10 & 94.24 & 96.82 \\
0.50 &  0.75 & -0.10 & 94.00 & 97.14 \\
0.50 &  0.75 & -0.05 & 95.28 & 97.36 \\
0.50 &  0.75 & 0.00 & 95.60 & 97.29 \\
0.50 & 0.75 & 0.05 & 95.43 & 96.98 \\
0.50 & 0.75 & 0.10 & 95.41 & 97.20 \\
1.00 & 1.08 & -0.10 & 89.48 & 97.76 \\
1.00 & 1.08 & -0.05 & 94.71 & 97.22 \\
1.00 &  1.08 & 0.00 & 95.82 & 97.29 \\
1.00 & 1.08 & 0.05 & 94.98 & 96.78 \\
1.00 & 1.08 & 0.10 & 93.84 & 96.78 \\
1.00 &  1.25 & -0.10 & 94.23 & 97.31 \\
1.00 &  1.25 & -0.05 & 95.71 & 97.52 \\
1.00 &  1.25 & 0.00 & 95.77 & 97.22 \\
1.00 &  1.25 & 0.05 & 95.48 & 97.04 \\
1.00 & 1.25 & 0.10 & 95.28 & 96.99 \\
\bottomrule
\end{tabular}

\label{table:forward_coverage}
\end{table}
\begin{table}
    \caption{Empirical coverage rates for nominal-$95\%$ predictive intervals for barrier options. All entries as explained in percentages. These results complement Table~\ref{tab:coverage_summary}.}\label{tab:barrier-coverage}
    \begin{tabular}{rrrrr}
\toprule
T & $k_K$ & $k_B$ & Posterior only $95\%$ & Full UQ $95\%$ \\
\midrule
0.25 & -0.05 & -0.20 & 82.92 & 96.17 \\
0.25 & -0.05 & -0.15 & 79.34 & 95.70 \\
0.25 & -0.05 & -0.10 & 63.54 & 94.72 \\
0.25 & 0.00 & -0.20 & 87.08 & 96.56 \\
0.25 & 0.00 & -0.15 & 86.06 & 95.78 \\
0.25 & 0.00 & -0.10 & 80.14 & 95.01 \\
0.50 & -0.05 & -0.20 & 83.20 & 96.19 \\
0.50 & -0.05 & -0.15 & 75.06 & 95.88 \\
0.50 & -0.05 & -0.10 & 53.94 & 94.96 \\
0.50 & 0.00 & -0.20 & 87.94 & 96.40 \\
0.50 & 0.00 & -0.15 & 83.93 & 96.18 \\
0.50 & 0.00 & -0.10 & 73.69 & 95.54 \\
1.00 & -0.05 & -0.20 & 78.31 & 95.76 \\
1.00 & -0.05 & -0.15 & 66.32 & 95.14 \\
1.00 & -0.05 & -0.10 & 43.90 & 95.27 \\
1.00 & 0.00 & -0.20 & 84.29 & 96.34 \\
1.00 & 0.00 & -0.15 & 77.46 & 95.74 \\
1.00 & 0.00 & -0.10 & 63.64 & 95.28 \\
\bottomrule
\end{tabular}
\label{table:barrier_coverage}
\end{table}

\begin{table}
    \caption{Empirical coverage rates for nominal-$95\%$ predictive intervals for realized-variance options. All entries as explained in percentages. These results complement Table~\ref{tab:coverage_summary}.}\label{tab:rv-coverage}
    \begin{tabular}{rrrrr}
\toprule
T & c & K & Posterior only $95\%$ & Full UQ $95\%$ \\
\midrule
0.08 & 0.25 & 0.03 & 91.70 & 96.42 \\
0.08 & 0.45 & 0.06 & 93.24 & 96.29 \\
0.25 & 0.25 & 0.03 & 93.10 & 96.62 \\
0.25 & 0.45 & 0.06 & 94.16 & 96.73 \\
0.50 & 0.25 & 0.03 & 94.32 & 96.73 \\
0.50 & 0.45 & 0.06 & 95.11 & 97.09 \\
0.50 & 0.65 & 0.08 & 94.80 & 97.15 \\
1.00 & 0.25 & 0.03 & 94.77 & 97.05 \\
1.00 & 0.45 & 0.06 & 94.83 & 97.05 \\
1.00 & 0.65 & 0.08 & 91.59 & 97.34 \\
\bottomrule
\end{tabular}
\label{table:rv_coverage}
\end{table}

\clearpage
\begin{table}
    \caption{Percentage of variance attributed by the law of total variance to the posterior uncertainty and to the surrogate pricer for contracts in the forward start family. These results complement Table~\ref{tab:var_decomposition_summary}.}
    \label{fig:forward_start_variance_decomposition}
   \begin{tabular}{rrrrr}
\toprule
$T_1$ & $T_2$ & $k$ & TRE variance share (\%) & Surrogate variance share (\%) \\
\midrule
0.08 & 0.17 & -0.10 & 49.99 & 50.01 \\
0.08 & 0.17 & -0.05 & 69.29 & 30.71 \\
0.08 & 0.17 & 0.00 & 80.20 & 19.80 \\
0.08 & 0.17 & 0.05 & 80.18 & 19.82 \\
0.08 & 0.17 & 0.10 & 78.41 & 21.59 \\
0.08 & 0.33 & -0.10 & 64.66 & 35.34 \\
0.08 & 0.33 & -0.05 & 76.50 & 23.50 \\
0.08 & 0.33 & 0.00 & 80.36 & 19.64 \\
0.08 & 0.33 & 0.05 & 82.23 & 17.77 \\
0.08 & 0.33 & 0.10 & 83.13 & 16.87 \\
0.25 & 0.33 & -0.10 & 50.85 & 49.15 \\
0.25 & 0.33 & -0.05 & 71.90 & 28.10 \\
0.25 & 0.33 & 0.00 & 82.27 & 17.73 \\
0.25 & 0.33 & 0.05 & 83.43 & 16.57 \\
0.25 & 0.33 & 0.10 & 81.12 & 18.88 \\
0.25 & 0.50 & -0.10 & 69.18 & 30.82 \\
0.25 & 0.50 & -0.05 & 79.81 & 20.19 \\
0.25 & 0.50 & 0.00 & 82.96 & 17.04 \\
0.25 & 0.50 & 0.05 & 84.67 & 15.33 \\
0.25 & 0.50 & 0.10 & 84.96 & 15.04 \\
0.50 & 0.58 & -0.10 & 46.42 & 53.58 \\
0.50 & 0.58 & -0.05 & 76.98 & 23.02 \\
0.50 & 0.58 & 0.00 & 84.37 & 15.63 \\
0.50 & 0.58 & 0.05 & 85.02 & 14.98 \\
0.50 & 0.58 & 0.10 & 83.11 & 16.89 \\
0.50 & 0.75 & -0.10 & 74.80 & 25.20 \\
0.50 & 0.75 & -0.05 & 80.15 & 19.85 \\
0.50 & 0.75 & 0.00 & 83.73 & 16.27 \\
0.50 & 0.75 & 0.05 & 86.84 & 13.16 \\
0.50 & 0.75 & 0.10 & 86.76 & 13.24 \\
1.00 & 1.08 & -0.10 & 53.79 & 46.21 \\
1.00 & 1.08 & -0.05 & 78.33 & 21.67 \\
1.00 & 1.08 & 0.00 & 85.79 & 14.21 \\
1.00 & 1.08 & 0.05 & 87.01 & 12.99 \\
1.00 & 1.08 & 0.10 & 82.87 & 17.13 \\
1.00 & 1.25 & -0.10 & 76.22 & 23.78 \\
1.00 & 1.25 & -0.05 & 83.61 & 16.39 \\
1.00 & 1.25 & 0.00 & 85.92 & 14.08 \\
1.00 & 1.25 & 0.05 & 86.55 & 13.45 \\
1.00 & 1.25 & 0.10 & 87.40 & 12.60 \\
\bottomrule
\end{tabular}
\end{table}
\begin{table}
    \caption{Percentage of variance attributed by the law of total variance to the posterior uncertainty and to the surrogate pricer for contracts in the barrier family. These results complement Table~\ref{tab:var_decomposition_summary}.}
    \label{fig:barrier_variance_decomposition}
    \begin{tabular}{rrrrr}
\toprule
$T$ & $k_K$ & $k_B$ & TRE variance share (\%) & Surrogate variance share (\%) \\
\midrule
0.25 & 0.00 & -0.10 & 48.38 & 51.62 \\
0.25 & 0.00 & -0.15 & 56.33 & 43.67 \\
0.25 & 0.00 & -0.20 & 55.64 & 44.36 \\
0.25 & -0.05 & -0.10 & 27.84 & 72.16 \\
0.25 & -0.05 & -0.15 & 43.04 & 56.96 \\
0.25 & -0.05 & -0.20 & 47.09 & 52.91 \\
0.50 & 0.00 & -0.10 & 38.99 & 61.01 \\
0.50 & 0.00 & -0.15 & 51.73 & 48.27 \\
0.50 & 0.00 & -0.20 & 57.88 & 42.12 \\
0.50 & -0.05 & -0.10 & 20.75 & 79.25 \\
0.50 & -0.05 & -0.15 & 38.62 & 61.38 \\
0.50 & -0.05 & -0.20 & 47.56 & 52.44 \\
1.00 & 0.00 & -0.10 & 28.96 & 71.04 \\
1.00 & 0.00 & -0.15 & 43.09 & 56.91 \\
1.00 & 0.00 & -0.20 & 51.83 & 48.17 \\
1.00 & -0.05 & -0.10 & 14.08 & 85.92 \\
1.00 & -0.05 & -0.15 & 30.45 & 69.55 \\
1.00 & -0.05 & -0.20 & 42.21 & 57.79 \\
\bottomrule
\end{tabular}
\end{table}
\begin{table}
    \caption{Percentage of variance attributed by the law of total variance to the posterior uncertainty and to the surrogate pricer for contracts in the Realized variance family. These results complement Table~\ref{tab:var_decomposition_summary}.}\label{fig:rv_variance_decomposition}

    \begin{tabular}{rrrrr}
\toprule
T & c & K & TRE variance share (\%) & Surrogate variance share (\%) \\
\midrule
0.08 & 0.25 & 0.03 & 87.63 & 12.37 \\
0.08 & 0.45 & 0.06 & 86.40 & 13.60 \\
0.25 & 0.25 & 0.03 & 89.05 & 10.95 \\
0.25 & 0.45 & 0.06 & 89.84 & 10.16 \\
0.50 & 0.25 & 0.03 & 91.92 & 8.08 \\
0.50 & 0.45 & 0.06 & 91.63 & 8.37 \\
0.50 & 0.65 & 0.08 & 89.54 & 10.46 \\
1.00 & 0.25 & 0.03 & 91.89 & 8.11 \\
1.00 & 0.45 & 0.06 & 91.65 & 8.35 \\
1.00 & 0.65 & 0.08 & 86.01 & 13.99 \\
\bottomrule
\end{tabular}
\end{table}
\clearpage
\newpage
\section{Extended XAI analysis}\label{app:sec_XAI}

We complement the main posterior-information analysis with comparisons to KL-SHAP and point-estimator explanations. The point-estimator maps explain the outputs of separately
trained MSE- and MAE-based inverse networks, whereas Hellinger-SHAP and
KL-SHAP explain posterior contraction relative to the chosen prior.  These
are different inferential targets.  Thus, qualitative agreement between
point and posterior maps is evidence of convergent feature importance across
distinct fitted estimators, not equality of the underlying estimands.
Differences for individual parameters are correspondingly informative and
should not be regarded as inconsistencies by default.

\begin{figure}[!htbp]
    \centering

    \begin{subfigure}[t]{0.48\linewidth}
        \centering
        \includegraphics[width=\linewidth]{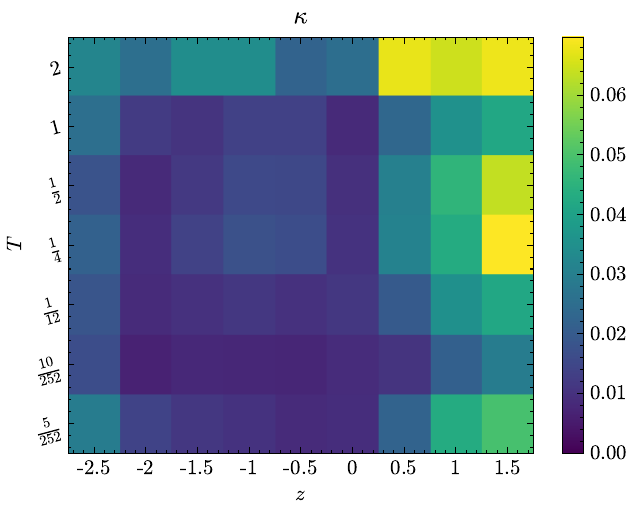}
    \end{subfigure}
    \hfill
    \begin{subfigure}[t]{0.48\linewidth}
        \centering
        \includegraphics[width=\linewidth]{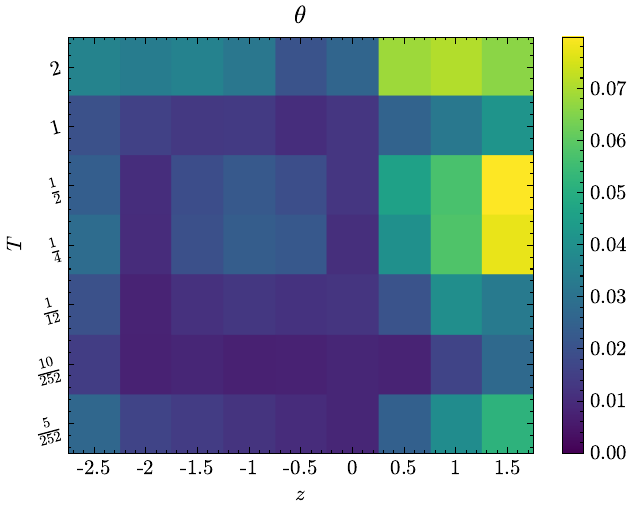}
    \end{subfigure}

    \vspace{0.5em}

    \begin{subfigure}[t]{0.48\linewidth}
        \centering
        \includegraphics[width=\linewidth]{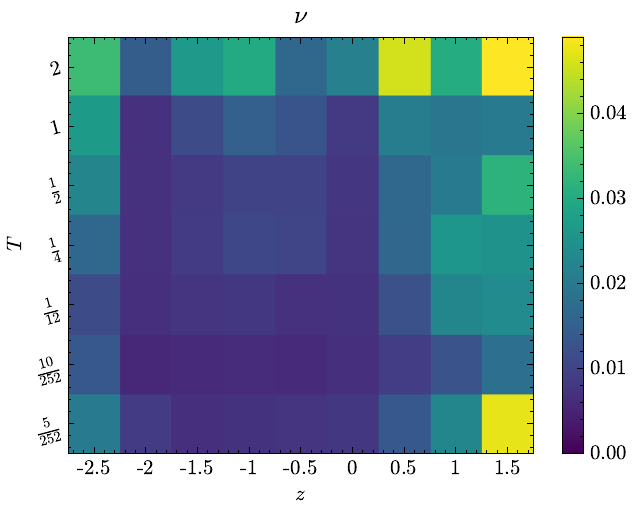}
    \end{subfigure}
    \hfill
    \begin{subfigure}[t]{0.48\linewidth}
        \centering
        \includegraphics[width=\linewidth]{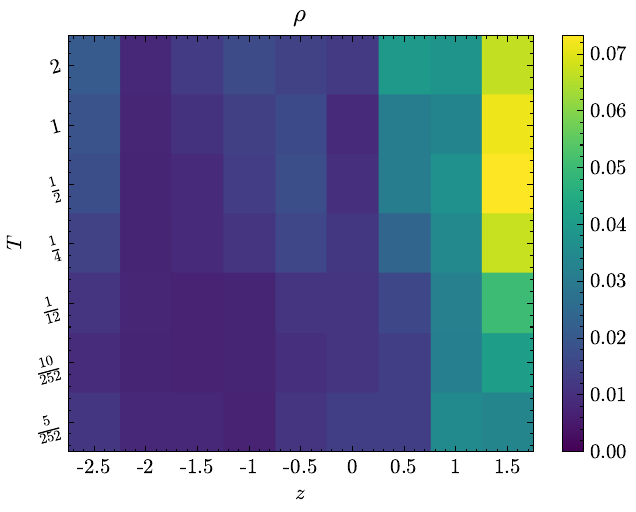}
    \end{subfigure}

    \vspace{0.5em}

    \begin{subfigure}[t]{0.48\linewidth}
        \centering
        \includegraphics[width=\linewidth]{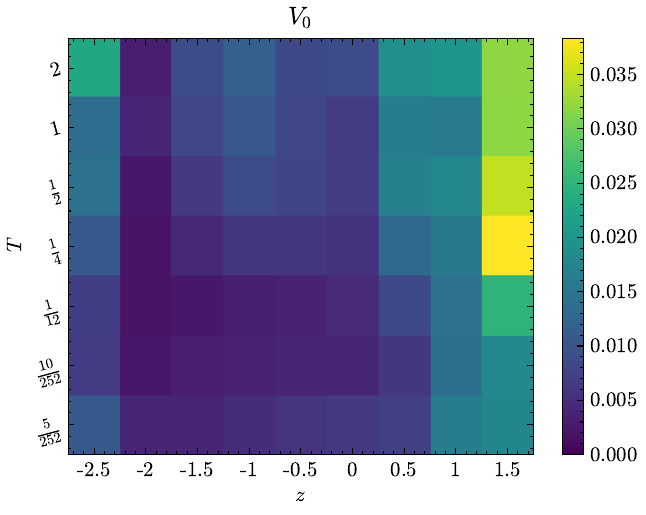}
    \end{subfigure}
    \hfill
    \begin{subfigure}[t]{0.48\linewidth}
        \centering
        \includegraphics[width=\linewidth]{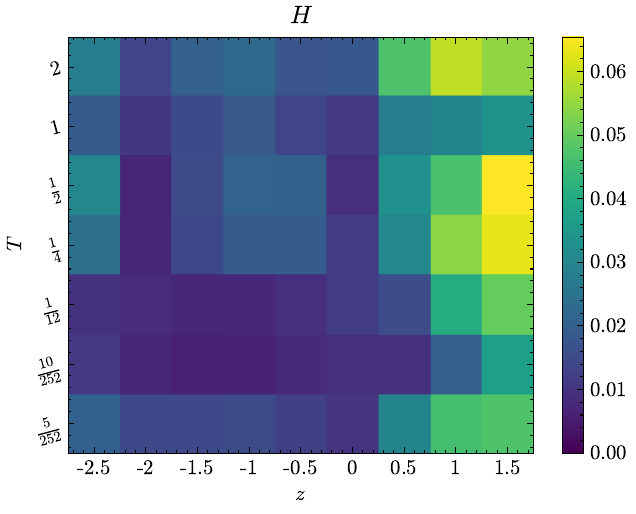}
    \end{subfigure}

    \caption{Mean absolute KL Kernel-SHAP attributions for each of the six rough Heston parameters, averaged over 500 explained held-out IV surfaces.}
    \label{fig:KL_XAI_heatmap}
\end{figure}

\begin{figure}[p]
    \centering
    \includegraphics[width=\textwidth]
    {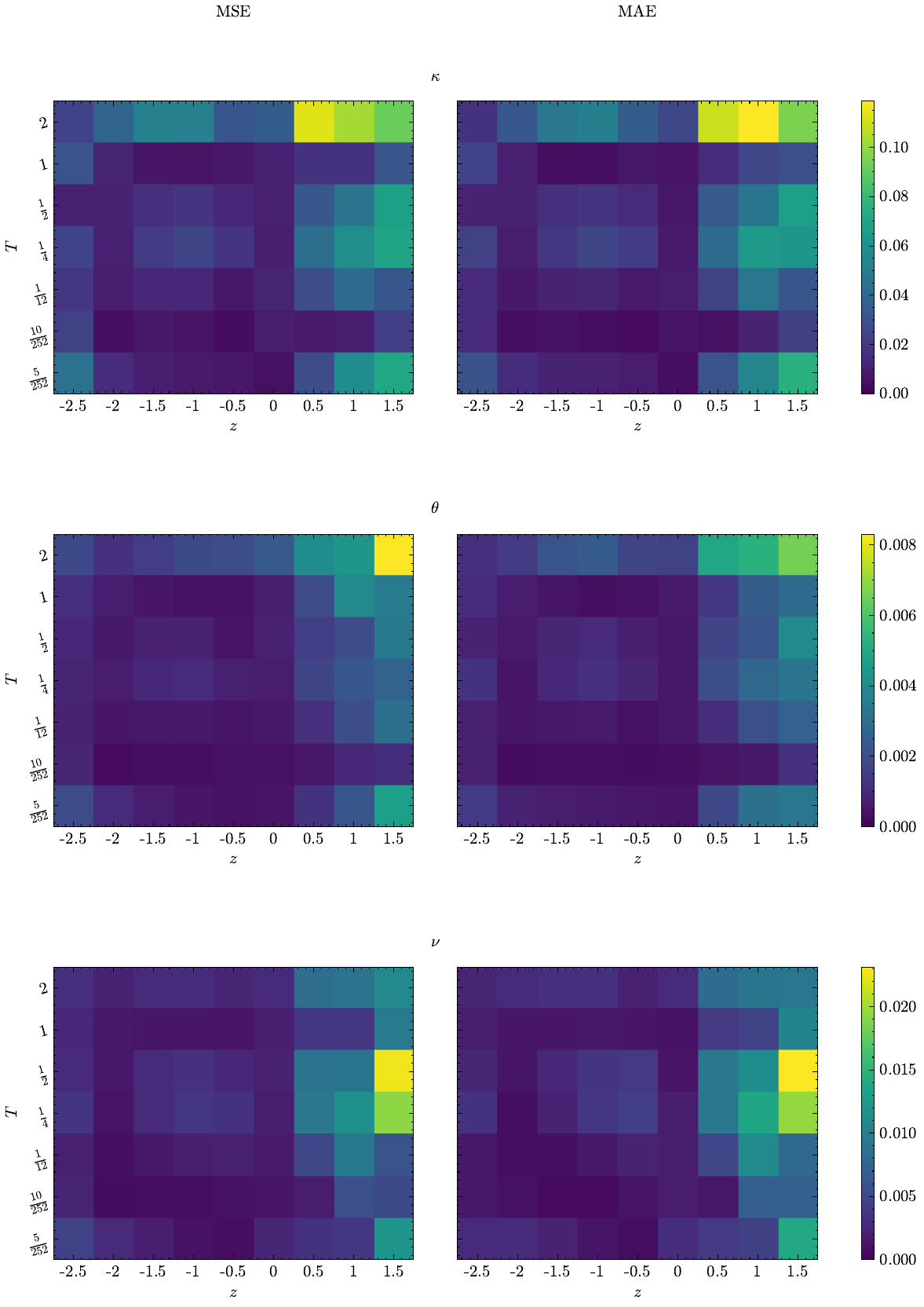}
        \caption{Mean absolute Kernel-SHAP attributions for MSE- and MAE-trained point estimator for the $\kappa, \theta, \nu$ rough Heston parameters. Results are averaged over 500 explained held-out IV surfaces. See Figure~\ref{fig:point_XAI_2} for results on the remaining three parameters $\rho,V_o,H$. These results complement the Hellinger-SHAP attributions from Figure~\ref{fig:KL_XAI_heatmap}.}
            \label{fig:point_XAI_1}
\end{figure}

\clearpage

\begin{figure}[p]
    \centering
    \includegraphics[width=\textwidth]
    {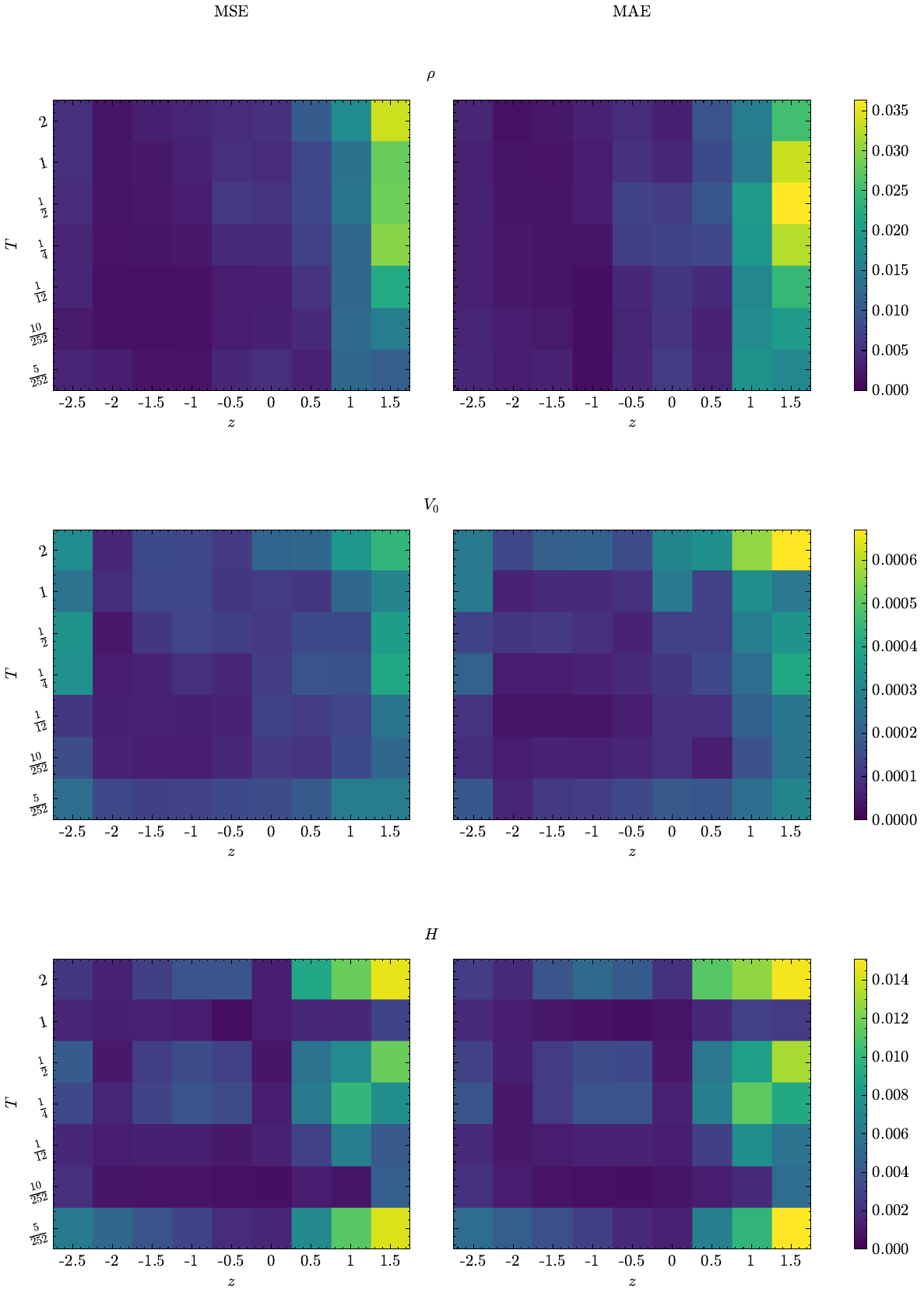}
    \caption{Mean absolute Kernel-SHAP attributions for MSE- and MAE-trained point estimator for the $\rho,V_o,H$ rough Heston parameters. Results are averaged over 500 explained held-out IV surfaces.  See Figure~\ref{fig:point_XAI_1} for results on the first three parameters $\kappa,\theta,\nu$.}
    \label{fig:point_XAI_2}
\end{figure}

\clearpage
\newpage
\section{Methodological considerations on the training of the neural estimators}
\label{app:further_training_details}
This appendix section is structured as follows:
\begin{itemize}
    \item Section~\ref{app:pre_processing} describes the preprocessing and standardization steps applied to the neural-network inputs.
    \item Section~\ref{app:architectures_and_loss_functions} details the network architectures and loss functions used for the neural point estimators, ratio estimators, and exotic-option surrogate pricers. The last part of Section~\ref{app:TRE_and_marginal_NRE} additionally describes the post-training calibration of the telescoping ratio estimator (TRE) and assesses the quality of the learnt posterior densities through simulation-based calibration (SBC) checks.
    \item Section~\ref{app:nn_hyperparameter_search} reports the hyperparameter grids used during neural network training.
\end{itemize}
\subsection{Data preprocessing and standardization procedures}\label{app:pre_processing}
We remind the reader that in this paper we consider two learning problems. In the inverse problem, we observe an IV surface $\x$ and learn either a point estimator $\x \to \widehat{\bt}(\x)$ or an approximation $ \x, \bt \to \hat{p}(\bt\mid\x)$ of the full posterior density, as described in Section~\ref{sec:sbi-landscape}. In addition, for the inverse problem, we require learning approximations $\x, \,\bt \to \hat{p}(\theta^i \mid \x)$ of the corresponding posterior marginals for $i=1,\ldots,m$ for the interpretability (XAI) analysis described in Section~\ref{sec:xai_hellinger}. In our case, $m=6$. In the forward problem, the direction is reversed: the input is the rough Heston parameter vector $\bt$ and the outputs are estimators $ \hat{P}^{\mathrm{exo}}$ of exotic-option prices. We apply preprocessing steps to both $\x$ and $\bt$ before passing them as inputs to the neural networks, as follows.

\textbf{$\x$ preprocessing:} The IV surface is a $63$-dimensional vector, see Section~\ref{sec:gridconst}. Let $\bar{\x}_{\mathrm{tr}}$ be the componentwise mean of the
IV surfaces in the training dataset and let $\widehat{\Sigma}_{\mathrm{tr}} = V\Lambda V^{\top}$ be the
eigendecomposition of their empirical covariance, with eigenvalues
$\lambda_1 \geq \cdots \geq \lambda_{63}$ on the diagonal of $\Lambda$ and the
corresponding orthonormal eigenvectors as the columns of $V$. Writing $V_{20}$ for the
first $20$ columns of $V$ and $\Lambda_{20} = \mathrm{diag}(\lambda_1,\ldots,\lambda_{20})$,
the network input is the vector of whitened scores
\begin{equation}
    \widetilde{\x}
    =
    \Lambda_{20}^{-1/2}V_{20}^{\top}
    (\x-\bar{\x}_{\mathrm{tr}})
    \in\mathbb{R}^{20},
    \label{eq:training_pca_whitening}
\end{equation}
which projects $\x$ onto the $20$ leading principal directions and rescales each score to
unit variance. The transform is fitted on the training split only and subsequently applied to the validation and test datasets. The retained components capture over $99.9999\%$ of the variability in the training dataset; the discarded directions correspond to near-zero eigenvalues, and retaining
them causes numerical instability of the whitening procedure in $\mathrm{float32}$. To ensure numerical stability, we apply
\eqref{eq:training_pca_whitening} in $\mathrm{float64}$ and subsequently train the networks in
$\mathrm{float32}$.

\paragraph{\bf{PCA representation for local XAI backgrounds.}}
The local-background Kernel-SHAP analyses in Section~\ref{sec:xai_hellinger}
use the same training-split PCA fit, but select nearest neighbours in the
unwhitened retained-score representation
\[
    \x \longmapsto V_{20}^{\top}(\x-\bar{\x}_{\mathrm{tr}}).
\]
For each explained surface, its $100$ nearest IV surfaces are selected from
the full $500000$-surface dataset using Euclidean distance in this
$20$-dimensional representation. Thus, locality is assessed in the dominant
principal directions without rescaling them to equal variance. The resulting
raw IV-surface backgrounds are subsequently supplied to Kernel SHAP for
feature-wise completion; every hybrid surface is then transformed according
to \eqref{eq:training_pca_whitening} before evaluation by the neural
estimators. 

\textbf{$\bt$ preprocessing} Let $[l_i,u_i]$ be the prior ranges from Table~\ref{tab:rheston_supports}, $i=1,\ldots,m$. Each component $\theta^i$ is standardized to $[-1,1]$, as follows
\begin{equation}
    \tilde\theta^i
    =2\frac{\theta^i-l_i}{u_i-l_i}-1
    =\frac{2 \theta_i - u_i - l_i }{u_i - l_i}.
    \label{eq:training_theta_scaling}
\end{equation}
In the schematic representation we discuss below, i.e., Figures~\ref{fig:NBE_multi_head_architecture}-~\ref{fig:marginal_NRE_multi_head_architecture}, we keep using $\x$ and $\bt$ instead of $\tilde{x}$ and $\tilde{\bt}$, for simplicity of notation. 

\subsection{Neural network multi-head architecture and loss functions}\label{app:architectures_and_loss_functions}
All networks use a multi-head architecture to alleviate multi-output optimization issues. A shared multi-layer perceptron (MLP), called the encoder, extracts a set of informative summary statistics $s(\cdot)$ from the input. This representation is then passed to separate, independently parameterized MLP networks, called the heads. The encoder therefore learns features that are informative for all outputs, while the parallel heads can perform output-specific finetuning. We discuss networks one at a time. 

All estimation methods we consider are amortized, which makes the considered approaches feasible for real-life applications. Specifically, each network is fitted once and then can be reused across many inputs at inference time. We train one network per estimator for the inverse problem and one surrogate pricer per exotic family for the forward exotic pricing problem. Nevertheless, the harder the inference task or the more tasks solved at once, e.g., the more contracts considered in one exotic family, the more likely it is that at least some of the learnt approximations are sub-optimal. The loss in performance relative to the true estimator is often called the \emph{amortization gap} \citep{cremer2018inference}. The preprocessing steps
described above already reduce this gap by presenting the networks with well-conditioned
inputs; further, all networks use the multi-head architecture to alleviate potential multi-output optimization
issues. For each estimator, we state the loss function used and report its corresponding values, as estimated on the validation set, at the end of each training epoch.

\subsubsection{Neural point estimators}

\begin{figure}[!t]
    \centering
    \includegraphics[width=8cm]{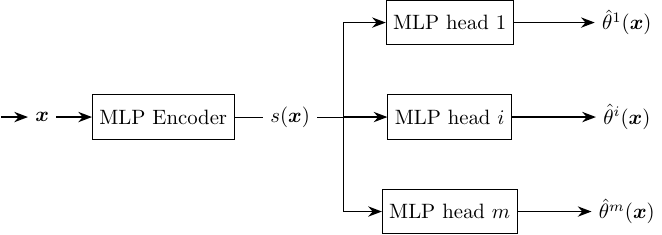}
    \caption{Schematic representation of the neural point estimator architecture. The observed IV surface $\x$ is passed to an MLP network to obtain a set of informative summary statistics, call them $s(\x)$. Next $s(\x)$ is passed as input to $m$ MLP heads to obtain estimates of $\theta^1,\ldots,\theta^m$.}
    \label{fig:NBE_multi_head_architecture}
\end{figure}

\begin{figure}[!t]
    \centering
    \begin{subfigure}[t]{0.48\linewidth}
        \centering
        \includegraphics[width=\linewidth]
        {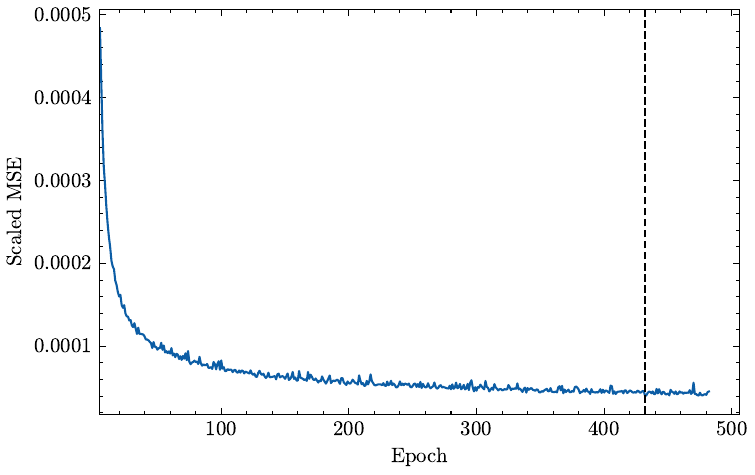}
    \end{subfigure}
    \hfill
    \begin{subfigure}[t]{0.48\linewidth}
        \centering
        \includegraphics[width=\linewidth]
        {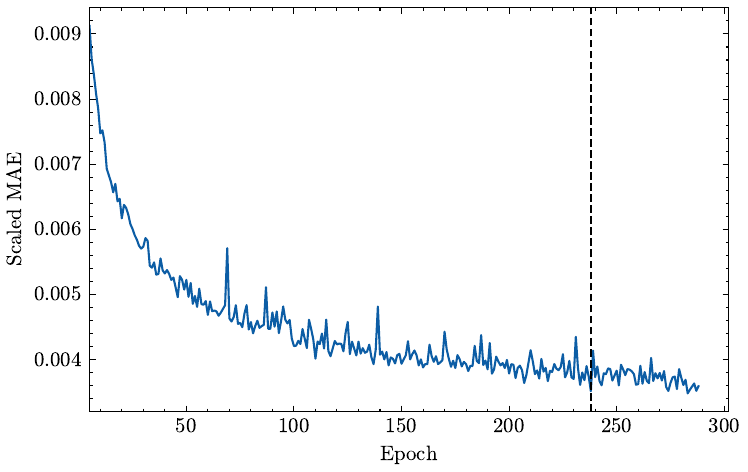}
    \end{subfigure}
    \caption{Validation losses for the neural point estimators trained with the scaled MSE (left) and MAE (right) losses. The epoch selected through early stopping is marked in black.}
    \label{fig:validation_losses_npe_scaled}
\end{figure}

\begin{figure}[!hbtp]
    \centering
    \begin{subfigure}[t]{0.48\linewidth}
        \centering
        \includegraphics[width=\linewidth]
        {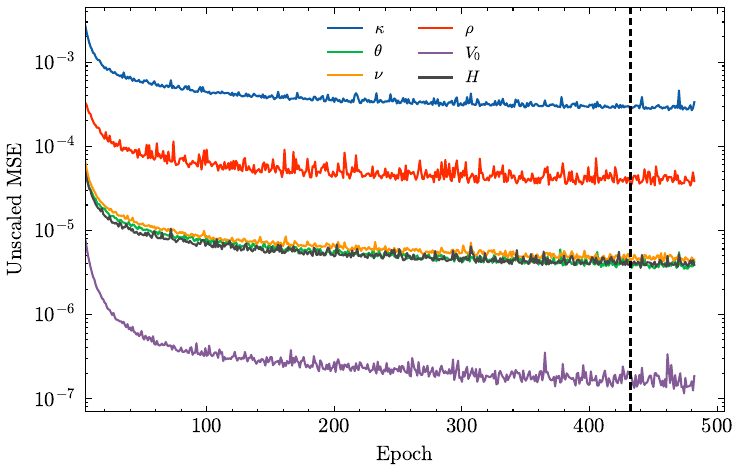}
    \end{subfigure}
    \hfill
    \begin{subfigure}[t]{0.48\linewidth}
        \centering
        \includegraphics[width=\linewidth]
        {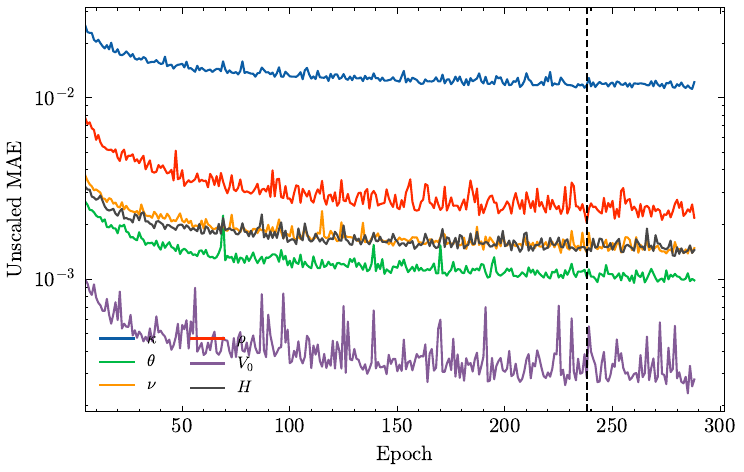}
    \end{subfigure}
    \caption{Breakdown of the validation losses on the original parameter scale, i.e., unscaled, for the two neural point estimators from Figure~\ref{fig:validation_losses_npe_scaled}. The $y$ axis is displayed on log scale.} 
    \label{fig:validation_losses_npe_unscaled}
\end{figure}

For the neural point estimators, let $\Delta_i=u_i-l_i$ denote the width of the prior support of the $i^\text{th}$ parameter for $i=1,\ldots,m$, see Table~\ref{tab:rheston_supports}. The two training objectives are then given by scaled MSE and MAE losses, as follows:
\begin{align}
    \mathcal L_{\mathrm{MSE}}
    &=
    \frac{1}{m}\sum_{i=1}^{m}
     \ev \left[
        \left(\frac{\hat\theta^i(\x)-\theta^i}{\Delta_i}\right)^2
    \right],
    \label{eq:npe_scaled_mse}\\
    \mathcal L_{\mathrm{MAE}}
    &=
    \frac{1}{m}\sum_{i=1}^{m}
    \ev\left[
        \frac{|\hat\theta^i(\x)-\theta^i|}{\Delta_i}
    \right].
    \label{eq:npe_scaled_mae}
\end{align}
The schema for this estimator is given in Figure~\ref{fig:NBE_multi_head_architecture},
while the validation losses are given in Figure~\ref{fig:validation_losses_npe_scaled}. We
further provide the unscaled, componentwise losses in
Figure~\ref{fig:validation_losses_npe_unscaled}, noting that early stopping and best-epoch
selection are done on the scaled losses. The scaled and unscaled objectives share the same
theoretical optimum, i.e., the posterior mean and median of $\bt \mid \x$ respectively, but their empirical minimizers can differ in practice. Indeed, coordinates with small prior support (short intervals) can result in smaller loss values and consequently, gradients. Scaling the loss as in~\eqref{eq:npe_scaled_mse} and in~\eqref{eq:npe_scaled_mae} ensures componentwise MSE and MAE losses are on the same scale.


\subsubsection{Neural surrogate pricers}\label{app:surrogate_pricers}

\begin{figure}[!t]
    \centering
    \includegraphics[width=8cm]{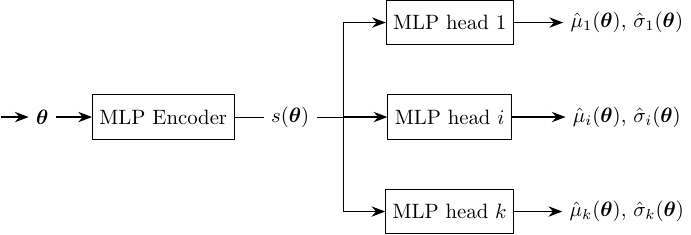}
    \caption{Schematic representation of the surrogate pricer architecture. For each family the exotic contracts (forward-start, barrier, realized-variance), the parameter vector $\bt$ is passed to an MLP encoder to obtain a set of informative summary statistics $s(\bt)$. Next $s(\bt)$ is passed to $k$ MLP heads to obtain estimates of the exotic options prices $P^{\text{exo},1},\ldots,P^{\text{exo},k}$, where $k$ is the number of contracts within the exotic family. 
    }
    \label{fig:surrogate_multi_head_architecture}
\end{figure}

\begin{figure}[!t]
    \centering
    \begin{subfigure}[t]{0.32\linewidth}
        \centering
        \includegraphics[width=\linewidth]
        {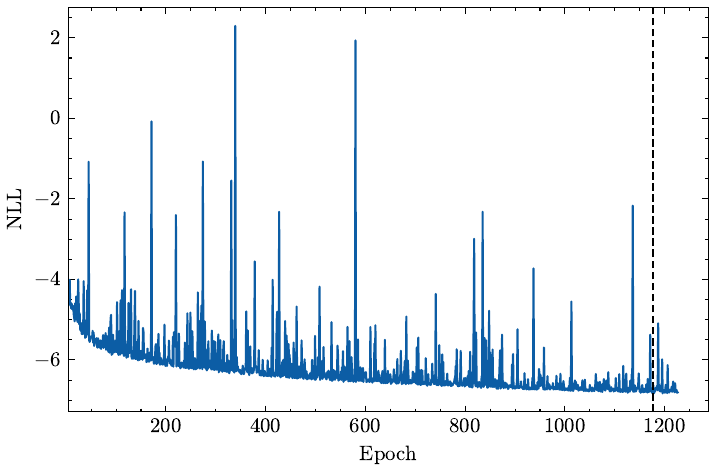}
        \caption{Forward-start}
    \end{subfigure}
    \hfill
    \begin{subfigure}[t]{0.32\linewidth}
        \centering
        \includegraphics[width=\linewidth]
        {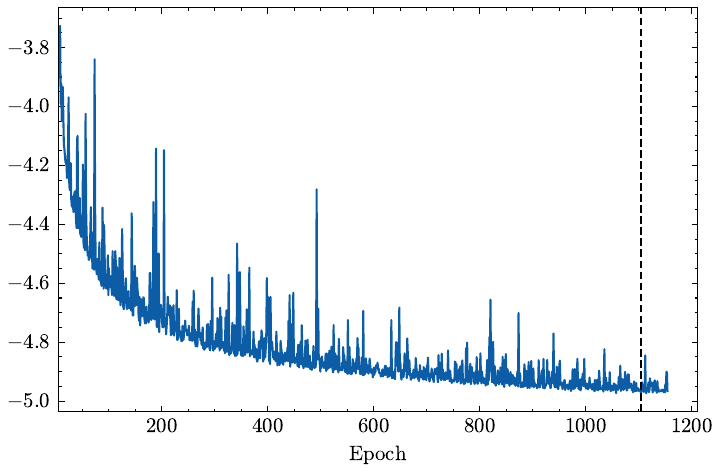}
        \caption{Barrier}
    \end{subfigure}
    \hfill
    \begin{subfigure}[t]{0.32\linewidth}
        \centering
        \includegraphics[width=\linewidth]
        {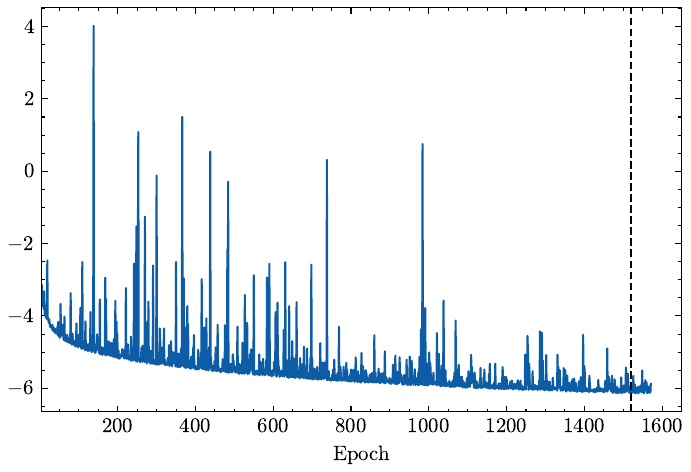}
        \caption{Realized variance}
    \end{subfigure}
    \caption{Validation negative log-likelihoods for the exotic-option surrogate models. The epoch selected through early stopping is marked in black.}
    \label{fig:validation_losses_surrogates}
\end{figure}

The neural surrogate pricers are similar to the neural point estimator above, but solve the forward problem. We train one network per exotic family, with one head for each contract in the aforementioned family. All forward-start and barrier contracts are retained. For realized-variance (RV) options, we retain $10$ active RV contracts: $c\in\{0.25,0.45\}$ at all four maturities $T\in\{1/12,1/4,1/2,1\}$, and $c=0.65$ at $T\in\{1/2,1\}$. This contract-level filtering removes some of the higher-strike contracts that contain a large fraction of MC prices $\hat{P}_i^{\mathrm{exo}}$ that are either exactly $0$ or economically irrelevant. Within the retained contracts, assume
\begin{equation}
    P_i^{\mathrm{exo}}+\epsilon\mid\bt
    \sim
    \operatorname{LogNormal} \left(\mu_i(\bt),\sigma_i^2(\bt)\right),
    \label{eq:LogNormal_eps_appendix}
\end{equation}
for some $\mu_i(\bt) \in \R$ and $\sigma_i(\bt) \ge 0$, where $\epsilon=0$ for forward-start and barrier options and $\epsilon=10^{-5}$ for RV options. This $\epsilon$ is used for numerical stability. Even within the retained RV contracts, there is still a small fraction of MC prices that are exactly zero, or extremely small, see Table~\ref{tab:rv_MC_diagnostics}. Thus a small, strictly-positive $\epsilon$ is required  to prevent such observations from disproportionately influencing the loss functions (which we discuss below) and damaging optimization. To this end, we use the multi-head architecture from Figure~\ref{fig:surrogate_multi_head_architecture}, in which each active head outputs a location $\hat{\mu}_i(\bt)$ and log-scale $\log\hat{\sigma}_i(\bt)$. Say we retain $k$ RV contracts. Then the network is trained by minimizing the average Gaussian negative log-likelihood (NLL)
\begin{equation}
    \mathcal L_{\mathrm{sur}}
    =
    \frac{1}{k}
    \sum_{i=1}^k
    \ev \left[
        \log\hat{\sigma}_i(\bt)
        +
        \frac{\left(\log{(\hat{P}_i^{\mathrm{exo}}+\epsilon)}-\hat{\mu}_i(\bt)\right)^2}{2\hat{\sigma}_i^2(\bt)}
    \right],
    \label{eq:training_surrogate_nll}
\end{equation}
where $\hat{P}_i^{\mathrm{exo}}$ denotes the MC approximation of the corresponding exotic contract price.
Note that the LogNormal assumption from ~\eqref{eq:LogNormal_eps_appendix} allows for mass on $[-\eps,0]$, a limitation that can in principle be removed by using more complex predictive distribution for $P^\textrm{exo} \mid \bt$. 
Validation losses are reported in Figure~\ref{fig:validation_losses_surrogates}.


\subsubsection{Telescoping neural ratio estimator (TRE) and marginal neural ratio estimator (NRE)}
\label{app:TRE_and_marginal_NRE}
\begin{figure}[t]
    \centering
    \includegraphics[width=0.8\linewidth]{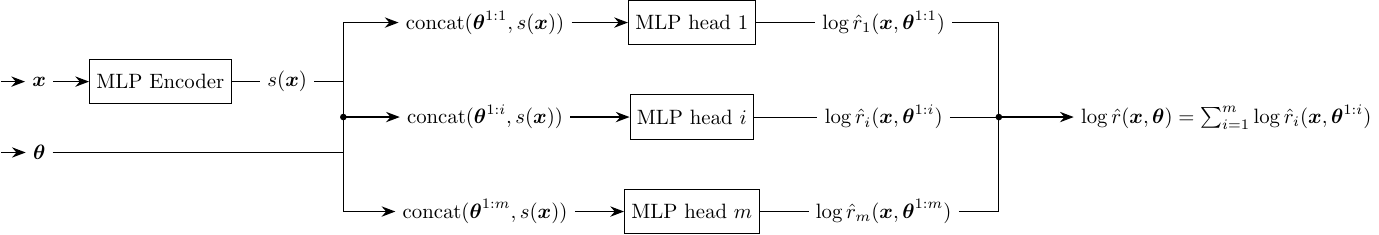}
    \caption{Schematic representation of the telescoping ratio estimator (TRE) architecture. The observed IV surface $\x$ is passed to an MLP network to obtain a set of informative summary statistics. Then, for $i \in \{1,\ldots,m\}$, the $i^\text{th}$ head inputs $s(\x)$ concatenated with the first $i$ components $\bt^{1:i} =(\theta^1,\ldots,\theta^i)$ of $\bt$ and outputs an estimate of $\log r_i(\x,\bt^{1:i})$. Summing the $m$ outputs yields an estimate of $\log r(\x,\bt)$.}
    \label{fig:TRE_multi_head_architecture}
\end{figure}

\begin{figure}[!ht]
    \centering
    \begin{subfigure}[t]{0.48\linewidth}
        \centering
        \includegraphics[width=\linewidth]
        {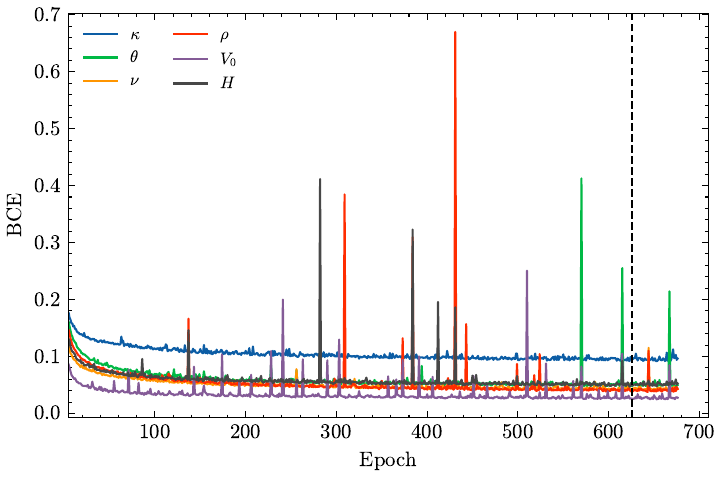}
        \caption{TRE.}
        \label{fig:val_loss_BCE_TRE}
    \end{subfigure}
    \hfill
    \begin{subfigure}[t]{0.48\linewidth}
        \centering
        \includegraphics[width=\linewidth]
        {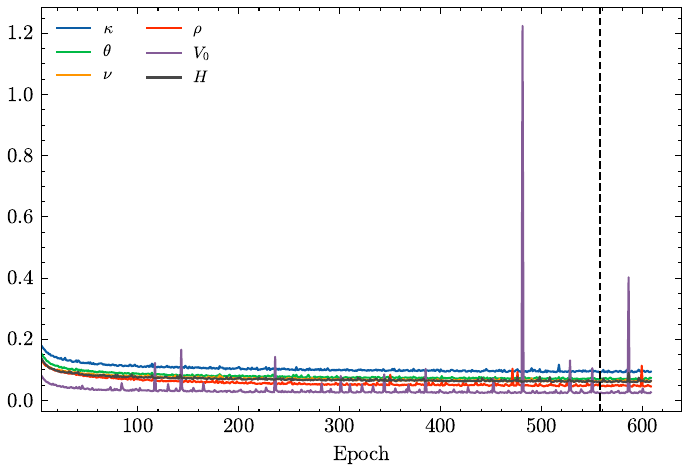}
        \caption{Marginal NRE.}
        \label{fig:val_loss_BCE_mNRE}
    \end{subfigure}
    \caption{Binary cross-entropy validation losses $\mathcal{L}_\textrm{TRE}$ and   $\mathcal{L}_\textrm{mNRE}$ for the two ratio-estimation models. The epoch selected through early stopping is marked in black.}
    \label{fig:validation_losses_ratio}
\end{figure}

We start with the TRE. Although it solves the inverse problem, it inputs both $\x$ and $\bt$ and outputs an approximation of the posterior density $p(\bt \mid \x)$. We employ the multi-head architecture from Figure~\ref{fig:TRE_multi_head_architecture}, in which the $i^\textrm{th}$ heads inputs $\x$ and the first $i$ coordinates $\bt^{1:i} = (\theta^1,\ldots,\theta^i$) and outputs an approximation of 
\begin{equation*}
    r_i(\x,\bt^{1:i}) = \frac{p(\theta^i \mid \x, \theta^1,\ldots,\theta^{i-1})}{p(\theta^i \mid \theta^1,\ldots, \theta^{i-1})}, \quad i =1,\ldots,m.
\end{equation*}
The full log likelihood-to-evidence is then recovered as a product. 
In our case, the chosen prior factorizes as $p(\bt) = p(\theta^1) \cdot \ldots \cdot p(\theta^m)$, see Table~\ref{tab:rheston_supports}, and the above ratio simplifies to
\begin{equation*}
        r_i(\x,\bt^{1:i}) = \frac{p(\theta^i \mid \x, \theta^1,\ldots,\theta^{i-1})}{p(\theta^i)}, \quad i =1,\ldots,m.
\end{equation*}
The full ratio is then recovered by
\begin{equation*}
    r(\x,\bt) = \prod_{i=1}^m r_i(\x,\bt^{1:i}).
\end{equation*}
We next discuss the loss functions used for each head. Let $c_{\textrm{TRE},i}$ be the classifier associated with the $i^\textrm{th}$ head. The corresponding binary-cross entropy (BCE) is then given by 
\begin{equation*}
    \mathcal{L}_{\textrm{TRE},i} =- \ev_{q_i^1(\x,\bt^{1:i})}\left[\log{c_{\textrm{TRE},i} \left(\x,\bt^{1:i} \right)}\right] - \ev_{q_i^0(\x,\bt^{1:i})}  \left[ \log{\left(1 - c_{\textrm{TRE},i} \left(\x,\bt^{1:i} \right)\right)}\right],
\end{equation*}
where 
\begin{align*}
    q_i^1(\x,\bt^{1:i})& = p(\x,\bt^{1:i}) \\
    q_i^0(\x,\bt^{1:i}) &= p(\x,\bt^{1:i-1}) p(\theta^i),
\end{align*} 
for $i=1,\ldots,m$, and the full loss function is given by
\begin{equation*}
    \mathcal{L}_\textrm{TRE}(\x,\bt) = \frac{1}{m} \sum_{i=1}^m \mathcal{L}_{\textrm{TRE},i}(\x,\bt^{1:i}).
\end{equation*}
Under the above BCE losses, the optimal Bayes classifiers satisfy 
\begin{equation*}
    c^{*}_{\text{TRE},i}(\x.\bt^{1:i}) = \frac{q_i^1(\x,\bt^{1:i})}{q_i^1(\x,\bt^{1:i}) + q_i^{0}(\x,\bt^{1:i})} = \frac{p(\theta^i \mid \x, \bt^{1:i-1})}{p(\theta^i \mid \x, \bt^{1:i-1}) + p(\theta^i)}
\end{equation*}
and yield the conditional posteriors
\begin{equation*}
    r_i(\x,\bt^{1:i}) = \frac{c_{\textrm{TRE},i}^{*}(\x,\bt^{1:i})}{1 - c_{\textrm{TRE},i}^{*}(\x,\bt^{1:i})} ,\quad i = 1, \ldots,m.
\end{equation*}
We defer the reader to the end of this subsection for post-training calibration technique and assessing the quality of the learnt estimators. A detailed analysis of the TRE methodology is available in \cite{leonte2025tre}. Per-head validation BCE losses are reported in Figure~\ref{fig:val_loss_BCE_TRE}.

We now turn to the marginal NRE, which is needed for the XAI analysis 
from Section~\ref{sec:simstudy-xai}. Indeed, there we need to learn the posterior marginals $p(\theta^i \mid \x)$ for $i=1,\ldots,m$, rather than the full posterior $p(\bt \mid \x)$. Note that marginalizing posterior samples produces samples from the marginal posteriors, but it does not give direct access to the marginal posterior densities, hence a second neural estimator is required. 

\begin{figure}[!ht]
    \centering
    \includegraphics[width=0.8\linewidth]{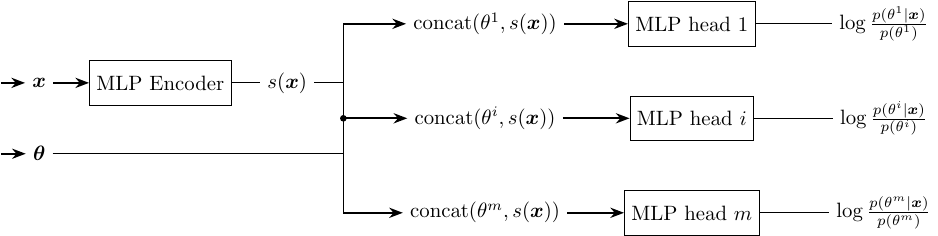}
    \caption{Schematic representation of the marginal neural ratio estimator (NRE) architecture. The observed IV surface $\x$ is passed to an MLP network to obtain a set of informative summary statistics. Then, for $i \in \{1,\ldots,m\}$, the $i^\text{th}$ head inputs $s(\x)$ concatenated with $\theta^i$ of $\bt$ and outputs an estimate of $\log \frac{p(\theta^i \mid x)}{p(\theta^i)}$, $i=1,\ldots,m.$}
    \label{fig:marginal_NRE_multi_head_architecture}
\end{figure}

For the marginal NRE, we employ the architecture from Figure~\ref{fig:marginal_NRE_multi_head_architecture}. The loss functions associated with head $i$ is the following BCE expression: 
\begin{equation*}
    \mathcal{L}_{\textrm{mNRE},i} = -\ev_{p(\x,\theta^i)}\left[\log{c_{\textrm{mNRE},i} \left(\x,\theta^i \right)}\right] - \ev_{p(\x) p(\theta^i)} \left[\log{\left(1 - c_{\textrm{mNRE},i} \left(\x,\theta^i \right)\right)}\right]
\end{equation*}
and the full loss function is given by
\begin{equation*}
    \mathcal{L}_\textrm{mNRE}(\x,\bt) = \frac{1}{m} \sum_{i=1}^m \mathcal{L}_{\textrm{mNRE},i}(\x,\theta^i).
\end{equation*}
Under the above BCE losses, the optimal Bayes classifiers satisfy 
\begin{equation*}
    c^{*}_{\text{mNRE},i}(\x.\bt^{1:i}) =  \frac{p(\theta^i \mid \x)}{p(\theta^i \mid \x) + p(\theta^i)},
\end{equation*}
and yield the marginal conditional posteriors
\begin{equation*}
   \frac{p(\theta^i \mid x)}{p(\theta^i)} = \frac{c_{\textrm{mNRE},i}^{*}(\x,\bt^{i})}{1 - c_{\textrm{mNRE},i}^{*}(\x,\bt^{i})} ,\quad i = 1, \ldots,m.
\end{equation*} 
Per-head validation BCE losses are provided in Figure~\ref{fig:val_loss_BCE_mNRE}.

\clearpage

\begin{remark}
    The ratio-estimators and, to a lesser extent the other estimators, can exhibit occasional spikes. For the ratio-estimators, this likely reflects
rare high-confidence classification errors. This is not an issue, as we select the set of weights and biases corresponding to the epoch achieving the lowest validation loss, and not those corresponding to the last epoch. The substantive validation of the fitted models is
therefore based on simulation-based calibration checks, together with the held-out posterior-coverage and downstream predictive-coverage diagnostics reported in this paper, rather than on visual smoothness of the training curves.
\end{remark}

\textbf{Further TRE details and diagnoistcs.} 
We discuss two aspects in this last part: i) post-training calibration of the TRE and ii) checks on the quality of the approximate posterior.

Having trained the $m$ classifiers $\hat{c}_i \defeq  \hat{c}_{\textrm{TRE},i}$ underlying the TRE, these classifiers yield estimated conditional density ratios $\hat{r}_i = \hat{c}_i/(1-\hat{c}_i)$, and hence approximations $\hat{p}(\theta^i \mid \x, \bt^{1:i-1})$ for $i=1,\ldots,m$. By definition, these approximations are not exact, and classification errors can propagate to the estimated posterior densities. A useful feature of classifier-based ratio estimators, and one of the reasons we chose to employ this methodology is that we can apply a post-training calibration step and thereby obtain (potentially) improved posterior approximations without retraining the neural networks. Specifically, for each head, we fit a calibration map $T_i:[0,1]\to[0,1]$ and define 
\begin{align*}
    \hat{c}_{i,\mathrm{cal}} &= T_i\circ\hat{c}_i,\\
    \hat{r}_{i,\mathrm{cal}} &= \hat{c}_{i,\mathrm{cal}}/(1-\hat{c}_{i,\mathrm{cal}}).
\end{align*}
This post-training classifier calibration is distinct from \emph{parameter calibration}, by which we mean recovering the rough Heston parameters from an observed IV surface.

Among common post-training calibration approaches, we consider Platt scaling \citep{platt1999probabilistic}, $\beta$-calibration \citep{kull2017beta}, and isotonic regression \citep{caruana2004predicting}. We test all three and empirically find that both Platt scaling and isotonic regression can degrade performance. Platt scaling is relatively restrictive, while isotonic regression can produce nearly degenerate posterior approximations. We therefore use $\beta$-calibration, which applies the monotone transformation 
\begin{equation*}
    T(s;a,b,c)= 1/ \left(1+e^{-c}(1-s)^b s^{-a}\right),
\end{equation*}
 with $a,b>0$ and $c\in \R$. Importantly, this family contains the identity map for $a=b=1$ and $c=0$, so an already well-calibrated classifier can remain unchanged.  Further, $\beta$-calibration is more flexible than Platt scaling while remaining parametric and well behaved in our setting. The calibration parameters are fitted separately for each TRE head, with the BCE loss. We now turn our attention to simulation-based calibration (SBC) checks.

\textbf{Sequential checks.} Having trained and $\beta$-calibrated the TRE, we assess the empirical coverage of the learnt approximate conditional posterior densities. For any density $q$, let $\Theta_q(1-\alpha)$ denote its $1-\alpha$ highest posterior density (HPD) region, defined as
\begin{equation}
    \Theta_q(1-\alpha)
    =
    \left\{
        \vartheta :
        q(\vartheta) \geq \gamma_{q}(\alpha)
    \right\},
    \label{eq:general_HPD_region}
\end{equation}
where $\gamma_{q}(\alpha)$ is chosen so that
\begin{equation}
    \int_{\vartheta \in \Theta_q(1-\alpha)} q(\vartheta)\,\mathrm{d}\vartheta
    =1-\alpha.
    \label{eq:general_HPD_probability}
\end{equation}
For the $i^\textrm{th}$ TRE head, we take $q(\vartheta^i)=\hat{p}(\vartheta^i\mid \x,\bt^{1:i-1})$. For each parameter $\theta^i$, define the (averaged) coverage of the corresponding approximate $1-\alpha$ HPD regions by
\begin{equation}
    \mathcal{C}^{i}_{1-\alpha}
    =
    \ev_{p(\x,\bt^{1:i})}\left[
        \mathbf{1}\left(
            \theta^i \in
            \Theta_{\hat{p}(\cdot\mid \x,\bt^{1:i-1})}(1-\alpha)
        \right)
    \right], \qquad i=1,\ldots,m.
    \label{eq:sequential_coverage}
\end{equation}
In practice, we estimate \eqref{eq:sequential_coverage} from $N$ pairs $(\x_j,\bt_j)$. For each pair and parameter $i$, we draw $L$ samples $\vartheta^i_{j,1},\ldots,\vartheta^i_{j,L}\sim \hat{p}(\theta^i\mid\x_j,\bt_j^{1:i-1})$, evaluate and sort the corresponding approximate posterior densities $p(\vartheta^i_{j,1} \mid \x_j, \bt^{1:i-1}),\ldots,p(\vartheta^i_{j,L} \mid \x_j, \bt^{1:i-1})$ in descending order. Then, the smallest value among the top $(1-\alpha)L$ samples defines the empirical HPD threshold $\hat{\gamma}^i_{j}(\alpha)$. The empirical coverage is then given by
\begin{equation}
    \widehat{\mathcal{C}}^{i}_{1-\alpha}
    =
    \frac{1}{N}\sum_{j=1}^{N}
    \mathbf{1}\left[
        \hat{p}(\theta^i_j\mid\x_j,\bt_j^{1:i-1})
        \geq \gamma^i_{j}(\alpha)
    \right].
    \label{eq:sequential_coverage_empirical}
\end{equation}

\begin{figure}[t]
    \centering
    \begin{subfigure}[t]{0.48\linewidth}
        \centering
        \includegraphics[width=\linewidth]{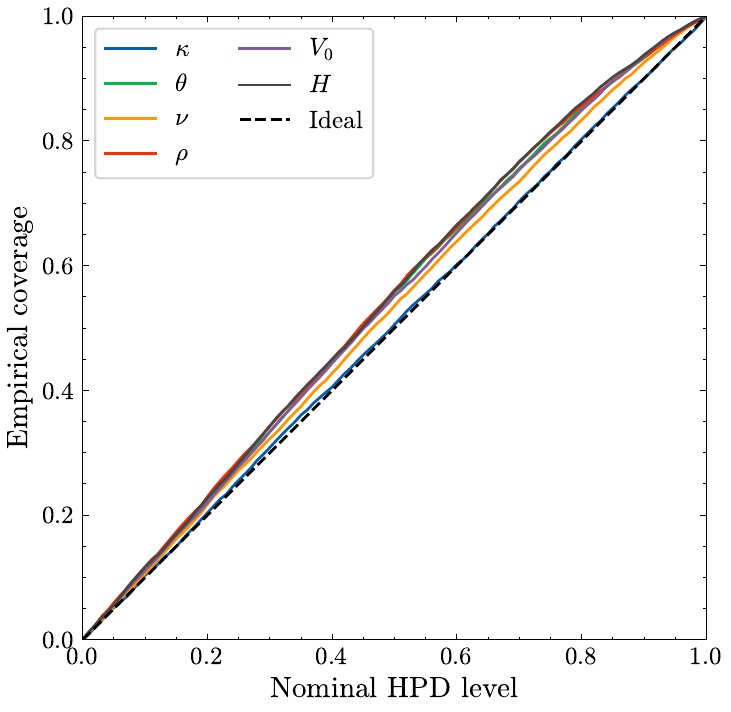}
        \caption{Uncalibrated.}
        \label{fig:TRE_seq_coverage_beta_appendix}
    \end{subfigure}
    \hfill
    \begin{subfigure}[t]{0.48\linewidth}
        \centering
        \includegraphics[width=\linewidth]{Paper1/figures/seq_coverage_beta_calibrated.pdf}
        \caption{Beta-calibrated.}
        \label{fig:TRE_seq_coverage_uncal}
    \end{subfigure}
    \caption{Empirical coverage of the HPD regions of the learnt posterior approximations $\hat{p}(\kappa \mid \x), \ldots, \hat{p}(H \mid \x, \kappa, \theta, \nu, \rho, V_0)$. We note that the $\beta$-calibrated classifiers yield posterior densities with near-perfect coverage, and that $\beta$-calibration qualitatively improves the coverage properties of the learnt conditional distributions.}
    \label{fig:TRE_seq_coverage}
\end{figure}

If the $i^\textrm{th}$ TRE classifier is well-calibrated, or equivalently if the corresponding conditional posterior approximation is accurate, then $\mathcal{C}^{i}_{1-\alpha}=1-\alpha$. All in all, we apply this check and presents results in Figure~\ref{fig:TRE_seq_coverage}. While the uncalibrated TRE classifiers display some miscalibration, specifically under-confidence, the $\beta$- calibrated versions have near-perfect coverage.

\subsection{Grid hyperparameter search}
\label{app:nn_hyperparameter_search}

Hyperparameters are selected based on the loss evaluated on the validation set. After each epoch, we evaluate and compare the validation loss with its previous best value. If the validation loss improves by more than a tolerance $\delta$, the current weights and biases are set as the new best checkpoint and the patience counter is reset; otherwise, the counter is incremented, and training stops once $50$ consecutive epochs pass without improvement above $\delta$. We use $\delta=10^{-6}$ for the (scaled) MSE-trained neural point estimator, $\delta=10^{-5}$ for the surrogate pricers and $\delta=10^{-4}$ for the (scaled) MAE-trained point estimator and ratio estimators, specifically the TRE and marginal NRE. Inspection of the validation losses confirms that these tolerances are sufficiently small not to induce materially earlier stopping. Dropout, when used, is applied to the shared encoder, while the output heads use no dropout.

\subsubsection{Neural point and ratio estimators}

For the inverse estimators we use the hyperparameter grid in Table~\ref{tab:inverse_nn_grid}. This gives $6\times2\times4\times2\times2=192$ configurations for each of three estimators. Neural point estimators are ranked by the corresponding scaled validation MSE or MAE losses, whereas marginal and telescoping ratio estimators are ranked by validation BCE.

\begin{table}[t]
\centering
\caption{Hyperparameter grid for the neural point estimator (NPE), marginal NRE and TRE.}
\label{tab:inverse_nn_grid}
\begin{tabular}{ll}
\toprule
Hyperparameter & Values \\
\midrule
Shared-encoder widths
& $(64,64,64)$, $(128,64)$, $(64,64)$, $(64,32)$, $(32,32)$, $(32,16)$ \\
Encoder dropout &   $0.1$, $0$ \\
Head widths & $(16,16,1)$, $(8,8,1)$, $(16,1)$, $(8,1)$ \\
Learning rate & $10^{-3}$, $5\times10^{-4}$ \\
AdamW weight decay & $10^{-3}$, $10^{-4}$ \\
\bottomrule
\end{tabular}
\end{table}

\subsubsection{Surrogate pricers}

For the forward estimators, a separate hyperparameter search is performed for each exotic family. The search grid is given in Table~\ref{tab:surrogate_nn_grid}, yielding $4\times4\times2\times2\times2=128$ configurations per family. Models are ranked by validation Gaussian negative log-likelihood on the log-prices.

\begin{table}[t]
\centering
\caption{Hyperparameter grid for the exotic surrogate pricers.}
\label{tab:surrogate_nn_grid}
\begin{tabular}{ll}
\toprule
Hyperparameter & Values \\
\midrule
Shared-encoder widths
& $(16,32,64,128)$, $(16,32,64)$, $(16,16)$, $(16,32)$ \\
Encoder dropout &  $0.1$, $0$ \\
Head widths & $(16,8,4,1)$, $(16,16,1)$, $(8,8,1)$, $(8,4,1)$,  \\
Learning rate & $10^{-3}$, $5\times10^{-4}$ \\
AdamW weight decay &   $10^{-3}$ , $10^{-4}$ \\
\bottomrule
\end{tabular}
\end{table}

\end{document}